\documentclass[lettersize,journal]{IEEEtran}
\usepackage{amsmath,amsfonts}
\usepackage{algorithmic}
\usepackage{algorithm}
\usepackage{array}
\usepackage[caption=false,font=normalsize,labelfont=sf,textfont=sf]{subfig}
\usepackage{textcomp}
\usepackage{stfloats}
\usepackage{url}
\usepackage{verbatim}
\usepackage{graphicx}
\usepackage{cite}
\usepackage[utf8]{inputenc} 
\usepackage[T1]{fontenc}    
\usepackage{hyperref}       
\usepackage{url}            
\usepackage{booktabs}       
\usepackage{amsfonts}       
\usepackage{nicefrac}       
\usepackage{microtype}      
\usepackage{xcolor}         

\newcommand{\ie}{\emph{i.e.}\;}

\newcommand{\etc}{etc.\;}
\usepackage[scr=boondoxo,scrscaled=1.05]{mathalfa}
\usepackage{amsmath}
\usepackage{graphicx}
\usepackage{wrapfig}
\usepackage{multirow}
\usepackage{bigstrut}
\usepackage{comment}
\usepackage{arydshln}
\usepackage{makecell}
\usepackage{siunitx}
\begin{document}

\title{Deep LoRA-Unfolding Networks for Image Restoration}
\author{Xiangming Wang, Haijin Zeng, Benteng Sun, Jiezhang Cao, Kai Zhang, Qiangqiang Shen, Yongyong~Chen,~\IEEEmembership{Member,~IEEE}
\thanks{This work was supported by the National Natural Science Foundation of China under Grant 62576118 and by Guangdong Major Project of Basic and Applied Basic Research under Grant 2023B0303000010. {(Corresponding authors: Haijin Zeng and Yongyong Chen.)}}
\thanks{Xiangming Wang, Haijin Zeng, Benteng Sun and Yongyong~Chen are with the School of Computer Science and Technology, Harbin Institute of Technology (Shenzhen), Shenzhen 518055, Guangdong, China, (Email: YongyongChen.cn@gmail.com).} 
\thanks{Jiezhang Cao is with the Institute of Image Communication and Network Engineering, Shanghai Jiao Tong University, Shanghai, China, (Email: caojiezhang@gmail.com).}
\thanks{Kai Zhang is with School of Intelligence Science and Technology, Nanjing University, Suzhou, China, (Email: kaizhang@nju.edu.cn).}
\thanks{Qiangqiang~Shen is with the School of Electronics and Information Engineering, Harbin Institute of Technology (Shenzhen), Shenzhen 518055, Guangdong, China, (Email: 1120810623@hit.edu.cn).}
\thanks{Manuscript received April 19, 2021; revised August 16, 2021.}}

\markboth{Journal of \LaTeX\ Class Files,~Vol.~14, No.~8, August~2021}%
{Shell \MakeLowercase{\textit{et al.}}: A Sample Article Using IEEEtran.cls for IEEE Journals}


\maketitle

\begin{abstract}

Deep unfolding networks (DUNs), combining conventional iterative optimization algorithms and deep neural networks into a multi-stage framework, have achieved remarkable accomplishments in Image Restoration (IR), such as spectral imaging reconstruction, compressive sensing and super-resolution.
It unfolds the iterative optimization steps into a stack of sequentially linked blocks.
Each block consists of a Gradient Descent Module (GDM) and a Proximal Mapping Module (PMM) which is equivalent to a denoiser from a Bayesian perspective, operating on Gaussian noise with a known level.
However, existing DUNs suffer from two critical limitations: (i) their PMMs share identical architectures and denoising objectives across stages, ignoring the need for stage-specific adaptation to varying noise levels; and (ii) their chain of structurally repetitive blocks results in severe parameter redundancy and high memory consumption, hindering deployment in large-scale or resource-constrained scenarios.
To address these challenges, we introduce generalized Deep Low-rank Adaptation (LoRA) Unfolding Networks for image restoration, named LoRun, harmonizing denoising objectives and adapting different denoising levels between stages with compressed memory usage for more efficient DUN.
LoRun introduces a novel paradigm where a single pretrained base denoiser is shared across all stages, while lightweight, stage-specific LoRA adapters are injected into the PMMs to dynamically modulate denoising behavior according to the noise level at each unfolding step.
This design decouples the core restoration capability from task-specific adaptation, enabling precise control over denoising intensity without duplicating full network parameters and achieving up to $N$ times parameter reduction for an $N$-stage DUN with on-par or better performance.
Extensive experiments conducted on three IR tasks validate the efficiency of our method.
\end{abstract}

\begin{IEEEkeywords}
Deep unfolding network, Low-rank adaption, Image restoration, Proximal mapping module.
\end{IEEEkeywords}

\section{Introduction}

\begin{figure}
\centering
\includegraphics[width=0.48\textwidth]{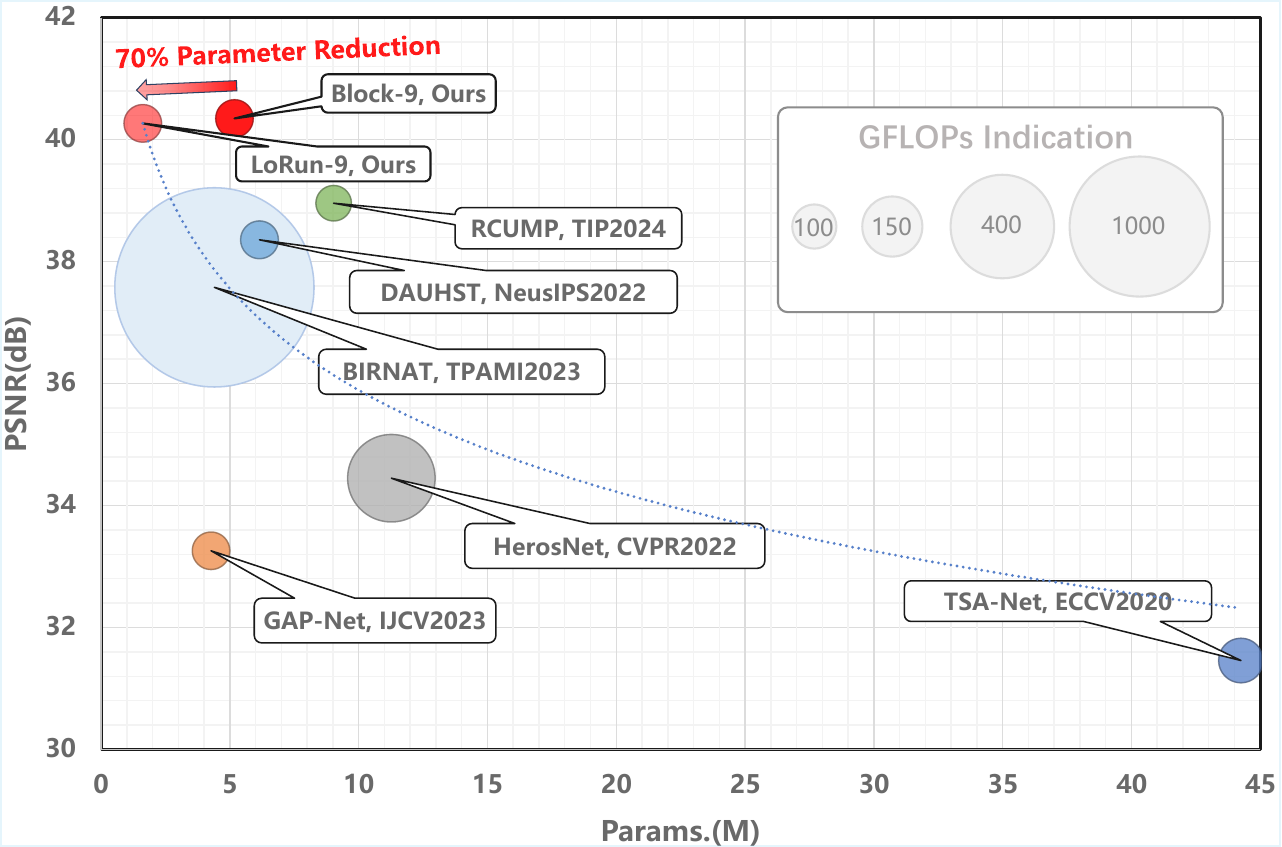}
\vspace{-1mm}
\caption{{PSNR-Params.-FLOPs comparison of our LoRun and SOTAs in CASSI task. Our LoRun reduces the parameters of traditional DUN methods by nearly \textbf{70\%} while maintaining comparable performance.}}
\vspace{-5mm}
\label{fig:brief}
\end{figure}

Among Image Restoration (IR) applications, the observed data usually suffer from various degradations, such as optical distortions \cite{chen2023hierarchical,pan2020cascaded} due to lens imperfections and shooting conditions, noise from photoelectric conversion and compression because of transmission and storage \cite{gap_tv,dong2014compressive,mou2022deep,herosnet,desci,cai2022degradation}. 
The degradation process can be formulated as 
\begin{equation}
\label{model}
    \mathbf{y} = \Phi \mathbf{x} + \mathbf{n},
\end{equation}
where $\mathbf{n}$ denotes the noise or errors in the imaging system and $\Phi$ represents the degradation matrix linked to the corresponding degradation.
Thus, how to uncover the ``clean'' image $\mathbf{x}$ from ``degraded'' measurement $\mathbf{y}$ has become the key part of IR.

To tackle the ill-posedness of the IR problem \cite{wu2025dswinir, zhou2024seeing, liang2025ntire, zhang2026clearair, gao2025learning, jiang2025survey}, a flurry of methods has been developed, which can be roughly divided into three categories: model-based methods, learning-based methods and Deep Unfolding Network (DUN) methods.
Model-based methods \cite{twist,gap_tv,NonLRMA,LLxRGTV,E3DTV,wang2025otlrm} mainly borrow manually-designed mathematical models as regularizers and optimize the reconstruction process by the following unconstrained optimization model:
\begin{equation}
\label{optimization}
    \hat{\mathbf{x}} = \mathop{\arg\min}\limits_{\mathbf{x}} \frac{1}{2} \Vert \mathbf{y}-\Phi \mathbf{x} \Vert^2_2 + \lambda \Psi(\mathbf{x}),
\end{equation}
where $\Psi(\cdot)$ denotes the regularization term which depends on the image priors such as the low-rank prior for singular value compression \cite{chen2019low}, sparse prior for redundant information removal \cite{li2018fusing} and total variation prior for local smoothness mining \cite{FCTV}.
Some iterative algorithms such as Alternating Direction Methods of Multipliers (ADMM) \cite{tra_2} and Iterative Shrinkage Thresholding Algorithm (ISTA) \cite{tra_3} are adopted to solve Eq. \eqref{optimization}.
While these model-based approaches are theoretically robust and interpretable, they are often less effective compared to deep learning approaches, mainly due to the weak adaptability and limited domain-specific knowledge of handcrafted priors.
Benefiting from the informative datasets and efficient structures, such as U-Net \cite{unet} and Transformer \cite{liang2021swinir}, more abundant and generalized underlying patterns are extracted by deep learning-based methods \cite{mst,cst,wang2025simulator,liu2025graph,ddrm,dds2m,zamir2021multi}.
However, how and why the modules work has not yet been completely and explicitly demonstrated, indicating the interpretability of deep learning methods is limited.

To incorporate theoretical properties and flexibility of the above two types, {DUN methods \cite{sun2016deep,song2021memory,cai2022degradation,tra_3,meng2023deep,admm-net,mou2022deep,zeng2025vision} unfold the iterative optimization steps into a multi-stage framework, inject the deep module as the regularizers and train the unfolding framework in an end-to-end manner.}
One iteration is converted into one block, which contains a Gradient Descent Module (GDM) and a Proximal Mapping Module (PMM) as shown in Fig. \ref{fig:main}(c).
GDM produces output from the previous block using the degradation transform $\Phi$ defined in Eq. \eqref{model}, integrating degradation information into the DUN model.
From the Bayesian perspective, PMM can be viewed as a denoising process with stage-specific known Gaussian noise level, \ie a denoiser \cite{chan2016plug} (Please refer to Section \ref{dun} for detailed representation).
This derivation enhances the performance and interpretability of stage-aware denoising capabilities.

However, conventional DUN methods usually treat multiple structurally identical blocks stacked together as parameter independent for better performance.
Under the multi-stage chain structure, this strategy is difficult to steer the accurate direction of each stage's optimization with only supervision of the last block and no explicit inter-block connections, leading to less effective multi-level denoising abilities.
Meanwhile, numerous structurally identical blocks need large and redundant parameters required for the target task, leading to high memory and parameter costs which poses significant challenges for real-world deployment especially when handling large-scale DUN models.
Therefore, a robust module construction and training mechanism for establishing hierarchical denoising capabilities with compressed but sufficient parameters is essential.

In this paper, we introduce Low-Rank Adaptation (LoRA) \cite{hu2021lora,ding2023sparse,valipour2022dylora,zhang2023adaptive} from the LLM fine-tuning field into the DUN framework and propose a novel trade-off strategy, LoRun which is a deep LoRA-induced unfolding network for IR.
Motivated by the optimization derivations and identically structured modules of the DUN framework, we leverage the strong similarity and the low ``intrinsic rank'' \cite{hu2021lora} of parameters among stages, promoting low-rank priors with different lightweight adapters and one same backbone.
Concretely, we decouple the weights in the DUN model into one fundamental denoiser across stages and several small LoRA modules for low-rank updates.
Thus our LoRun framework can not only generate the basic denoising target, but also enable multi-level denoising capability for each stage with much fewer parameters.
Fig. \ref{fig:brief} shows the PSNR-FLOPs-Params. comparison of our LoRun and the state-of-the-arts (SOTAs) in spectral imaging construction which shows the superiority in terms of parameters and performance.
And this strategy can be seamlessly applied to DUNs based on different iterative optimization algorithms, different structures of the denoiser and different applications, which demonstrates the superiority of our LoRun in terms of generalizability.
The primary contributions are outlined as follows:
\begin{itemize}
    \item We propose LoRun, a generic model-free and task-free DUN framework, which incorporates low-rank modules into each structurally identical block for dynamic adaptation, enabling backbone-guided multi-level denoising abilities for each stage without any reliance on specified optimization algorithms, models and tasks.
    \item LoRun trains only one backbone with small LoRA modules without full parameter tuning, which achieves on-par or better performance while much compressing memory usage and parameters with reduction up to $N$-times for an $N$-stage DUN.
    \item LoRun decouples the backbone from LoRA, enabling efficient and flexible switching between different tasks or modes by simply replacing the LoRA modules.
    \item Extensive experiments demonstrate comparable results with compressed parameters in three typical IR tasks including spectral imaging reconstruction, compressive sensing and image super-resolution.
\end{itemize}

The rest of the paper is organized as follows.
Section \ref{related_works} gives the brief related works about DUN methods for IR and LoRA.
Then we give the detailed description about our LoRun method (which is related to our motivation and the overview pipeline) in Section \ref{methods}.
Section \ref{applications_and_experiments} presents the experiment settings and results of three applications.
Section \ref{conclusion} concludes our LoRun on the proposed framework.

\section{Related Works}
\label{related_works}
\noindent\textbf{Deep Unfolding Networks for Image Restoration.}
Deep unfolding networks aim to enhance IR tasks by unfolding an iterative optimization algorithm into an end-to-end neural network architecture.
For example, ISTA-Net \cite{tra_3} unfolded the ISTA optimization algorithm and solved the proximal mapping module with sparsity-induced nonlinear transforms.
Based on the tensor singular value thresholding, ADMM-Net \cite{admm-net} followed the ADMM algorithm and constructed a layer-wise structure for video snapshot compressive imaging (SCI).
And GAP-Net \cite{gapnet}, which is based on the generalized alternating projection (GAP) algorithm, proved a probabilistic global convergence result and evaluated its performance in video and spectral SCI problems.
\cite{song2021memory} addressed the compressive sensing problem with a known gradient descent module.
DAUHST \cite{cai2022degradation} expanded the half-quadratic splitting algorithm and adopted a half-shuffle transformer as the denoiser to capture local and non-local information for SCI.
{Recently, DGUNet \cite{mou2022deep}, which is a generalized DUN method for multiple IR tasks, introduced a flexible gradient descent module with learnable degradation transform for real-world image degradation and built an inter-stage feature pathway across the proximal mapping modules.
\textit{However, the above generalized DUN methods mostly trained multiple identically structured blocks with independent parameters, ignoring the generalization of homogenous denoising target and hierarchical distinctions between stages, and the large parameters also limited their practical applications.}}

\noindent\textbf{Low-Rank Adaptation.}
The backbone method of LoRA \cite{hu2021lora} aims at the large-scale pretraining on general domain data and adaptation to particular sub-tasks for LLM.
It decomposes the update of weights into the product of two small-scale matrices $\mathbf{A}$ and $\mathbf{B}$.
LoRA freezes the weights of the pretrained model and injects trainable low-rank decomposition matrices into each layer of the target module.
With LoRA as a fine-tuning strategy, large general models can be fine-tuned by fewer parameters with no additional inference latency.
Some variants of LoRA have been proposed, such as DyLoRA \cite{valipour2022dylora} which aims to address the problem of rank estimation by setting a range of ranks instead of fixed values.
Dynamic rank estimation is important and beneficial for the demonstration of LoRA effects.
Based on singular value decomposition, AdaLoRA \cite{zhang2023adaptive} 
adopted the importance-driven allocation strategy to adaptively adjust the parameter budget of the weight matrices and finally automatically remove relatively unimportant singular values.
Furthermore, SoRA \cite{ding2023sparse}, which induced the sparse prior into the singular value matrix, enables sparse-based dynamic rank adjustments by constructing a gate unit optimized with the proximal gradient method in the training stage.
\textit{However, LoRA currently shows promising performance for fine-tuned training of large models (such as Diffusion or Large Language Models), and is still unknown in the DUN framework.}

\section{Methods}
\label{methods}

In this section, we first present the motivation and overview structure of our proposed LoRun in Section \ref{lora_unfolding}, then Section \ref{lora_details} and Section \ref{dun} explain the details of the fine-tuning strategy LoRA and two classical DUN frameworks.

\subsection{Deep LoRA-Unfolding Network for Image Restoration}
\label{lora_unfolding}
\begin{figure}
    \centering
    \includegraphics[width=1\linewidth]{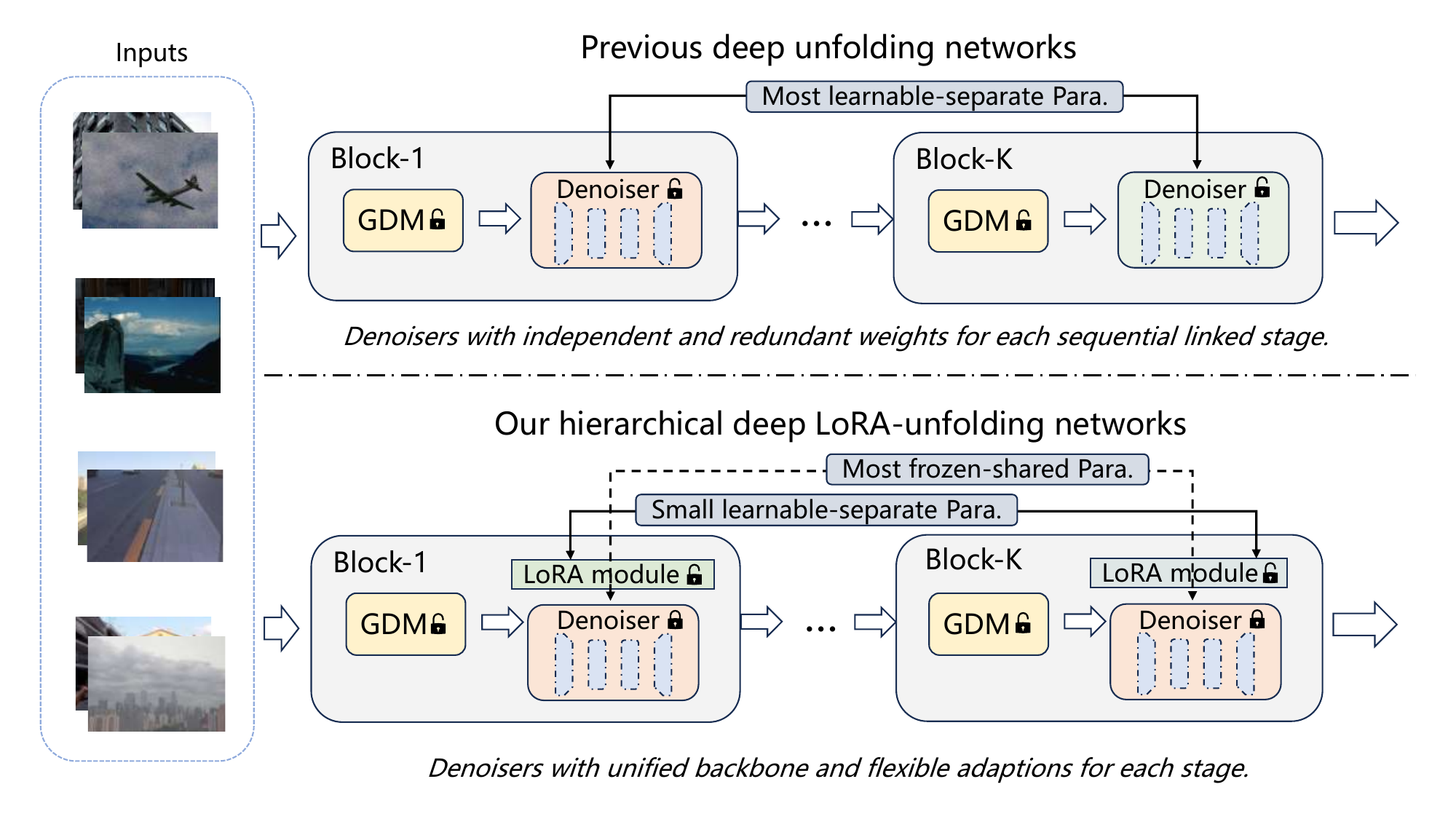}
    \vspace{-6mm}
    \caption{Structure comparison between the previous DUNs and our hierarchical deep LoRA-unfolding networks. Previous DUNs adopt independent parameters to all stages while our LoRun first leverages pre-trained denoiser for each stage then is trained with different LoRA layers for different denoising abilities.}
    \vspace{-4mm}
    \label{fig:LoRun_framework_clarify}
\end{figure}

\begin{figure}
\centering
\includegraphics[width=0.49\textwidth]{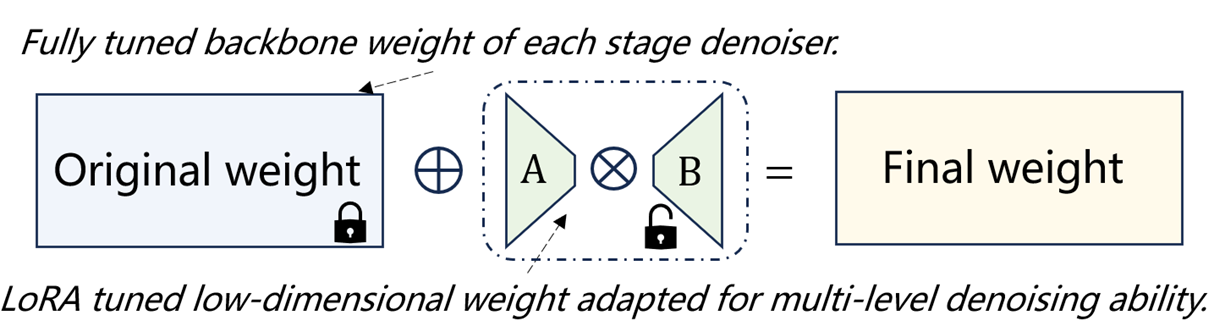}
\vspace{-6mm}
\caption{Representation for LoRA adopted in our LoRun framework.}
\label{fig:lora}
\vspace{-4mm}
\end{figure}

\begin{figure*}
    \centering
    \includegraphics[width=1\linewidth]{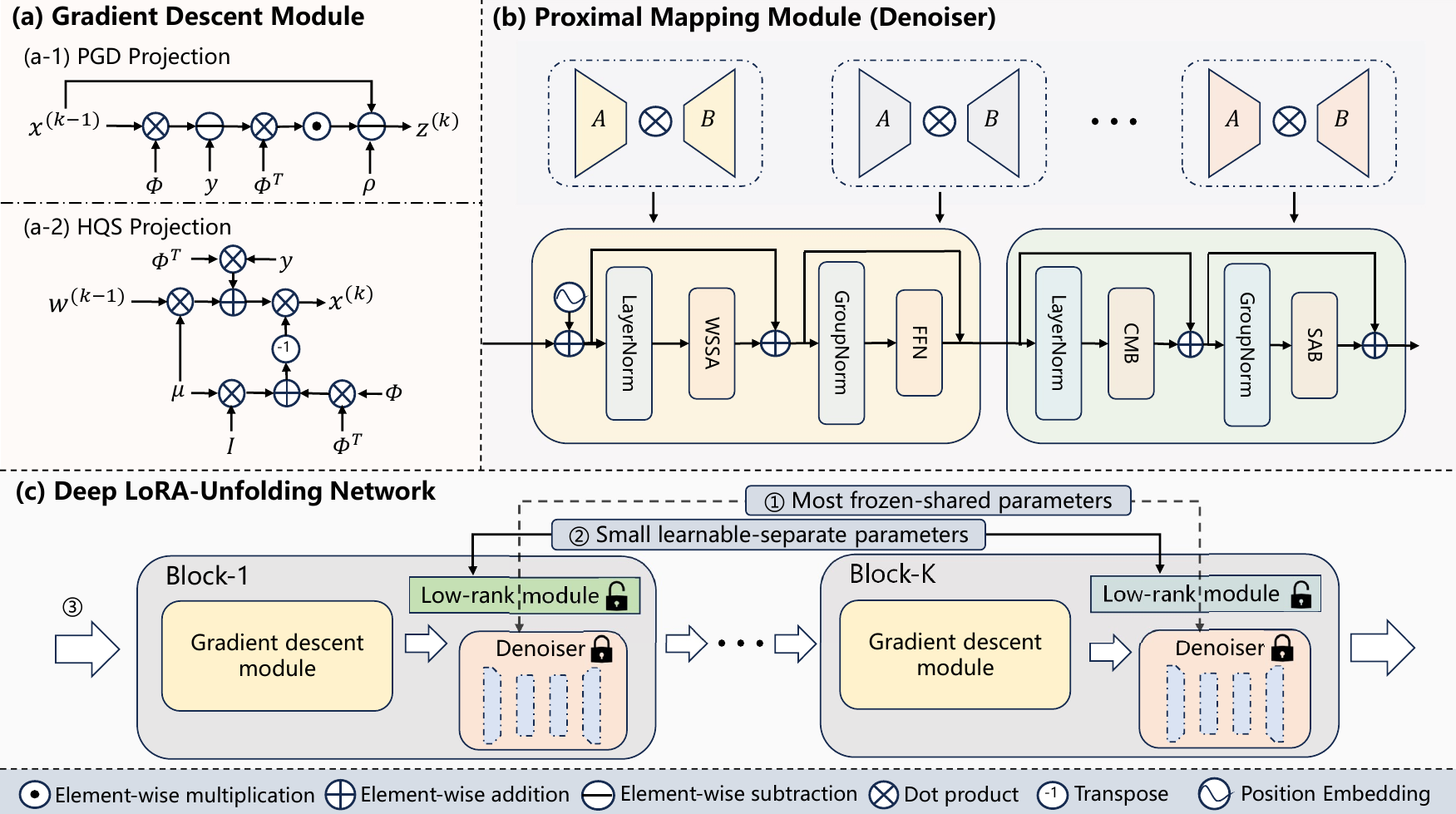}
    \vspace{-4mm}
    \caption{Illustration of the proposed LoRun method. \textbf{(a)} The gradient descent module. (a-1) The gradient descent process based on the PGD algorithm. (a-2) The gradient descent process based on the HQS algorithm. \textbf{(b)} The proximal mapping module (denoiser). \textbf{(c)} Framework of the proposed deep LoRA-Unfolding network. Our LoRun first adopts the same most frozen-shared backbone denoiser among all stages, followed by fine-tuning the denoising level and direction of each stage with small learnable-separate low-rank modules. This strategy mostly reduces the parameters and accelerates convergence.}
    \label{fig:main}
    \vspace{-4mm}
\end{figure*}

\noindent \textbf{Motivation}
Derived from the iterative optimization algorithms, Eqs. \eqref{gradient_descent} and \eqref{proximal_mapping} describe two operations in the $k$-th stage of DUN (below we expand on the PGD algorithm as an example).
GDM refines the stage image $\mathbf{x}^{(k)} \in \mathbb{R}^{H \times W \times 3}$ based on the previous stage output $\mathbf{x}^{(k-1)}$ and original degraded image $\mathbf{y} \in \mathbb{R}^{H \times W \times 3}$ through the degradation transform $\Phi(\cdot)$. 
Then PMM (or the denoiser) maps the output of GDM into the next stage by solving a denoising subproblem with Gaussian noise level $\sqrt{\rho\lambda}$.
By handling the following two issues, LoRun aims to fully leverage PMM's potential for efficient IR.

Firstly, current DUNs mostly have independently updated blocks which are linked sequentially, which neglects both consistency and hierarchy of denoising capabilities between stages, delivering unstable and oscillating stage recovery results.
Secondly, numerous independent blocks bring redundant parameters, leading to high memory, parameter and computation costs.
To tackle the above problems, inspired by the structurally identical properties of the blocks and the intrinsic rank of weights, our LoRun framework first constructs multi-stage information interoperability at the parameter level, and adds lightweight components to empower different stages with flexible adaptation capabilities.
Then LoRun divides the DUN training procedure into two parts: denoising backbone training and LoRA multi-level fine-tuning.
Fig. \ref{fig:LoRun_framework_clarify} clarifies the differences between our LoRun framework and current unfolding networks.

\noindent \textbf{Overview}
Here, we give the details of our LoRun framework and that how to train a $\mathbf{K}$-stage DUN model.
Fig. \ref{fig:main} shows the main structure of our LoRun framework.
Fig. \ref{fig:main}(a) and (b) denote the two parts of the proposed model, the GDM and the PMM denoiser, and Fig. \ref{fig:main}(c) shows the whole LoRun training strategy indicated by serial numbers.
For the structure of our LoRun, unlike previous DUNs which just stack $\mathbf{K}$ blocks and conduct end-to-end training, we design only one backbone PMM denoiser for all stages and $\mathbf{K}$ lightweight LoRA modules with low-rank weights to adapt to different stages.
And at the $\mathbf{k}$-th stage, we add the weights of the backbone and the $\mathbf{k}$-th LoRA, then we process the forward calculation.

{For the LoRun training procedure, we first establish the foundation by training the DUN with only one single block across stages to obtain the initial pre-trained PMM denoiser instance.
Critically, we ensure that the pre-trained denoiser and the subsequent LoRA fine-tuning are conducted under the identical task setting.}
This fundamental pretrained denoiser is subsequently leveraged and frozen across all $\mathbf{K}$ blocks of our LoRun framework, as shown by serial number 1 in Fig. \ref{fig:main}(c).
While maintaining the original model structure, LoRA decomposes the original weights into a product of low-rank matrices and is integrated into each block, applying dynamic adaptation under the guidance of the backbone denoiser with few independent parameters, illustrated in Fig. \ref{fig:main}(c), serial number 2.
Finally, as indicated by serial number 3, we fine-tune LoRun with most frozen parameters and small learnable LoRA modules end-to-end.
Thus most parameters in the weight matrix of the denoisers are frozen and shared.
We take PGD algorithm for example and provide the detailed description of our LoRun training pipeline in Algorithm \ref{alg:lorun_algorithmic}.

Based on the unique architecture and training strategy, LoRun leverages the low-rank prior of structurally similar modules, enabling more focused and multi-level denoising abilities with less parameters.
Also, decoupling independently updated denoisers into one backbone and adapters, our LoRun can seamlessly switch application tasks based on the same backbone by switching only the fine-tuning modules.
Therefore, LoRun improves DUN from the basic framework and training strategy, which does not rely on specific optimization iteration algorithms, denoiser structures and applications, and is a generic model-free and task-free DUN framework.

\begin{algorithm}[h]
    \caption{Training Pipeline of Our LoRun Algorithm}
    \label{alg:lorun_algorithmic}
    \begin{algorithmic}[1]
        \STATE {\bfseries Input:} The number of stages $\mathbf{K}$, the low-quality image $\mathbf{y}$, the high-quality image $\hat{\mathbf{y}}$, learnable degradation transform $\Phi(\cdot)$, learnable hyperparameters $\rho,\lambda$, \textbf{one} learnable PMM $\mathop{denoiser}(\cdot)$ and $\mathbf{K}$ LoRA modules.
        \STATE {\bfseries Output:} The restored image $\mathbf{x}$.
        \STATE {\bfseries Init:} $\rho,\lambda$ are initialized to $0.5$ and the other deep modules are initialized to the default settings of PyTorch.
        \STATE {\bfseries (1) Backbone Training:}
            \FOR{$k = 1$ {\bfseries to} $\mathbf{K}$}
                \STATE Update intermediate variable $\mathbf{z}^{(k)}$ via Eq. \ref{gradient_descent}.
                \STATE Update restored $\mathbf{x}^{(k)}$ via Eq. \ref{proximal_mapping_simple}.
            \ENDFOR
            \STATE Calculate the loss via Eq. \ref{loss}.
            \STATE Perform backpropagation to update degradation transform $\Phi(\cdot)$, hyperparameters $\rho$, $\lambda$ and only one $\mathop{denoiser}(\cdot)$.
        \STATE {\bfseries (2) LoRA Fine-tuning:}
            \STATE Inject pre-trained $\mathop{denoiser}(\cdot)$ into $\mathbf{K}$ stages, freeze its gradient update and assemble LoRA to each stage.
            \FOR{$k = 1$ {\bfseries to} $\mathbf{K}$}
                \STATE Update intermediate variable $\mathbf{z}^{(k)}$ via Eq. \ref{gradient_descent}.
                \STATE Update restored $\mathbf{x}^{(k)}$ via Eq. \ref{proximal_mapping_simple}.
            \ENDFOR
            \STATE Calculate the loss via Eq. \ref{loss}.
            \STATE Perform backpropagation to update degradation transform $\Phi(\cdot)$, hyperparameters $\rho$, $\lambda$ and $\mathbf{K}$ LoRA modules.
        \STATE {\bfseries Return} $\mathbf{x}$
    \end{algorithmic}
\end{algorithm}

\subsection{Low-Rank Adaptation}
\label{lora_details}
Given the pretrained weight matrix $\mathbf{W}_0 \in \mathbb{R}^{n1 \times n2}$, LoRA freezes the weight and decomposes its update weight matrix $\Delta\mathbf{W}$ into the product of two small factor matrices, $\mathbf{A} \in \mathbb{R}^{n1 \times r}$ and $\mathbf{B} \in \mathbb{R}^{r \times n2}$, where $r \ll \min\{n1, n2\}$.
Hyperparameter $r$ controls the ``intrinsic rank'' and trainable parameters of the update weight $\Delta\mathbf{W}$.
Fig. \ref{fig:lora} shows the details of the LoRA structure.
For input $\mathbf{x}$, the relevant formulas of the fine-tuning process are shown below:
\begin{equation}
\label{lora}
    \mathbf{y} = (\mathbf{W}_0 + \Delta \mathbf{W})\mathbf{x} = (\mathbf{W}_0 + \mathbf{A}\mathbf{B})\mathbf{x},
\end{equation}
where $\mathbf{B}$ is initialized with random Gaussian distribution and zero for $\mathbf{A}$, thus the starting state of $\Delta\mathbf{W}$ is zero.
For the basic module in deep neural networks, such as the convolutional and linear layers, LoRA can be seamlessly integrated and efficiently applied.
For example, given the weight of the convolutional layer $\mathbf{W}_c \in \mathbb{R}^{C_{\text{out}} \times C_{\text{in}} \times k \times k}$ with kernel size $k$, relevant equations can be formulated as:
\begin{equation}
    \Delta \mathbf{W}_c = \operatorname{reshape}(\mathbf{A}_c \mathbf{B}_c),
\end{equation}
where $\mathbf{A}_c \in  \mathbb{R}^{C_{\text{out}} k \times r k}, \mathbf{B}_c \in \mathbb{R}^{r k \times C_{\text{in}} k}$.
By freezing the core weights of the pre-trained model and updating only the low-rank matrices, LoRA guarantees that the model adapts to the updated multi-level capabilities while retaining the original abilities after fine-tuning. 
In addition, LoRA reduces the amount of parameter updates, which improves training efficiency and reduces computational and storage costs.

\subsection{Deep Unfolding Networks}
\label{dun}

DUNs expand the optimization into a GDM and a PMM (or called denoiser) in an end-to-end manner.
Most optimization algorithms, such as Proximal Gradient Descent (PGD) and Half-Quadratic Splitting (HQS), can be unfolded and applied to DUNs.
Since the specific solutions are different, the following illustrations are provided for the two algorithms.

\noindent\textbf{PGD-based Deep Unfolding Network.}
PGD algorithm is a generalized form of projection and can be used to solve non-differentiable convex optimization problems.
And ISTA, ADMM \etc are special instances of proximal algorithms.
Based on the representation in Eq. \eqref{model} and the optimization model in Eq. \eqref{optimization}, PGD uses the following update steps to iteratively solve the IR problem:
\begin{equation}
\label{gradient_descent}
    \mathbf{z}^{(k)} = \mathbf{x}^{(k-1)} - \rho\Phi^{\mathbf{T}}(\Phi \mathbf{x}^{(k-1)} - \mathbf{y}),
\end{equation}
\begin{equation}
\label{proximal_mapping}
    \mathbf{x}^{(k)} = \arg\min_{\mathbf{x}} \frac{1}{2} \|\mathbf{x} - \mathbf{z}^{(k)}\|_2^2 + \rho\lambda \Psi(\mathbf{x}),
\end{equation}
where $k$ denotes the iteration step index, $\rho$ is the step size and $\Psi(\cdot)$ is a handcrafted or deep prior.
Moreover, Eq. \eqref{proximal_mapping} can be regarded as a denoising process with a Gaussian noise at level $\sqrt{\rho\lambda}$ \cite{chan2016plug,ryu2019plug}, simplified as
\begin{equation}
\label{proximal_mapping_simple}
    \mathbf{x}^{(k)} = \mathop{denoiser}_{\rho,\lambda}(\mathbf{z}^{(k)}),
\end{equation}
which can be adjusted as the prior $\Psi(\cdot)$.
For example, if prior $\Psi(\cdot)$ represents the $\mathscr{l}_1$ norm, the $\mathop{denoiser}(\cdot)$ can be denoted as the soft thresholding function:
\begin{equation}
    \mathop{denoiser}_{\rho,\lambda}(\mathbf{z}^{(k)}) = \text{sign}(\mathbf{z}^{(k)}) \cdot \max\{0, |\mathbf{z}^{(k)}| - \rho\lambda\}.
\end{equation}
And if $\Psi(\cdot)$ is deep prior, the $\mathop{denoiser}(\cdot)$ can be substituted for a deep image recovery module, which is learnable and flexible.
Because the DUN models are trained in an end-to-end manner, the degradation matrix $\Phi$ and the hyperparameters ($\rho, \lambda$) can be simply set as learnable parameters and updated with gradient descent-based algorithms.
Thus, in the iterative framework of DUN models, Eq. \eqref{proximal_mapping} or \eqref{proximal_mapping_simple} is a hierarchical denoising process, which means that each block denoises the input $\mathbf{z}^{(k)}$ with independent noise level $\sqrt{\rho\lambda}$.
Generally, each block of the PGD-based DUN consists of a GDM in Eq. \eqref{gradient_descent} and a prior-guided denoiser in Eq. \eqref{proximal_mapping_simple}.

\noindent\textbf{HQS-based Deep Unfolding Network.}
HQS aims to decouple the fidelity term and the regularization term by introducing an auxiliary variable $\mathbf{w}$ for optimization problems.
According to the IR problem and model given in Eq. \eqref{model} and Eq. \eqref{optimization}, HQS contains the following two sub-problems:
\begin{equation}
\label{hqs_projection_last}
    \mathbf{x}^{(k)} = (\Phi^{\mathbf{T}} \Phi + \mu \mathbf{I})^{-1} (\Phi^{\mathbf{T}}\mathbf{y} + \mu\mathbf{w}^{(k-1)}).
\end{equation}
\begin{equation}
\label{hqs_mapping}
    \mathbf{w}^{(k)} = \arg \min_{\mathbf{w}} \frac{1}{2}\Vert \mathbf{w} - \mathbf{x}^{(k)} \Vert_2^2 + \frac{\lambda}{\mu} \Psi(\mathbf{w}).
\end{equation}
The denoising process in Eq. \eqref{hqs_mapping} in HQS is similar to that in Eq. \eqref{proximal_mapping}, so this process can also be treated as a $\mathop{denoiser}(\cdot)$ operator like Eq. \eqref{proximal_mapping_simple}, which is:
\begin{equation}
\label{proximal_mapping_simple_hqs}
    \mathbf{w}^{(k)} = \mathop{denoiser}_{\mu,\lambda}(\mathbf{x}^{(k)}).
\end{equation}

\vspace{-2mm}
\section{Applications and Experiments}
\label{applications_and_experiments}
To investigate the effectiveness and generalization of our proposed LoRun framework, we conduct experiments on three typical IR tasks including Coded Aperture Snapshot Spectral Imaging (CASSI), Compressive Sensing (CS) and image Super Resolution (SR).
We validate our LoRun approach in the following two ways.
(1) We refine the SOTA methods with LoRun framework and make comprehensive comparisons, verifying that LoRun can obtain comparable SOTA performance with less parameters, lower memory costs and faster convergence.
(2) We also employ classic different model structures (U-Net and Transformer) and different optimization algorithms (PGD and HQS) for different task settings (such as different blur kernels in SR task), showing that our strategy can be seamlessly applied to most DUN methods and applications with powerful generalizability.
We present the details of the denoiser structure for the three tasks in the supplementary materials.

\noindent\textbf{Overall settings.}
All experiments are implemented in PyTorch and conducted on a NVIDIA GeForce RTX 3090 GPU and 24GB RAM.
Adam \cite{adam} is used to optimize all applications.
For the number $\mathbf{K}$ of blocks, we set it as $9$ to verify the effectiveness when dealing with large blocks.
For the rank $r$, which controls the size and the ``intrinsic dimension'' of $\mathbf{A}$ and $\mathbf{B}$ in Eq. \eqref{lora}, we simply set it as:
\begin{equation}
\label{r_gamma}
    r = \lceil \min\{ \mathop{input\_dim}, \mathop{output\_dim} \} * \gamma / 100 \rceil
\end{equation}
for each weight matrix in the PMM denoiser, where the $\mathop{input\_dim}$ and $\mathop{output\_dim}$ denote the dimension arguments of the convolutions or the linear layers.
Thus we can tune $\gamma$ to explore the intrinsic rank $r$ efficiently.

\noindent\textbf{Metrics.}
For numerical comparison, we use peak signal-to-noise ratio (PSNR) and structural similarity (SSIM) as metrics to evaluate the effectiveness of the model in all the tasks.

\noindent\textbf{Loss Function.}
Given the training batch of images as $\{x_i\}_{i=1}^b$, where $b$ is the number of training images, we employ the $l_2$ loss to measure the difference between the restored image $\hat{x}_i$ and the ground truth $x_i$. 
Consequently, the loss function is formulated as follows: 
\begin{equation}
\label{loss}
L(\Theta) = \frac{1}{b} \sum_{i=1}^{b} \| \hat{x}_i - x_i \|_2^2,
\end{equation}
where $\Theta$ represents the learnable parameters for each model.

\noindent\textbf{Comparison Settings.}
Given a DUN model with $\mathbf{K}$ blocks, comparisons covers SOTA methods, DUN model with multiple independent blocks (Block-$\mathbf{K}$) and the LoRun model (LoRun-$\mathbf{K}$).
We provide the details of adopted model structures for each application in the supplementary materials.
For table settings, \textbf{best} and \underline{second-best} results are highlighted in bold and underlined, respectively.
For figures, we zoom in on detailed regions with the red box for clarity.

\subsection{Compressive Sensing}

\noindent\textbf{Background.}
CS aims to reconstruct original signal $\mathbf{x} \in \mathbb{R}^{m}$ from compressed measurement $\mathbf{y} \in \mathbb{R}^{n}$, where $n \ll m$.
Based on Eq. \eqref{model} and Eq. \eqref{optimization}, $\Phi \in \mathbb{R}^{n \times m}$ in CS is a linear random projection, which can also be referred to as a sampling matrix.
The CS ratio $\beta$ is defined as $\frac{n}{m}$.
Based on the PGD algorithm, we unfold the model according to Eq. \eqref{gradient_descent} to Eq. \eqref{proximal_mapping_simple}.

\noindent\textbf{Experiment Settings and Datasets.}
In the experiment, we set the sampling matrix $\Phi$ learnable and jointly optimize the sampling matrix with the whole model following \cite{zhang2020amp, mou2022deep}.
We randomly select 40,000 images from the \emph{COCO2017} unlabeled images dataset \cite{coco} for training and evaluate our model on \emph{General100}, \emph{Set11} \cite{Set11} and \emph{Set14} for the set of CS ratios $\beta \in\{ 1\%, 4\%, 10\%, 25\%\}$.

\begin{figure*}
	\footnotesize
	\setlength{\tabcolsep}{1pt}
 \newcommand{\tabincell}[2]{\begin{tabular}{@{}#1@{}}#2\end{tabular}}
	\begin{center}
            \scalebox{1}{
		\begin{tabular}{cccccccc}
		\includegraphics[width=0.12\linewidth]{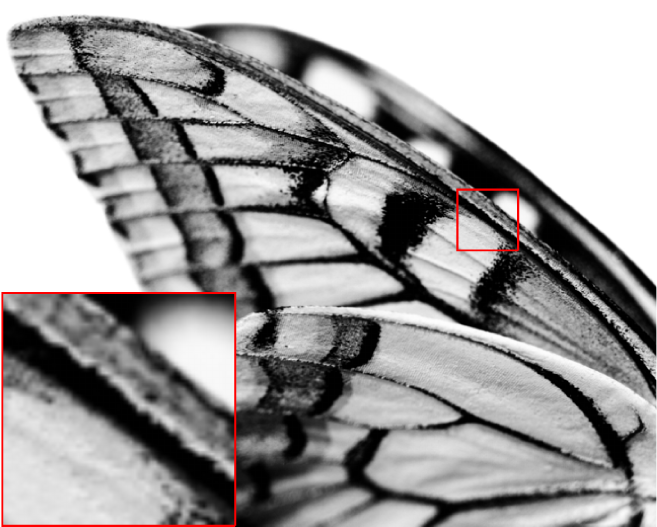}&
            \includegraphics[width=0.12\linewidth]{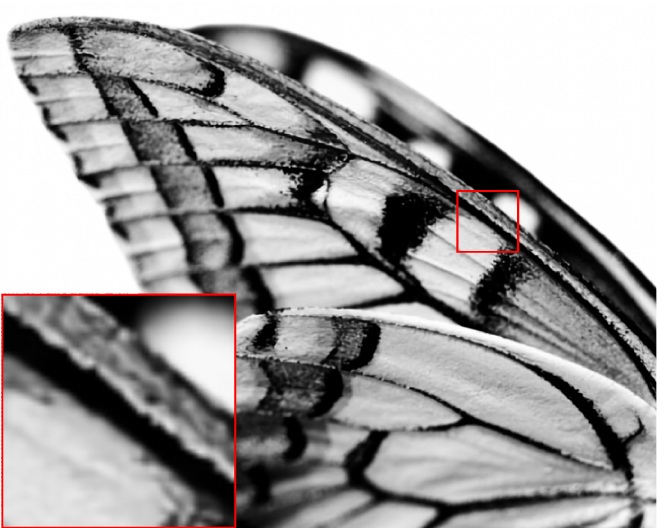}&
            \includegraphics[width=0.12\linewidth]{fig/cs/sota_cs/im_14_CASNet1.pdf}&
            \includegraphics[width=0.12\linewidth]{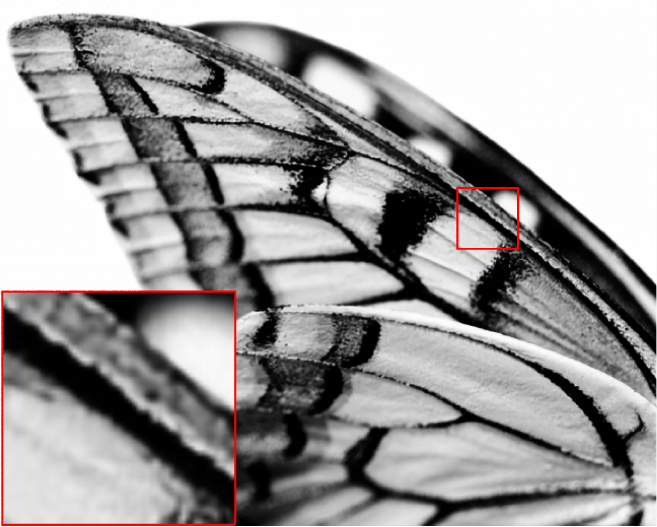}&
            \includegraphics[width=0.12\linewidth]{fig/cs/sota_cs/im_14_DGUNet1.pdf}&
            \includegraphics[width=0.12\linewidth]{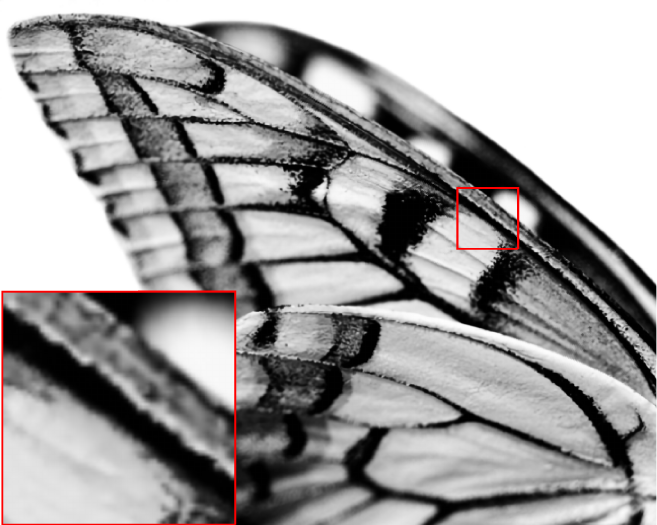}&
            \includegraphics[width=0.12\linewidth]{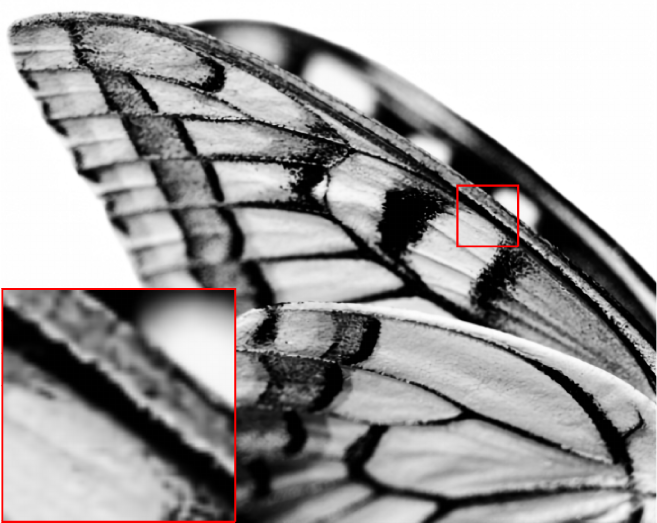}&
            \includegraphics[width=0.12\linewidth]{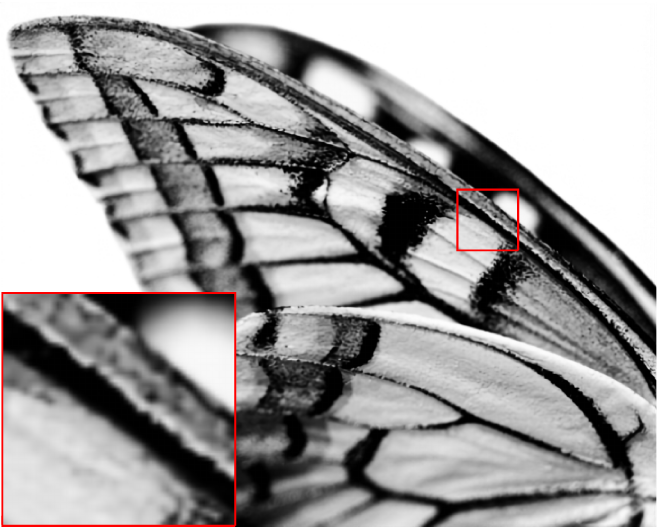}\\
            \includegraphics[width=0.12\linewidth]{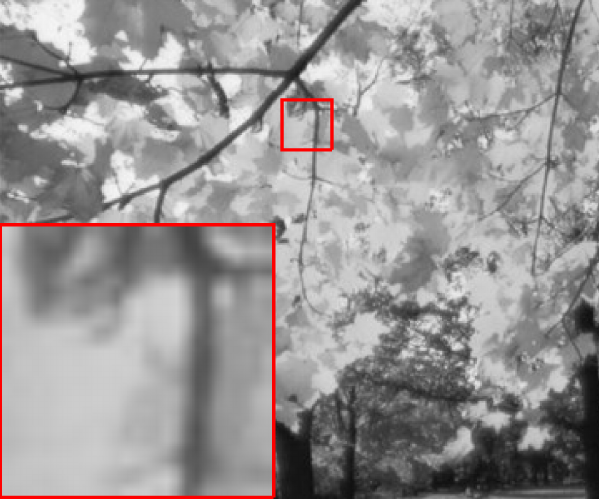}&
            \includegraphics[width=0.12\linewidth]{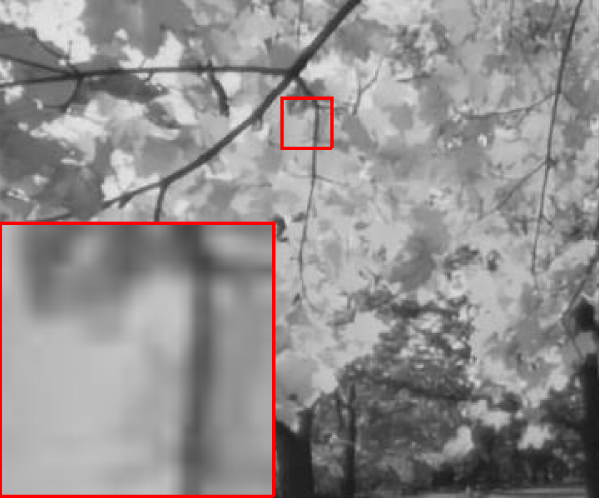}&
            \includegraphics[width=0.12\linewidth]{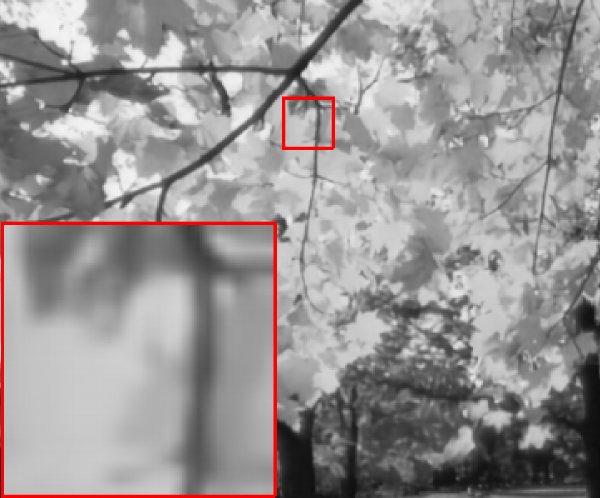}&
            \includegraphics[width=0.12\linewidth]{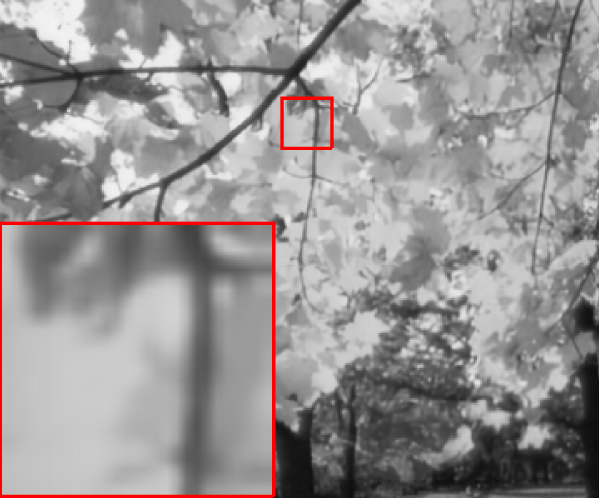}&
            \includegraphics[width=0.12\linewidth]{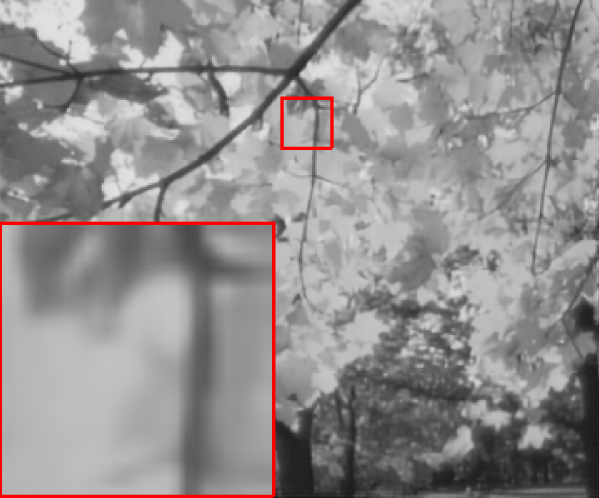}&
            \includegraphics[width=0.12\linewidth]{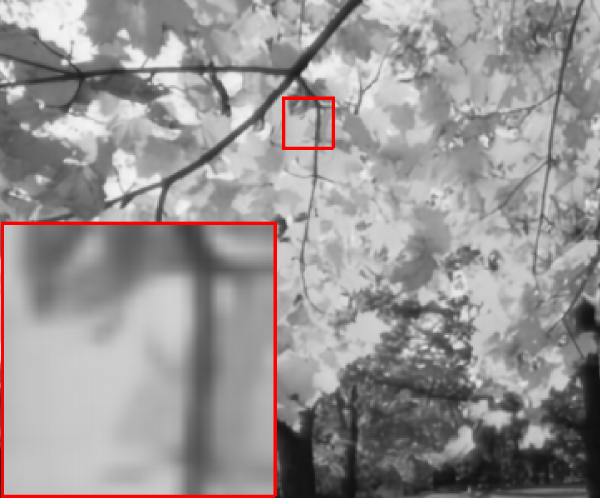}&
            \includegraphics[width=0.12\linewidth]{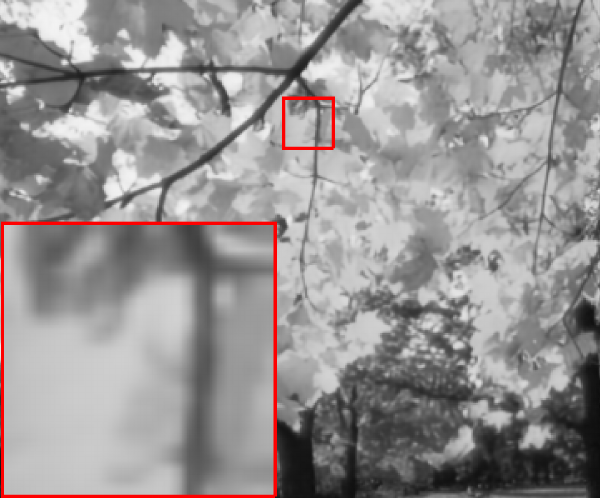}&
            \includegraphics[width=0.12\linewidth]{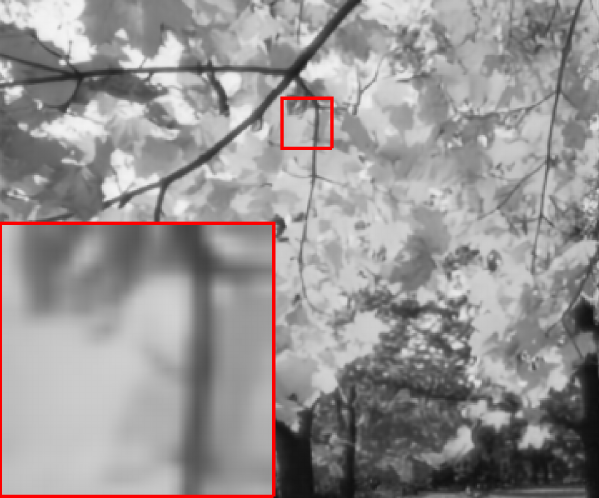}\\
            \hdashline[2pt/1pt]
        \noalign{\vskip 3.5pt}
            \includegraphics[width=0.12\linewidth]{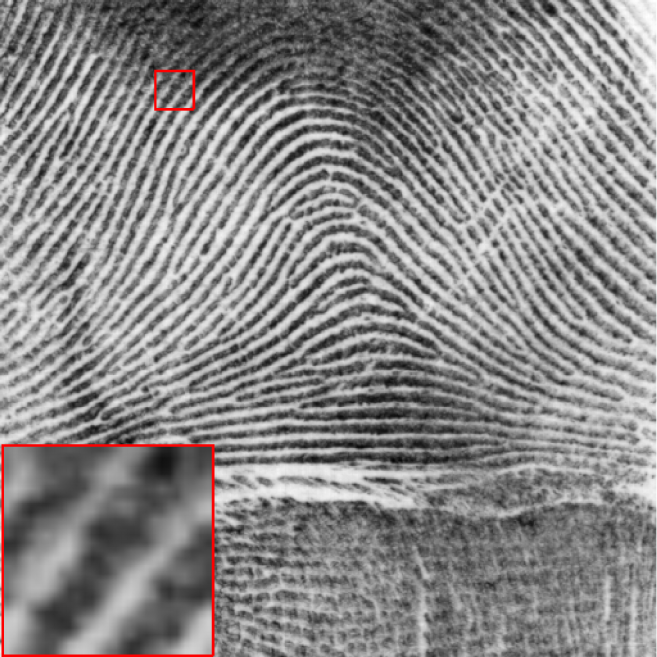}&
            \includegraphics[width=0.12\linewidth]{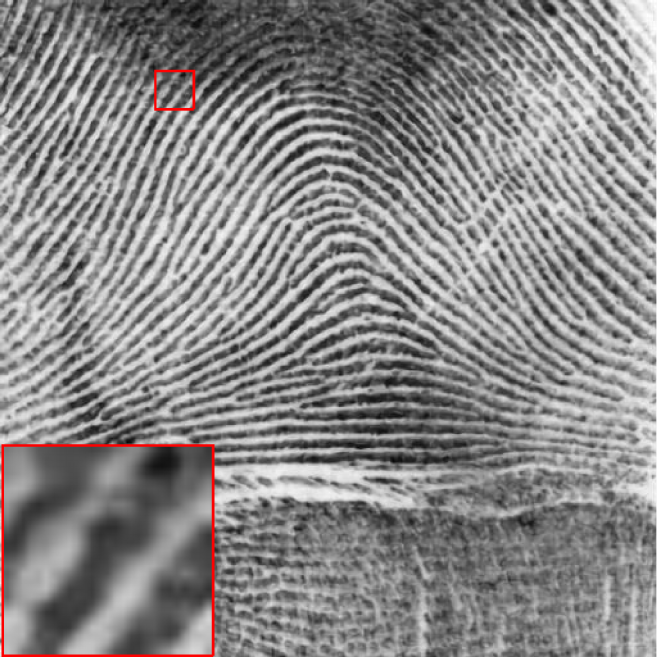}&
            \includegraphics[width=0.12\linewidth]{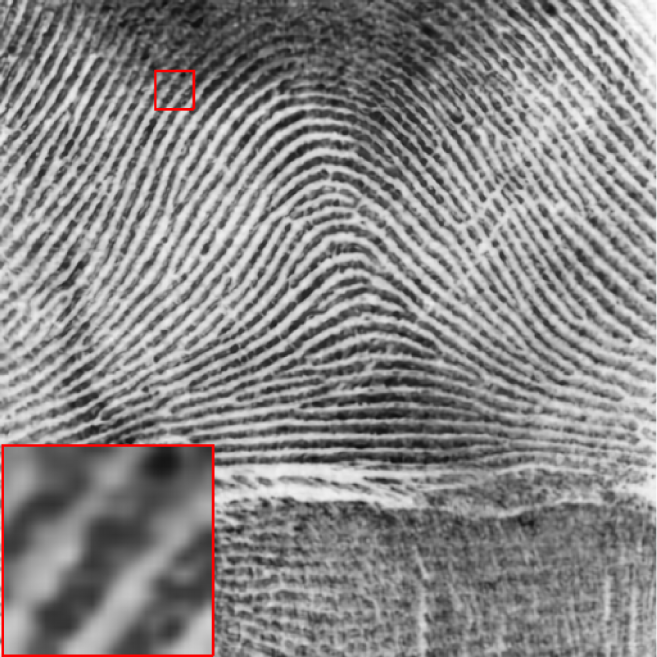}&
            \includegraphics[width=0.12\linewidth]{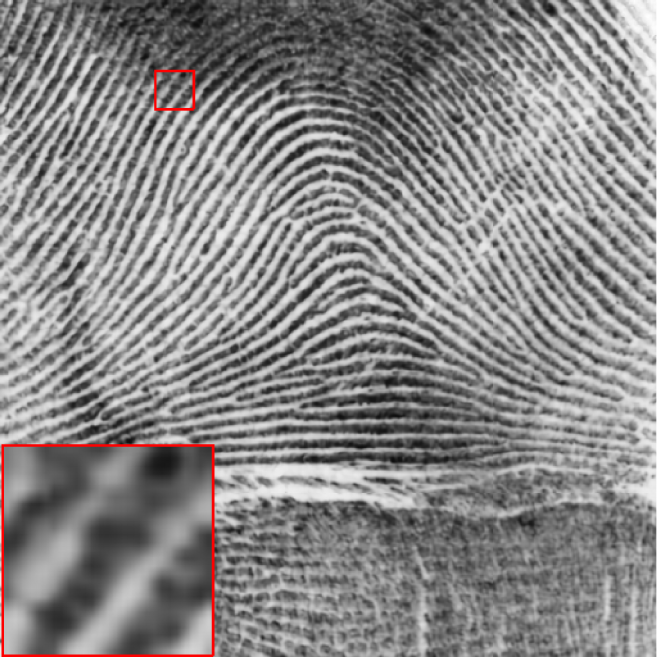}&
            \includegraphics[width=0.12\linewidth]{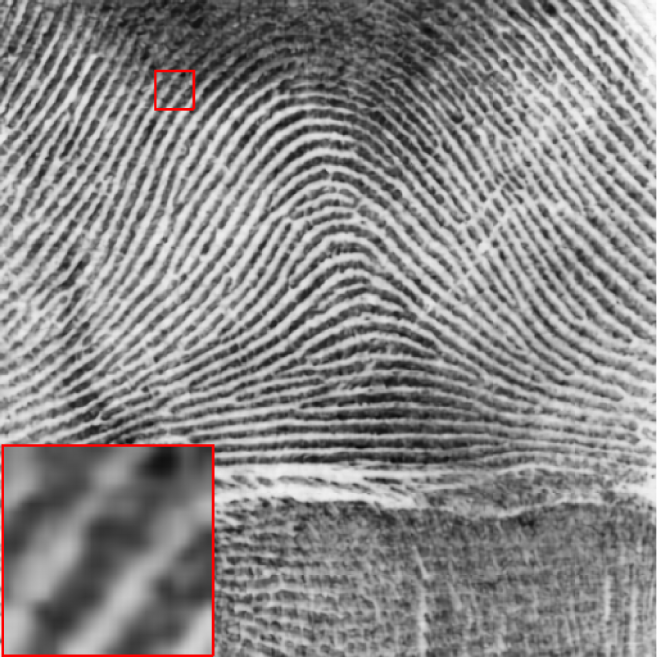}&
            \includegraphics[width=0.12\linewidth]{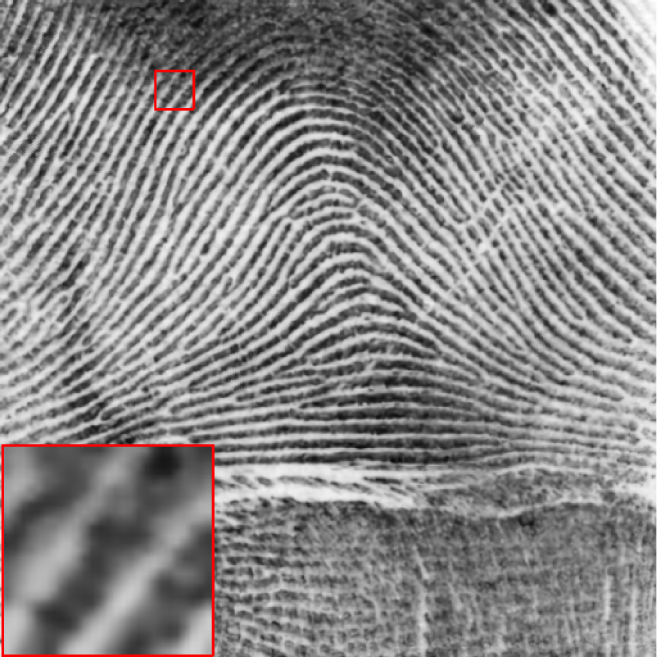}&
            \includegraphics[width=0.12\linewidth]{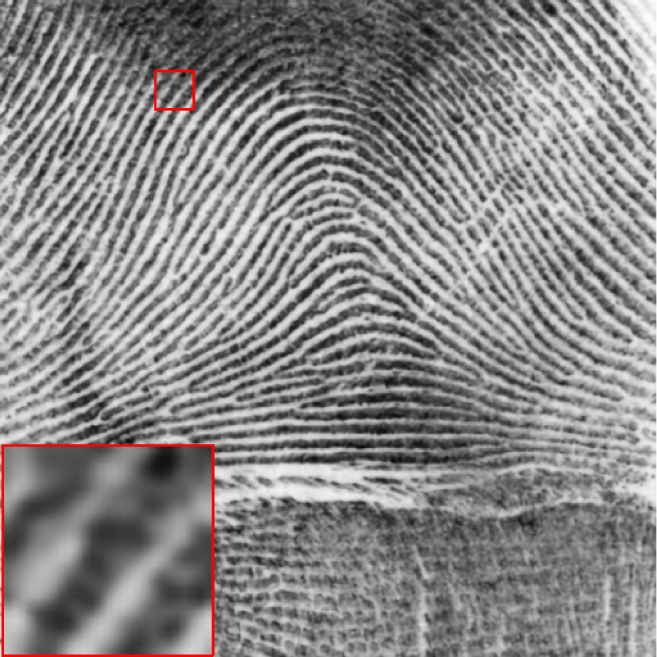}&
            \includegraphics[width=0.12\linewidth]{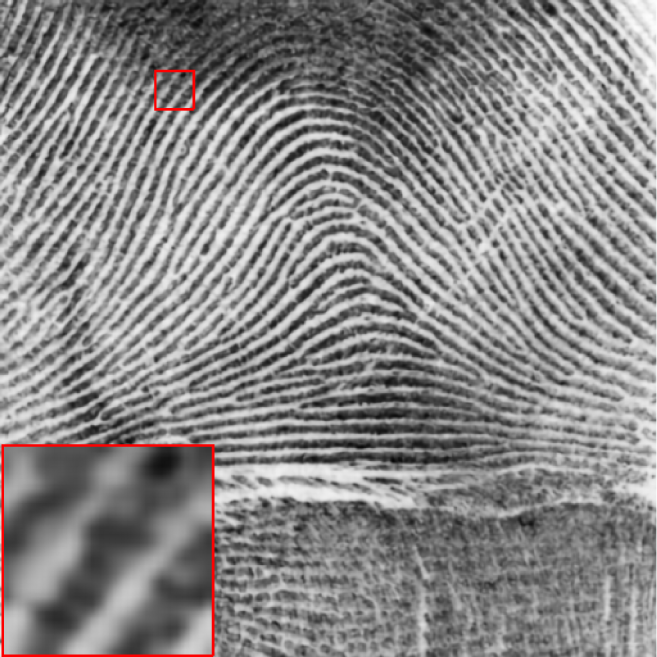}\\
            \includegraphics[width=0.12\linewidth]{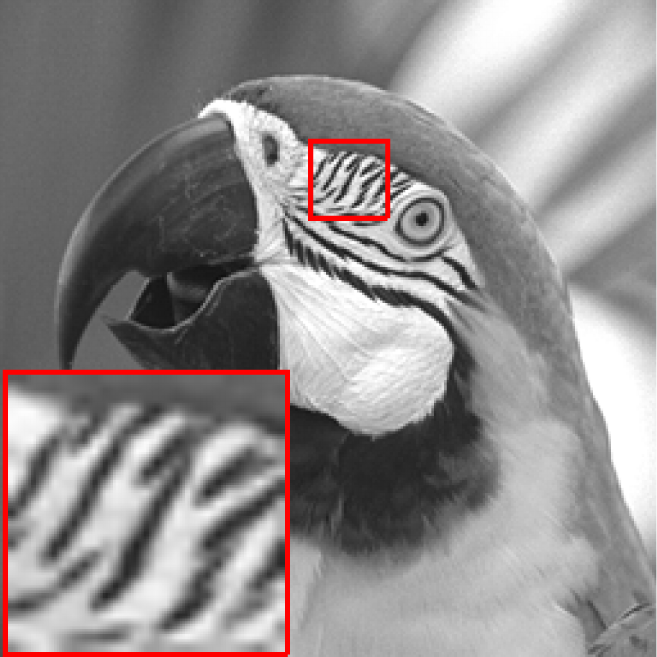}&
            \includegraphics[width=0.12\linewidth]{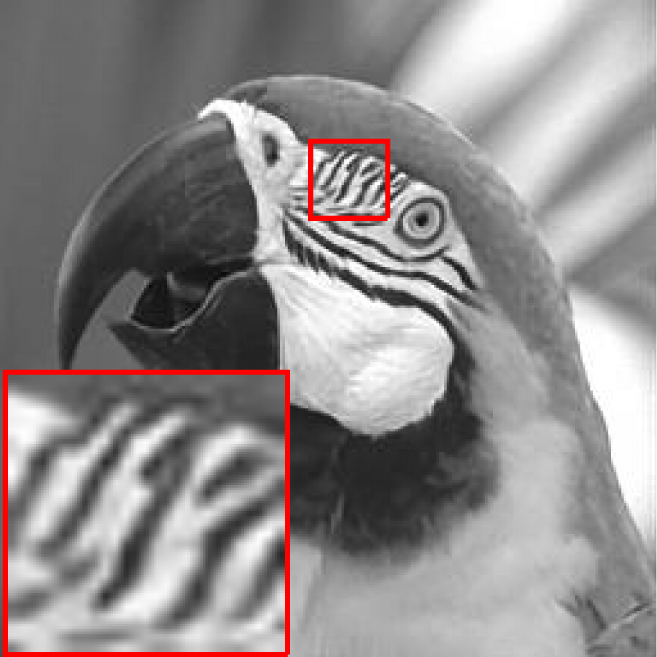}&
            \includegraphics[width=0.12\linewidth]{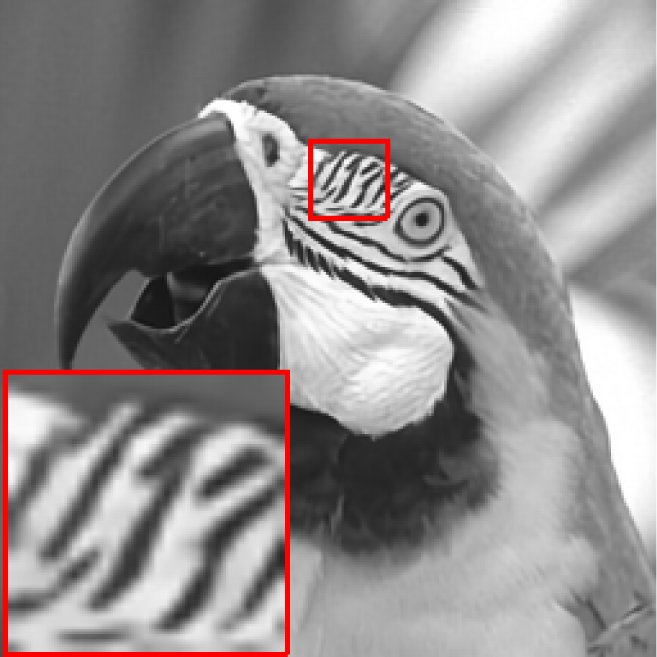}&
            \includegraphics[width=0.12\linewidth]{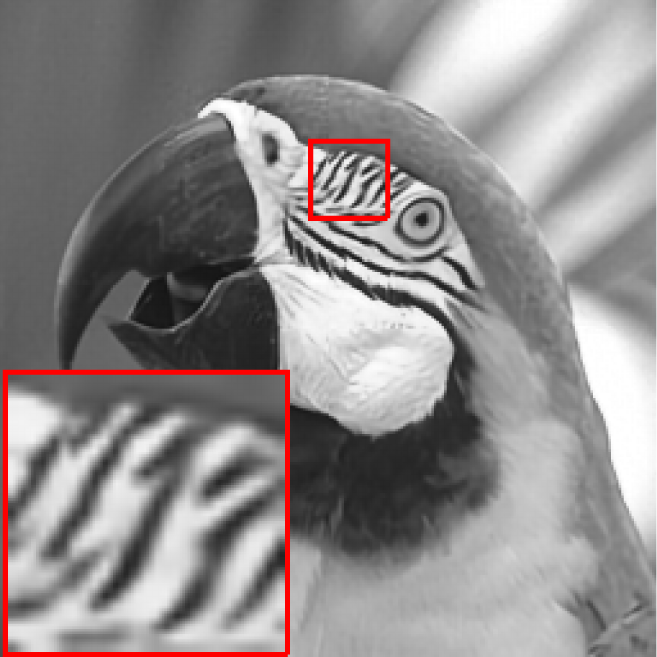}&
            \includegraphics[width=0.12\linewidth]{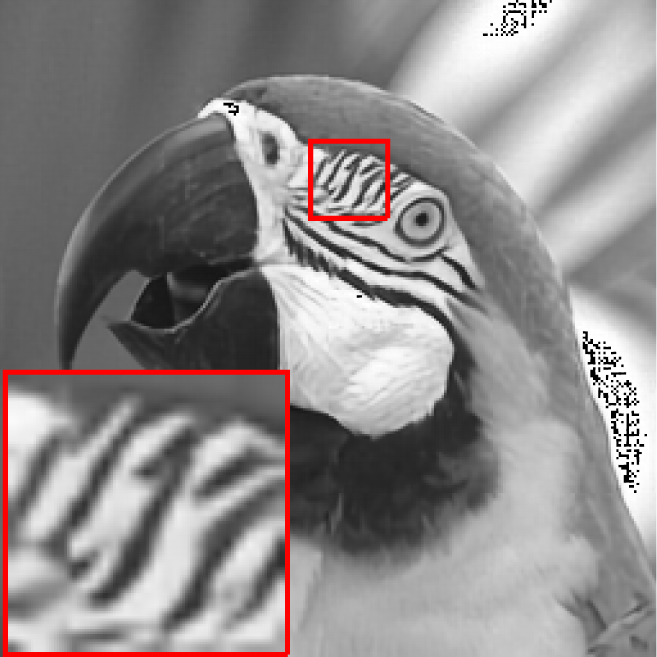}&
            \includegraphics[width=0.12\linewidth]{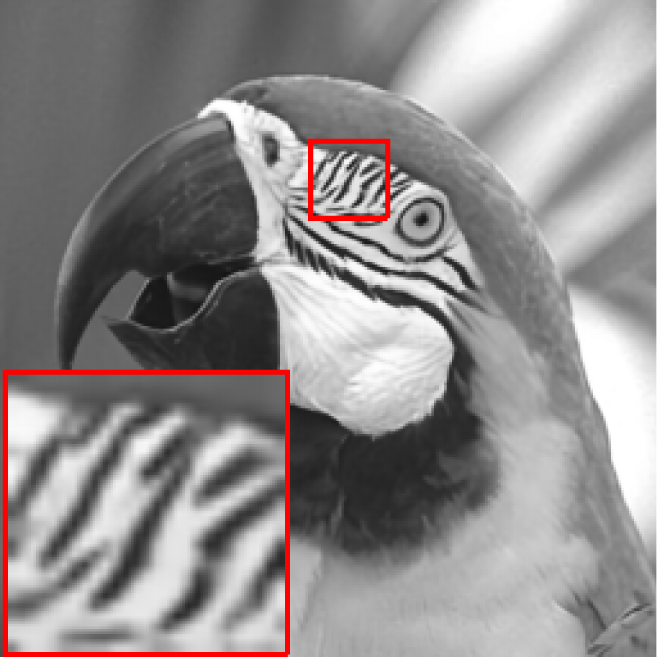}&
            \includegraphics[width=0.12\linewidth]{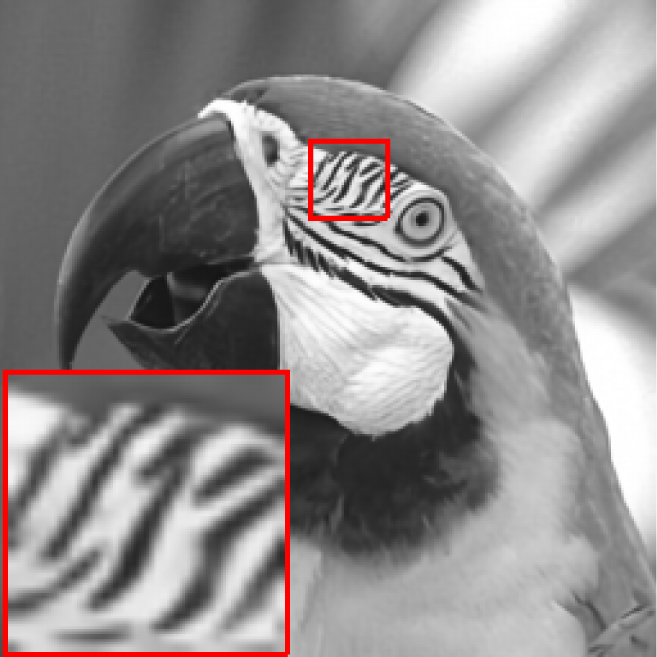}&
            \includegraphics[width=0.12\linewidth]{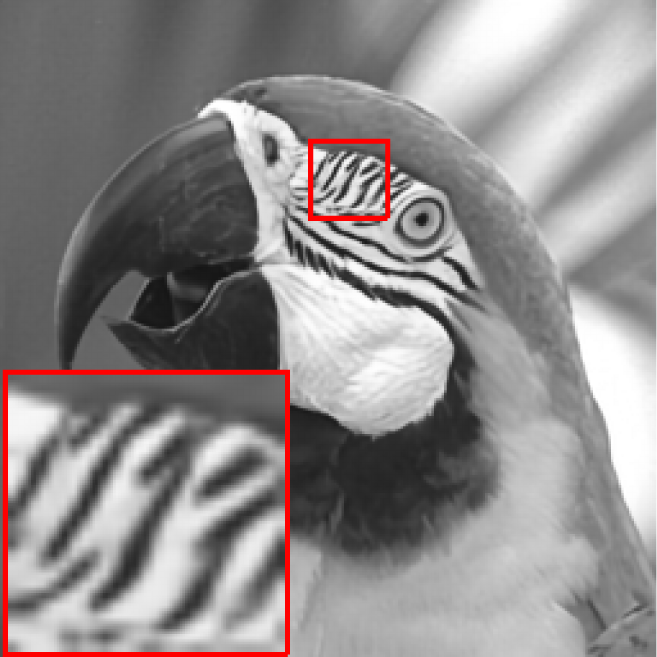}\\
            \hdashline[2pt/1pt]
        \noalign{\vskip 3.5pt}
            \includegraphics[width=0.12\linewidth]{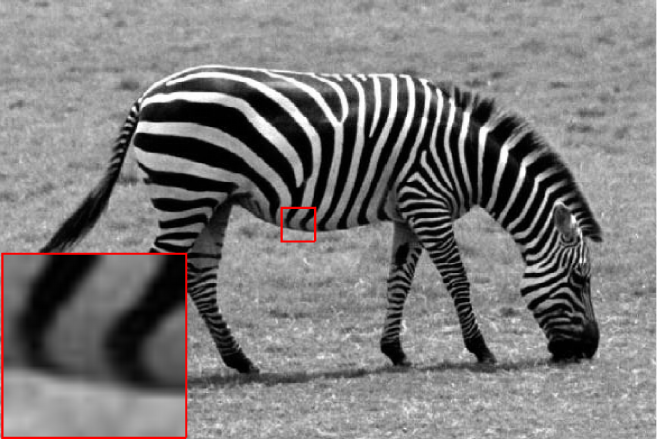}&
            \includegraphics[width=0.12\linewidth]{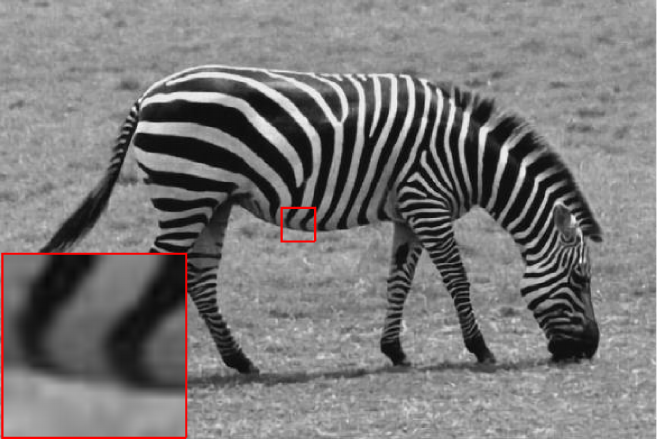}&
            \includegraphics[width=0.12\linewidth]{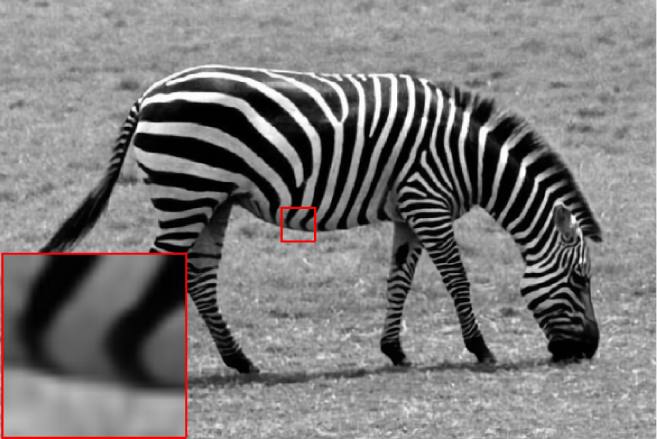}&
            \includegraphics[width=0.12\linewidth]{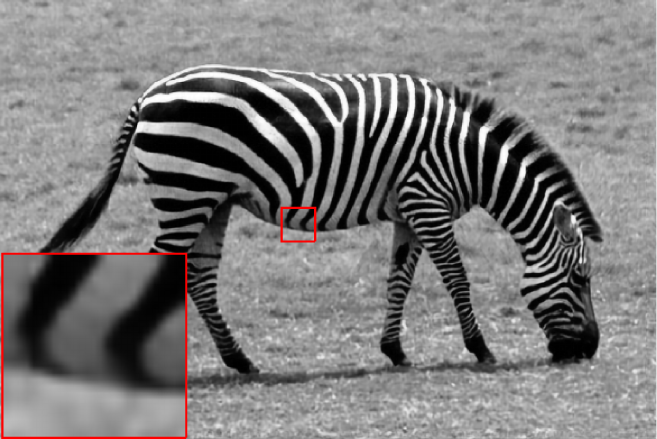}&
            \includegraphics[width=0.12\linewidth]{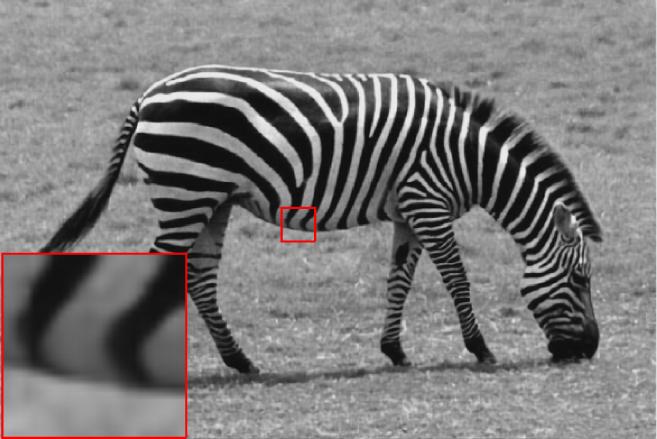}&
            \includegraphics[width=0.12\linewidth]{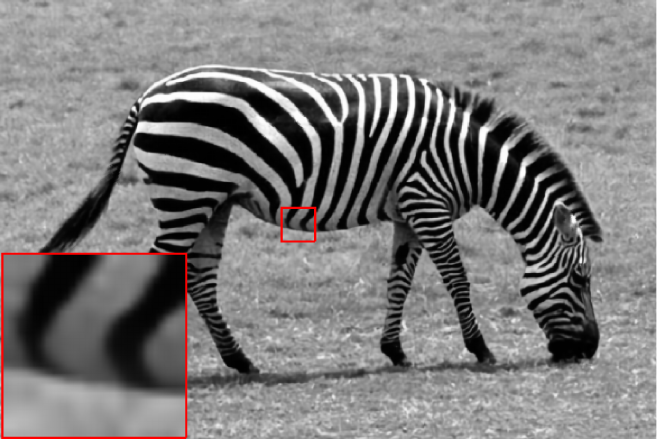}&
            \includegraphics[width=0.12\linewidth]{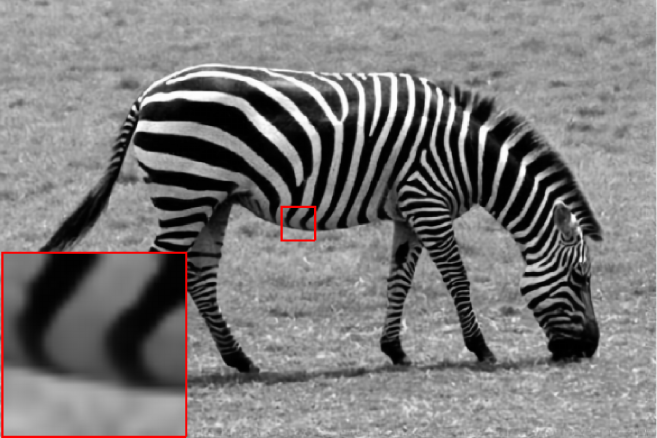}&
            \includegraphics[width=0.12\linewidth]{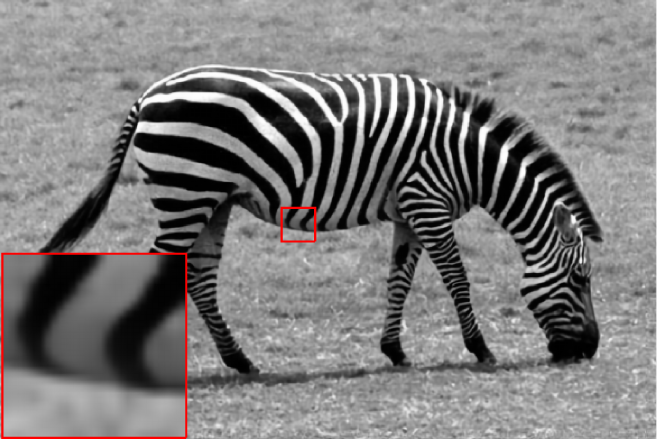}\\

            Truth&AMP-Net&CASNet&DGUNet+&CSformer&NesTD-Net&Block-9&LoRun-9\\
		\end{tabular}} 
		\caption{CS results under ratio $\beta=25\%$ of different methods and datasets, from top to bottom \emph{General100}, \emph{Set11} and \emph{Set14}.}
		\label{cs_sota_figs}
        \vspace{-4mm}
	\end{center}
\end{figure*}

\noindent\textbf{Comparisons and Results.}
We compare our method with deep unfolding networks AMP-Net\cite{zhang2020amp}, CASNet\cite{chen2022content}, DGU-Net+\cite{mou2022deep}, TransCS\cite{shen2022transcs} and NesTD-Net\cite{gan2024nestd}, and two deep learning-based methods AutoBCS\cite{gan2021autobcs} and CSformer \cite{ye2023csformer}.
Table \ref{tab:cs_sota} presents the numeral results of our LoRun and different SOTA methods under CS ratios $\beta \in\{1\%, 4\%, 10\%, 25\%\}$.
For each CS ratio, our LoRun method demonstrates better performance with just $32.8\%$ parameters and no additional inference time compared with Block-9 strategy.
{Also, our LoRun achieves $33\%$ GPU memory reduction compared to Block-9 strategy (Block-9 v.s. LoRun-9: 21G v.s. 7G per GPU).}
LoRun even surpasses the Block-9 strategy on \emph{General100} datasets under $\beta \in\{ 1\%, 4\%\}$ and the average PSNR results.
Compared to previous DUN based methods DGU-Net+ and NesTD-Net, the proposed LoRun method yields PSNR gains of \SI{0.70}{dB} and \SI{0.15}{dB} on the \textit{General100} dataset. 
Concurrently, it led to a decrease in parameter complexity by \SI{1.58}{M} and \SI{0.70}{M}, respectively.
Fig. \ref{cs_sota_figs} shows the CS results of different datasets (\textit{General100}, \textit{Set11} and \textit{Set14}) under CS ratio $\beta=25\%$.
In the zoomed-in detail segment, LoRun obtains neater and more explicit results than SOTA methods and Block-9 strategy.
We also provide the results of the classical denoising modules, Transformer and U-Net in Table \ref{tab:cs_transformer_unet}, showing that LoRun also works on lightweight and basic modules.

\begin{table*}[t]
\centering
\caption{Average PSNR (dB)/SSIM performance comparisons of recent CS methods on various datasets at different CS ratios.}
\vspace{-2mm}
\resizebox{\textwidth}{!}{
\begin{tabular}{ccccccccccc}
\toprule
Datasets & CS Ratios & \textbf{AMP-Net} & \textbf{CASNet }& \textbf{DGU-Net$^+$} & \textbf{TransCS} & \textbf{AutoBCS} & \textbf{CSformer} & \textbf{NesTD-Net} & \textbf{Block-9} & \textbf{LoRun-9}\\
& & TIP'21 & TIP'22 & CVPR'22 & TIP'22 & TCYB'23 & TIP'23 & TIP'24 & Ours & Ours\\
- & - & 0.94M & 16.90M & 6.81M & 1.49M & 2.01M & 6.65M & 5.93M & 15.96M & 5.23M\\
\midrule
\multirow{4}{*}{General100} 
& 0.01 & 22.71/\underline{0.68} & 23.48/0.65 & 22.86/0.62 & 21.66/0.54 & 22.24/0.62 & 23.35/\textbf{0.69} & 23.14/0.62 & \underline{23.89}/0.65 & \textbf{23.92}/0.65 \\
& 0.04 & 26.96/0.77 & 28.50/\textbf{0.82} & 27.92/\underline{0.81} & 27.25/0.74 & 27.91/0.80 & \underline{28.61}/0.80 & 28.58/\textbf{0.82} & 28.60/\textbf{0.82} & \textbf{28.70}/\textbf{0.82} \\
& 0.10 & 30.62/0.88 & 32.78/\textbf{0.91} & 32.74/\textbf{0.91} & 32.17/0.87 & 32.70/\underline{0.89} & \textbf{32.97}/0.88 & \underline{32.85}/\textbf{0.91} & 32.82/\textbf{0.91} & 32.78/\textbf{0.91} \\
& 0.25 & 36.01/0.95 & 38.07/\textbf{0.97} & 37.55/\underline{0.96} & 36.95/\underline{0.96} & 36.71/0.96 & 36.51/\underline{0.96} &  \textbf{38.42}/\textbf{0.97} & \underline{38.17}/\underline{0.96} & 38.16/\textbf{0.97} \\
& Avg. & 29.13/0.81 & 30.71/\textbf{0.84} & 30.19/0.82 & 29.35/0.79 & 29.10/0.82 & 29.82/0.82 & 30.74/\underline{0.83} & \underline{30.87}/\textbf{0.84} &  \textbf{30.89}/\textbf{0.84} \\
\midrule
\multirow{4}{*}{Set11} 
& 0.01 & 20.20/0.56 & 21.76/0.60 & \textbf{22.15}/\underline{0.61} & 20.15/0.50 & 19.63/0.51 & 21.86/\underline{0.61} & 21.40/0.59 & 21.85/ \textbf{0.62} &  \underline{21.89}/\textbf{0.62} \\
& 0.04 & 24.29/0.71 & 26.50/\underline{0.81} & \underline{26.82}/\textbf{0.82} & 23.41/0.71 & 22.45/0.61 & 25.27/0.75 & 26.73/\textbf{0.82} & \textbf{26.83}/\textbf{0.82} &  \textbf{26.83}/\textbf{0.82} \\
& 0.10 & 29.49/0.88 & 30.90/\textbf{0.94} & \underline{30.93}/0.91 & 27.60/0.89 & 26.15/0.80 & 30.09/0.90 & 30.91/0.91 &  \textbf{31.17}/\underline{0.92} & 30.79/0.91 \\
& 0.25 & 34.20/\underline{0.95} & 35.65/\textbf{0.96} & \underline{36.18}/\textbf{0.96} & 32.84/0.93 & 30.95/0.88 & 35.34/0.92 &  \textbf{36.27}/\textbf{0.96} & 36.04/\textbf{0.96} & 35.94/\textbf{0.96} \\
& Avg. & 27.37/0.79 & 28.49/0.81 & \textbf{29.02}/\textbf{0.83} & 25.50/0.76 & 24.30/0.70 & 28.14/0.79 & 28.83/\underline{0.82} &  \underline{28.97}/\textbf{0.83} & 28.86/\textbf{0.83} \\
\midrule
\multirow{4}{*}{Set14}
& 0.01 & 21.37/\underline{0.53} & 22.03/\textbf{0.56} & 21.86/\underline{0.54} & 19.88/0.44 & 21.21/0.50 & 22.27/0.51 & \underline{22.32}/\textbf{0.56} & \underline{22.32}/\textbf{0.56} &  \textbf{22.34}/\textbf{0.56} \\
& 0.04 & 25.50/0.70 & 26.04/\underline{0.73} & 25.88/0.73 & 25.50/0.71 & 25.87/0.72 & 26.04/0.66 & \underline{26.31}/ \textbf{0.74} &  \textbf{26.54}/\textbf{0.74} & 26.26/\textbf{0.74} \\
& 0.10 & 28.77/0.81 & 30.53/\underline{0.85} & 29.34/\textbf{0.85} & 28.33/0.81 & 28.33/0.80 & \underline{30.83}/0.80 & 29.62/\underline{0.85} &  \textbf{31.44}/\textbf{0.87} & 29.53/\underline{0.85} \\
& 0.25 & 33.21/0.91 & 33.95/\textbf{0.93} & 33.70/\textbf{0.93} & 32.14/\underline{0.92} & 29.88/0.91 & 32.45/0.91 &  \textbf{34.33}/\textbf{0.93} & \underline{34.09}/\textbf{0.93} & 34.03/\textbf{0.93} \\
& Avg. & 27.20/0.74 & 27.85/\underline{0.77} & 27.70/0.76 & 26.46/0.72 & 26.82/0.73 & 27.42/0.75 & \underline{28.15}/\underline{0.77} & \textbf{28.60}/\textbf{0.78} & 28.04/\underline{0.77} \\
\bottomrule
\end{tabular}
}
\label{tab:cs_sota}
\vspace{-2mm}
\end{table*}

\begin{table}[!t]
    \centering
    \setlength{\tabcolsep}{3pt}
    \newcommand{\tabincell}[2]{\begin{tabular}{@{}#1@{}}#2\end{tabular}}
    \caption{Comparisons on \emph{Set11} under different CS ratios for \textbf{CS} based on \textbf{Transformer and U-Net} denoiser. PSNR (upper entry), SSIM (lower entry) and parameters are reported.}
    \vspace{-2mm}
    \scalebox{0.9}{
    \begin{tabular}{cccccccc}
    \toprule
        Model & Method & $\beta$=1\% & $\beta$=4\% & $\beta$=10\% & $\beta$=25\% & $\beta$=30\% & Params.\\
    \midrule
        \multirow{2}{*}{Transformer} & Block-9 & 
        \tabincell{c}{20.73\\0.54} &
        \tabincell{c}{24.50\\0.74}&
        \tabincell{c}{28.44\\0.88}&
        \tabincell{c}{33.63\\\textbf{0.95}}&
        \tabincell{c}{34.72\\\textbf{0.95}}&
        \tabincell{c}{0.10M}\\
        &\tabincell{c}{LoRun-9\\$\gamma$=10}&
        \tabincell{c}{\textbf{21.02}\\\textbf{0.55}}&
        \tabincell{c}{\textbf{25.17}\\\textbf{0.75}}&
        \tabincell{c}{\textbf{29.14}\\\textbf{0.89}}&
        \tabincell{c}{\textbf{34.16}\\\textbf{0.95}}&
        \tabincell{c}{\textbf{35.54}\\\textbf{0.95}}&
        \tabincell{c}{0.06M}\\
    \midrule
        \multirow{2}{*}{U-Net} & Block-9 & 
        \tabincell{c}{\textbf{21.14}\\\textbf{0.56}}& 
        \tabincell{c}{\textbf{25.60}\\\textbf{0.79}}& 
        \tabincell{c}{\textbf{29.57}\\\textbf{0.89}}& 
        \tabincell{c}{\textbf{35.00}\\\textbf{0.96}}& 
        \tabincell{c}{\textbf{36.36}\\\textbf{0.96}}&
        \tabincell{c}{4.65M}\\
        &\tabincell{c}{LoRun-9\\$\gamma$=10} & 
        \tabincell{c}{21.13\\\textbf{0.56}}&
        \tabincell{c}{25.45\\\textbf{0.79}}&
        \tabincell{c}{29.29\\\textbf{0.89}}&
        \tabincell{c}{34.75\\0.95}&
        \tabincell{c}{36.17\\\textbf{0.96}}&
        \tabincell{c}{1.42M}\\
    \toprule
    \end{tabular}
    }
    \label{tab:cs_transformer_unet}
    \vspace{-6mm}
\end{table}

\subsection{Coded Aperture Snapshot Spectral Imaging}

\noindent\textbf{Background.}
CASSI, which aims at scanning the scenes with spatial and spectral dimensions, has a hardware-based encoder and a software-based decoder. With a coded aperture and a disperser, it encodes and shifts each band of the original multi-spectral images (MSIs) $\mathcal{X} \in \mathbb{R}^{H \times W \times C}$ with known mask $\mathbf{M} \in \mathbb{R}^{H \times W}$ and later blends all the bands to generate a 2-D measurement $\mathbf{Y} \in \mathbb{R}^{H \times (W + d \times (C - 1))}$, where $d$ is the shift step. Considering the measurement noise $\mathbf{N} \in \mathbb{R}^{H \times (W + d \times (C - 1))}$ generated in the coding system, the whole process can be formulated as
\begin{equation}
    \mathbf{Y} = \sum_{k = 1}^{C} shift(\mathcal{X}(:,:,k) \odot \mathbf{M}) + \mathbf{N}.
\end{equation}
For the decoder process, as Eq. \eqref{model} and Eq. \eqref{optimization} defined, we reformulate the model to determine the specific form: the vectorized measurement $\mathbf{y} \in \mathbb{R}^{n}$ where $n = \mathop{H}(\mathop{W}+\mathop{d}(\mathop{C}-1))$, the vectorized shifted MSI $\mathbf{x} \in \mathbb{R}^{nC}$ and the known degradation matrix $\Phi \in \mathbb{R}^{n \times nC}$.
We adopt the HQS algorithm for deep unfolding procedure (from Eq. \eqref{hqs_projection_last} to Eq. \eqref{proximal_mapping_simple_hqs}).

\noindent\textbf{Experiment Settings and Datasets.}
In the experiment, the \emph{CAVE} dataset \cite{cave}, which contains 32 MSIs with the spatial size of $512 \times 512$, is employed as the training set.
Following the settings of \cite{cai2022degradation}, we select 10 scenes (all of size $256 \times 256 \times 28$) from \emph{KAIST} dataset \cite{kaist} as testing datasets.
And the shift step $d$ is set to 2, which means that the measurement is of size $256 \times 310$ (shown in Fig. \ref{cassi_meas}).

\noindent\textbf{Comparisons and Results.}
We compare our method with one model-based method DeSCI\cite{desci}, two deep learning-based CNN methods $\lambda$-Net\cite{miao2019net} and TSA-Net\cite{meng2020end}, two Plug-and-play (PnP) methods PnP-CASSI \cite{zheng2021deep} and DIP-HSI\cite{meng2021self}, one Recurrent neural network BIRNAT\cite{birnat}, three DUN methods GAP-Net\cite{meng2023deep}, DAUHST\cite{cai2022degradation} and RCUMP\cite{zhao2024rcump}.
Table \ref{tab:sci_sota} demonstrates the performance of LoRun framework compared with SOTA methods on 10 simulation scenes (\emph{scene01}$\sim$\emph{scene10}) in \textit{KAIST} datasets for CASSI.
Fig. \ref{cassi_dbands} shows the visualization of \emph{scene01} in different bands ($\lambda \in \{\SI{481.5}{\nano\meter}, \SI{492.5}{\nano\meter}, \SI{522.5}{\nano\meter}, \SI{567.5}{\nano\meter}, \SI{604.5}{\nano\meter}\}$).
Fig. \ref{cassi_dimgs} presents visual results of five datasets (\emph{scene02}, \emph{scene06}, \emph{scene07}, \emph{scene08} and \emph{scene10}) in different bands ($\lambda \in \{\SI{487.0}{\nano\meter}, \SI{498.0}{\nano\meter}, \SI{584.5}{\nano\meter}, \SI{492.5}{\nano\meter}, \SI{476.5}{\nano\meter}\}$).
Our proposed LoRun strategy not only achieves better performance among all types of the methods, but also obtains comparable results with $31.3\%$ parameters against Block-9 strategy.
And compared to SOTA regular DUN method RCUMP, our LoRun achieves $\SI{1.3}{dB}$ higher PSNR metrics with only $17.9\%$ of the number of its parameters.
From the selected pseudo-color images, our LoRun strategy has yielded a clearer and more informative result with sharper grain boundaries.
Table \ref{tab:snapshot_unet_s1-5} shows the MSI recovery results with the U-Net denoiser which indicate that LoRun can also fit and work seamlessly into simple but classic modules with great generalization.

\begin{table*}
	\caption{Comparisons of PSNR (dB)/SSIM between LoRun ($\gamma = 10$) and SOTA methods on 10 simulation scenes in KAIST for \textbf{CASSI}.}
    \vspace{-2mm}
\def\arraystretch{1.2}
\setlength{\tabcolsep}{3pt}
	\centering
\scalebox{0.85}{
		\begin{tabular}{cccccccccccccc}
			\toprule
			\multicolumn{1}{c}{\textbf{-}} &
			\multicolumn{1}{c}{\textbf{DeSCI}} &
			\multicolumn{1}{c}{\textbf{$\lambda$-Net}} &
			\multicolumn{1}{c}{\textbf{TSA-Net}} &
			\multicolumn{1}{c}{\textbf{PnP-CASSI}} &
			\multicolumn{1}{c}{\textbf{DIP-HSI}} &
			\multicolumn{1}{c}{\textbf{BIRNAT}} &
			\multicolumn{1}{c}{\textbf{GAP-Net}} &
			\multicolumn{1}{c}{\textbf{DAUHST}} &
			\multicolumn{1}{c}{\textbf{DWMT}} &
			\multicolumn{1}{c}{\textbf{PSRSCI}} &
			\multicolumn{1}{c}{\textbf{RCUMP}} &
			\multicolumn{1}{c}{\textbf{Block-9}} &
			\multicolumn{1}{c}{\textbf{LoRun-9}} \\
            \midrule
            \multicolumn{1}{c}{\textbf{Category}} & Model & CNN  & CNN  & PnP & PnP & RNN & DUN  & DUN  & Transformer & Diffusion & DUN  & DUN  & DUN  \\
            - & - & Supervised & Supervised & Zero-Shot & Zero-Shot & - & Supervised & Supervised & Supervised & Supervised & Supervised & Supervised & Supervised \\
            \multicolumn{1}{c}{\textbf{Reference}} & TPAMI'19 & ICCV'19 & ECCV'20 & PR'21 & ICCV'21 & TPAMI'23 & IJCV'23 & NeurIPS'22 & AAAI'24 & ICLR'25 & TIP'24 & Ours & Ours \\
            \multicolumn{1}{c}{\textbf{Params.}} & - & 62.64M & 44.25M & 33.85M & 33.85M & 4.40M & 4.27M & 6.15M & 15.47M & 1312M & 9.03M & 5.18M & 1.62M \\
            \midrule
			\textbf{scene01}
			& 28.38/0.80
			& 30.10/0.85
			& 32.31/0.89
			& 29.09/0.80
			& 31.32/0.86
			& 36.79/0.95
			& 33.63/0.91
			& 37.25/\underline{0.96}
			& 36.46/\underline{0.96}
			& 37.18/\underline{0.96}
			& 37.63/\underline{0.96}
			& \textbf{38.88}/\textbf{0.97}
			& \underline{38.78}/\textbf{0.97} \\

			\textbf{scene02}
			& 26.00/0.70
			& 28.49/0.81
			& 31.03/0.86
			& 28.05/0.71
			& 25.89/0.70
			& 37.89/0.96
			& 33.19/0.90
			& 39.02/\underline{0.97}
			& 37.75/0.96
			& 38.74/\underline{0.97}
			& 39.98/\underline{0.97}
			& \textbf{41.70}/\textbf{0.98}
			& \underline{41.68}/\textbf{0.98} \\

			\textbf{scene03}
			& 23.11/0.73
			& 27.73/0.87
			& 32.15/0.92
			& 30.15/0.85
			& 29.91/0.84
			& 40.61/\underline{0.97}
			& 33.96/0.93
			& 41.05/\underline{0.97}
			& 38.47/\underline{0.97}
			& 41.07/\textbf{0.98}
			& 42.38/\underline{0.97}
			& \textbf{44.29}/\textbf{0.98}
			& \underline{44.11}/\textbf{0.98} \\

			\textbf{scene04}
			& 28.26/0.86
			& 37.01/0.93
			& 37.95/0.96
			& 39.17/0.94
			& 38.69/0.93
			& 46.94/\underline{0.98}
			& 39.14/0.97
			& 46.15/\underline{0.98}
			& 44.23/\underline{0.98}
			& 46.31/\textbf{0.99}
			& 46.94/\underline{0.98}
			& \underline{48.50}/\textbf{0.99}
			& \textbf{48.54}/\textbf{0.99} \\

			\textbf{scene05}
			& 25.41/0.78
			& 26.19/0.82
			& 29.47/0.88
			& 27.45/0.80
			& 27.45/0.80
			& 35.42/0.96
			& 31.44/0.92
			& 35.80/\underline{0.97}
			& 33.99/0.96
			& 35.81/\underline{0.97}
			& \underline{36.67}/\underline{0.97}
			& \textbf{38.16}/\textbf{0.98}
			& \textbf{38.16}/\textbf{0.98} \\

			\textbf{scene06}
			& 22.38/0.68
			& 28.64/0.85
			& 31.44/0.91
			& 26.16/0.75
			& 29.53/0.82
			& 35.30/0.96
			& 32.40/0.93
			& 37.08/\underline{0.97}
			& 36.17/\underline{0.97}
			& 36.76/\underline{0.97}
			& 37.11/\underline{0.97}
			& \textbf{38.22}/\textbf{0.98}
			& \underline{38.16}/\textbf{0.98} \\

			\textbf{scene07}
			& 24.45/0.74
			& 26.47/0.81
			& 30.32/0.88
			& 26.92/0.74
			& 27.46/0.70
			& 36.58/0.96
			& 32.27/0.90
			& 37.57/\underline{0.96}
			& 35.22/0.95
			& 37.38/\textbf{0.97}
			& 37.86/\underline{0.96}
			& \textbf{38.70}/\textbf{0.97}
			& \underline{38.63}/\textbf{0.97} \\

			\textbf{scene08}
			& 22.03/0.67
			& 26.09/0.83
			& 29.35/0.89
			& 24.92/0.71
			& 27.69/0.80
			& 33.96/0.96
			& 30.46/0.91
			& 35.10/\underline{0.97}
			& 34.56/\underline{0.97}
			& 34.55/0.96
			& 35.21/\underline{0.97}
			& \underline{36.16}/\underline{0.97}
			& \textbf{36.18}/\textbf{0.98} \\

			\textbf{scene09}
			& 24.56/0.73
			& 27.50/0.83
			& 30.01/0.89
			& 27.99/0.75
			& 33.46/0.86
			& 39.47/0.97
			& 33.51/0.92
			& 40.02/0.97
			& 37.41/0.97
			& 39.49/0.97
			& 40.94/0.97
			& \textbf{43.12}/\underline{0.98}
			& \underline{42.61}/\textbf{0.99} \\

			\textbf{scene10}
			& 23.59/0.59
			& 27.13/0.82
			& 29.59/0.87
			& 25.58/0.66
			& 26.10/0.73
			& 32.80/0.94
			& 30.24/0.90
			& 34.59/\underline{0.96}
			& 34.00/\underline{0.96}
			& 34.10/\underline{0.96}
			& 34.91/\underline{0.96}
			& \textbf{36.08}/\textbf{0.97}
			& \underline{35.88}/\textbf{0.97} \\

			\textbf{Avg}
			& 24.82/0.73
			& 28.54/0.84
			& 31.36/0.89
			& 28.55/0.77
			& 29.75/0.80
			& 37.58/0.96
			& 33.03/0.92
			& 38.36/\underline{0.97}
			& 36.83/0.96
			& 38.14/\underline{0.97}
			& 38.97/\underline{0.97}
			& \textbf{40.39}/\textbf{0.98}
			& \underline{40.27}/\textbf{0.98} \\
            \toprule
		\end{tabular}
}
	\label{tab:sci_sota}
    \vspace{-3mm}
\end{table*}

\begin{figure*}
	\footnotesize
	\setlength{\tabcolsep}{1pt}
 \newcommand{\tabincell}[2]{\begin{tabular}{@{}#1@{}}#2\end{tabular}}
	\begin{center}
            \scalebox{0.94}{
		\begin{tabular}{cccccccccc}
			\includegraphics[width=0.1\linewidth]{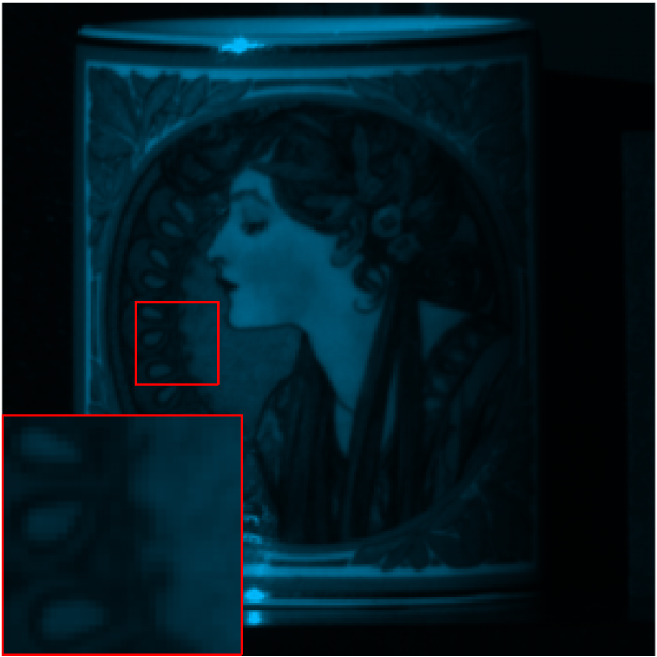}&
			\includegraphics[width=0.1\linewidth]{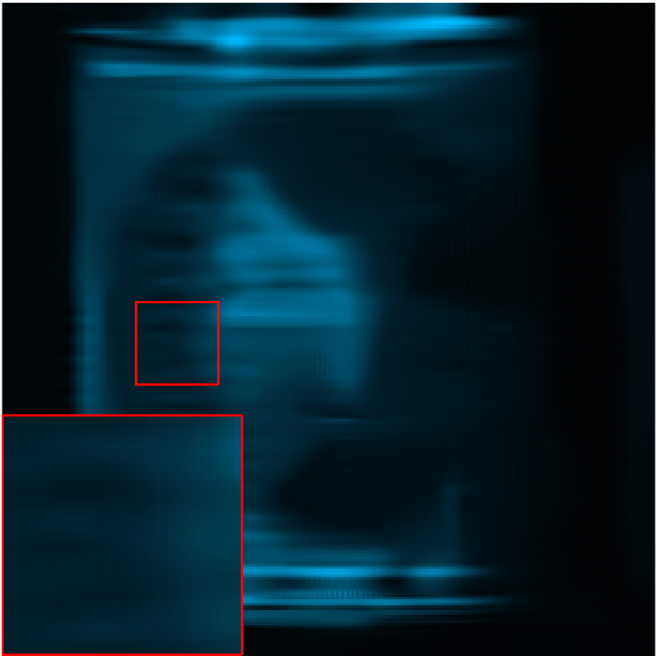}&
			\includegraphics[width=0.1\linewidth]{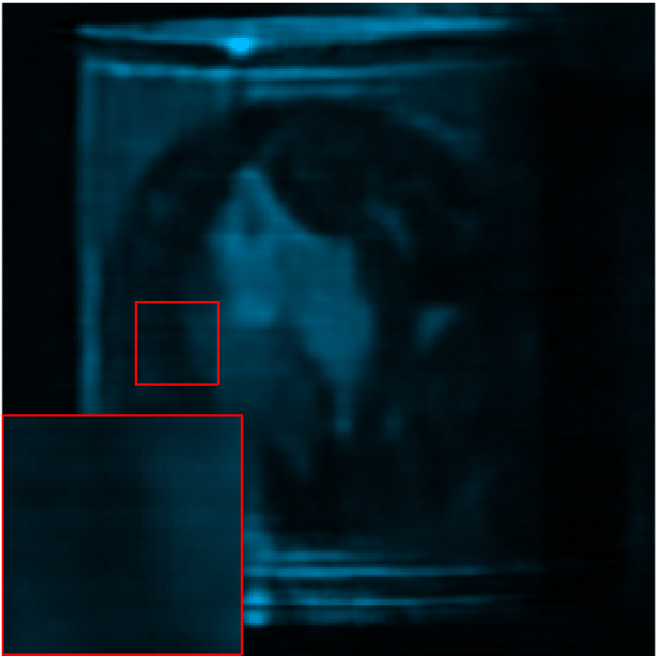}&
			\includegraphics[width=0.1\linewidth]{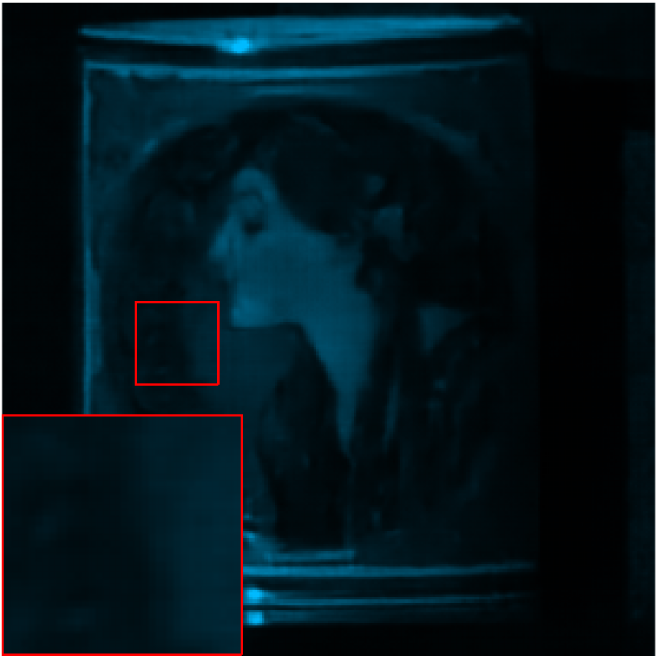}&
			\includegraphics[width=0.1\linewidth]{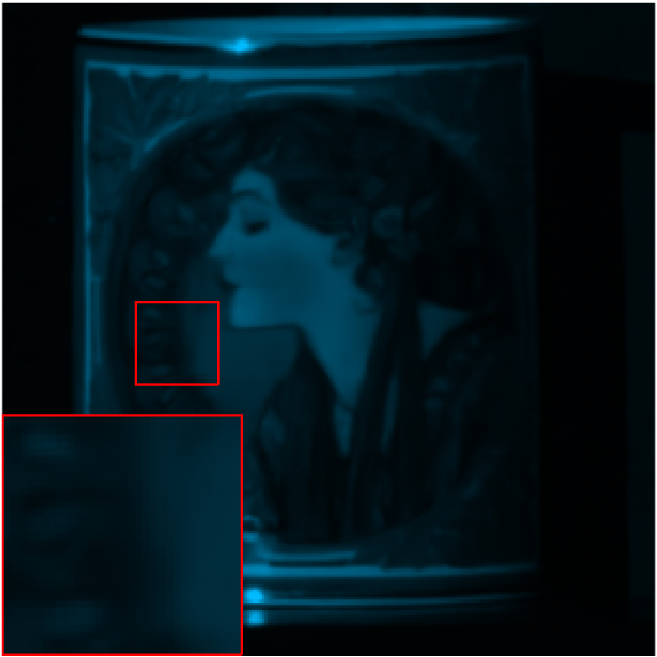}&
                \includegraphics[width=0.1\linewidth]{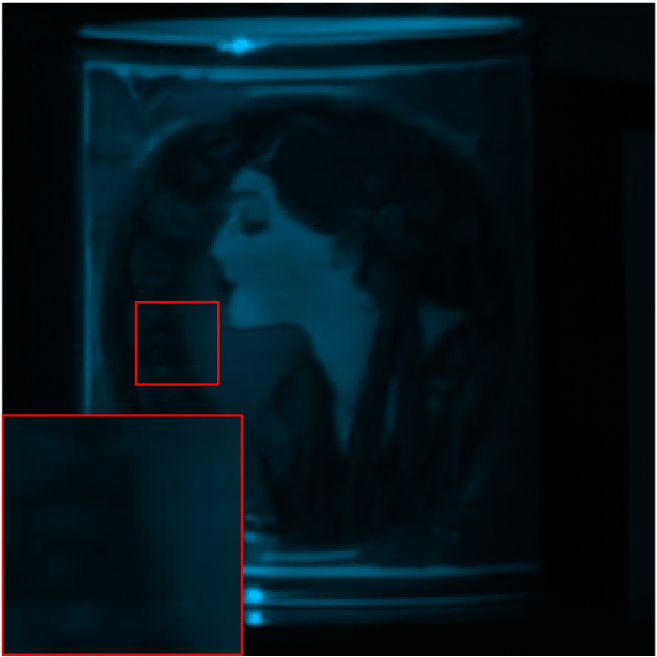}&
			\includegraphics[width=0.1\linewidth]{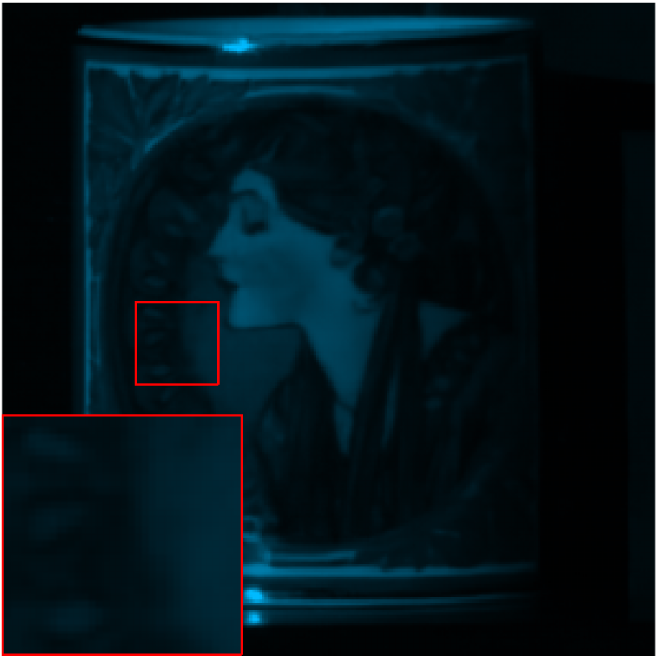}&
			\includegraphics[width=0.1\linewidth]{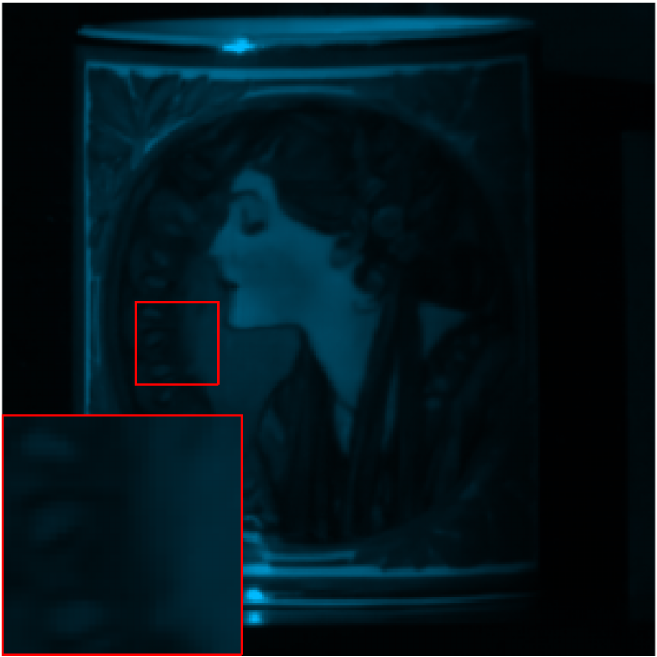}&
			\includegraphics[width=0.1\linewidth]{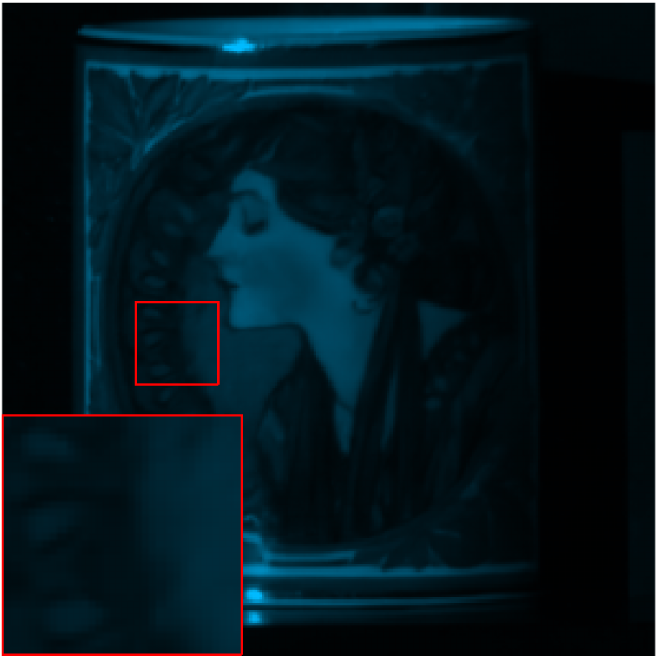}&
			\includegraphics[width=0.1\linewidth]{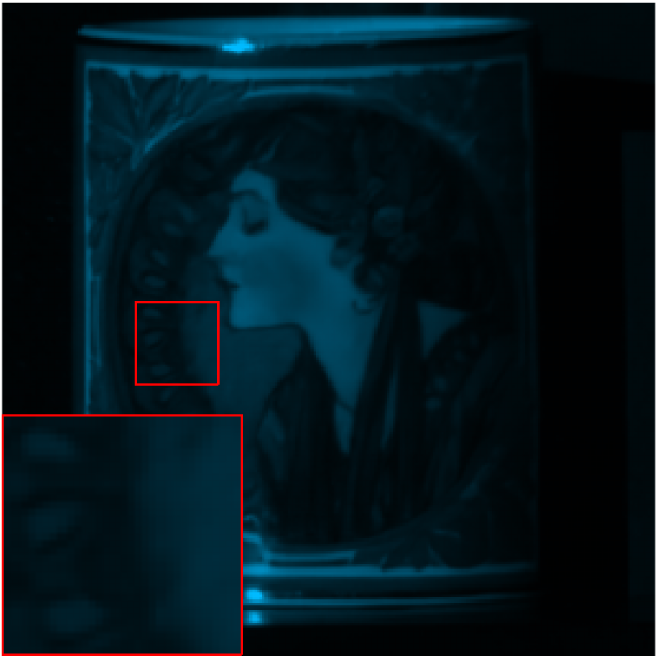}\\
            \includegraphics[width=0.1\linewidth]{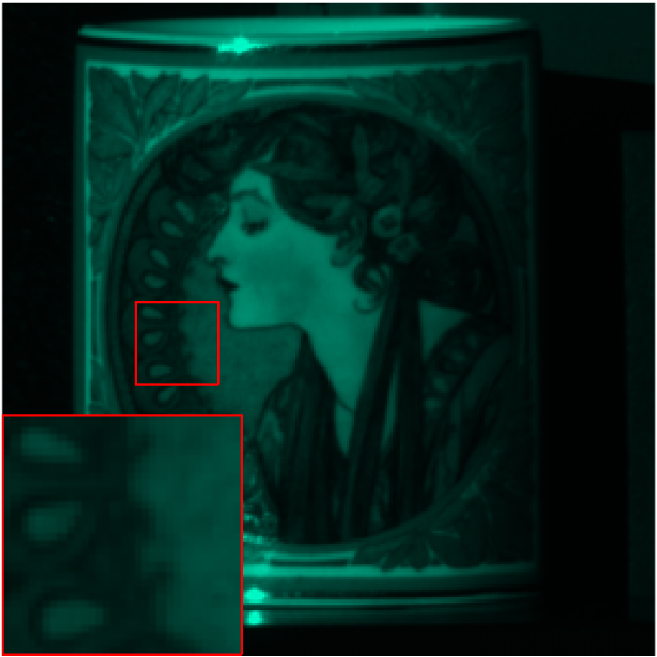}&
			\includegraphics[width=0.1\linewidth]{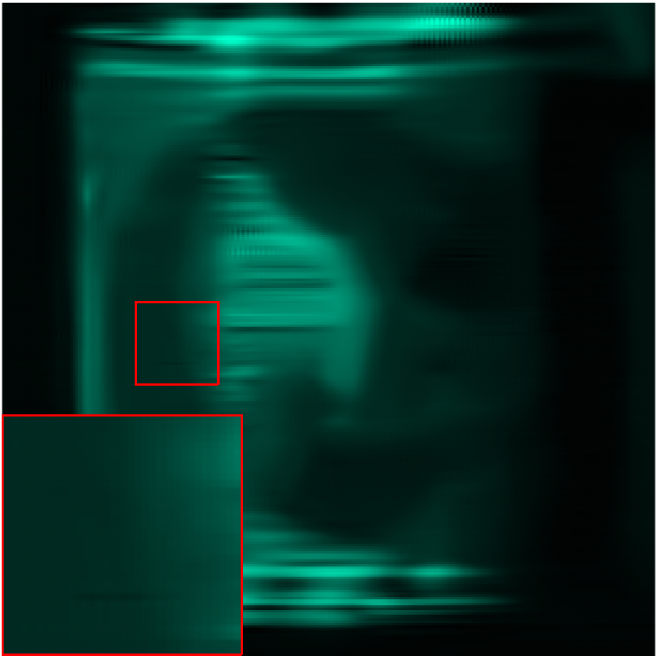}&
			\includegraphics[width=0.1\linewidth]{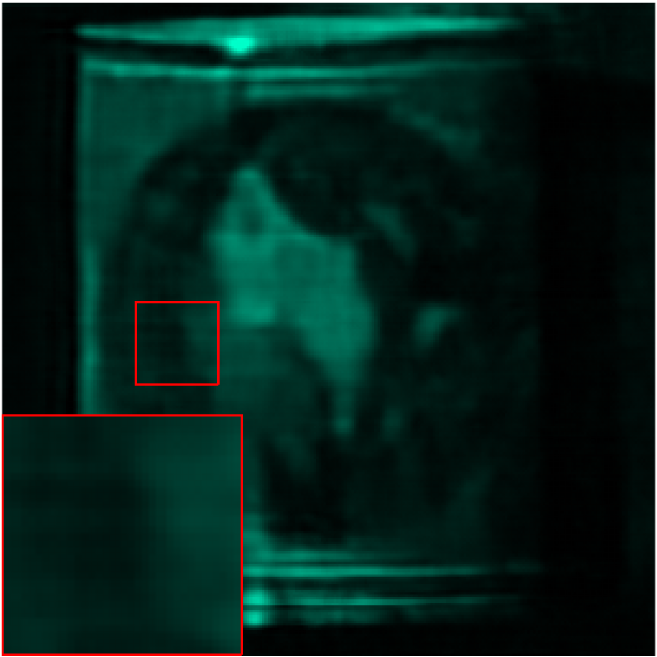}&
			\includegraphics[width=0.1\linewidth]{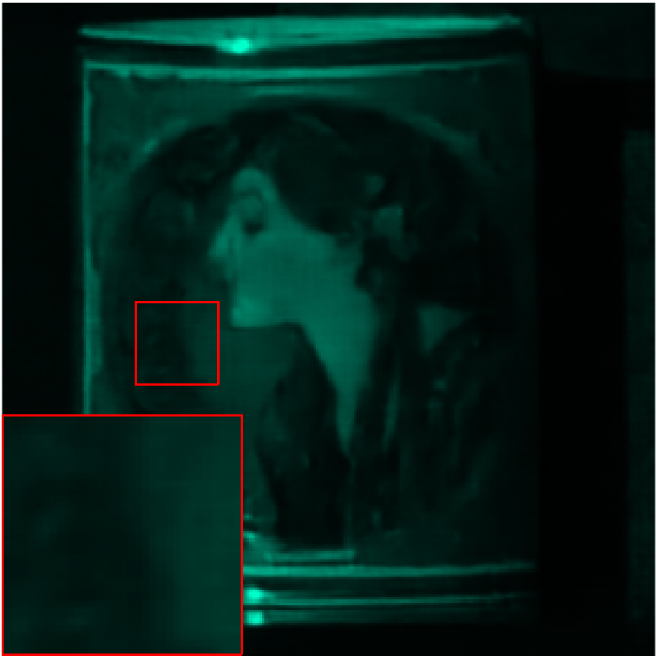}&
			\includegraphics[width=0.1\linewidth]{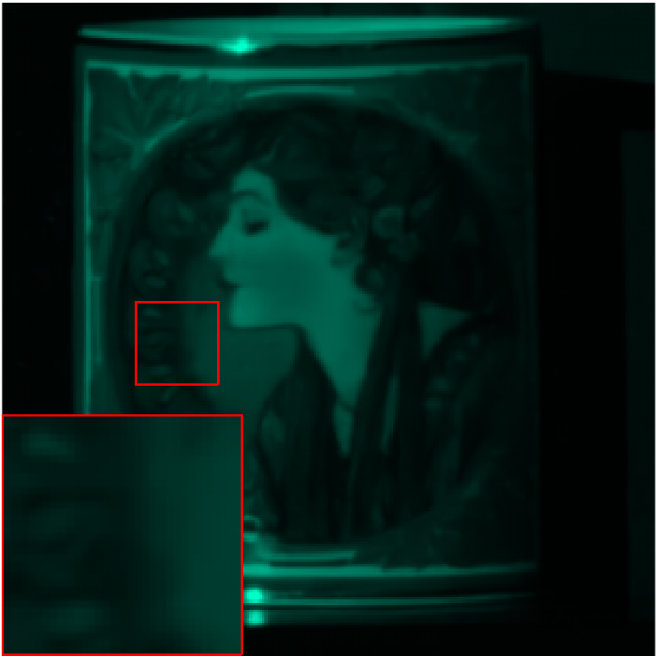}&
                \includegraphics[width=0.1\linewidth]{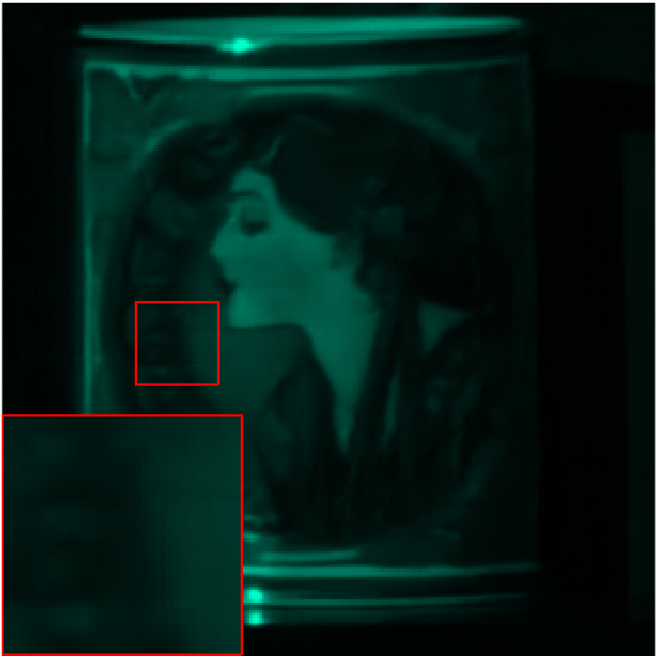}&
			\includegraphics[width=0.1\linewidth]{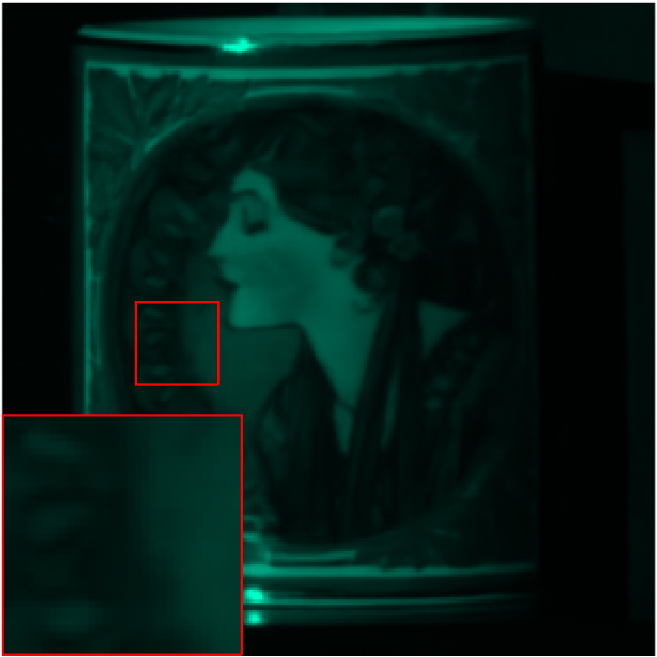}&
			\includegraphics[width=0.1\linewidth]{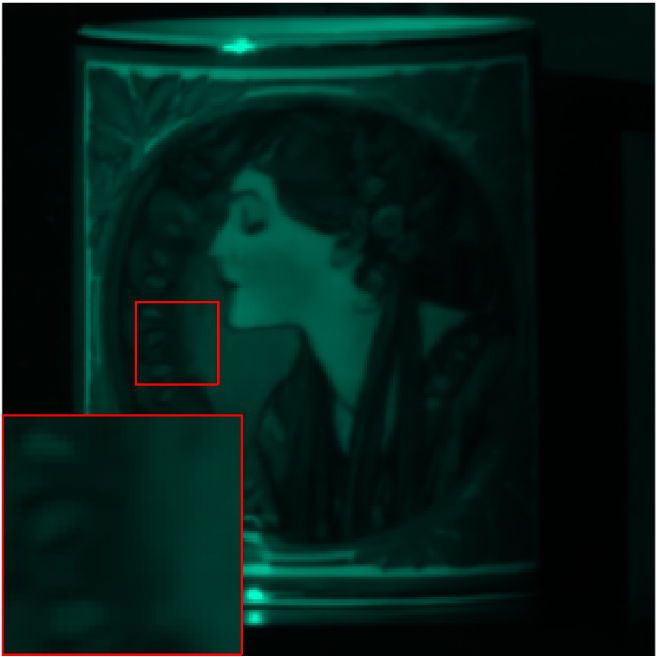}&
			\includegraphics[width=0.1\linewidth]{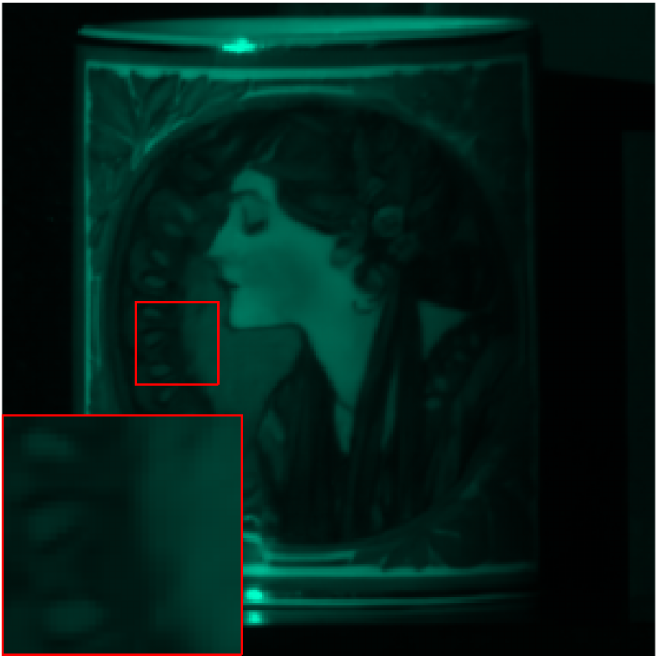}&
			\includegraphics[width=0.1\linewidth]{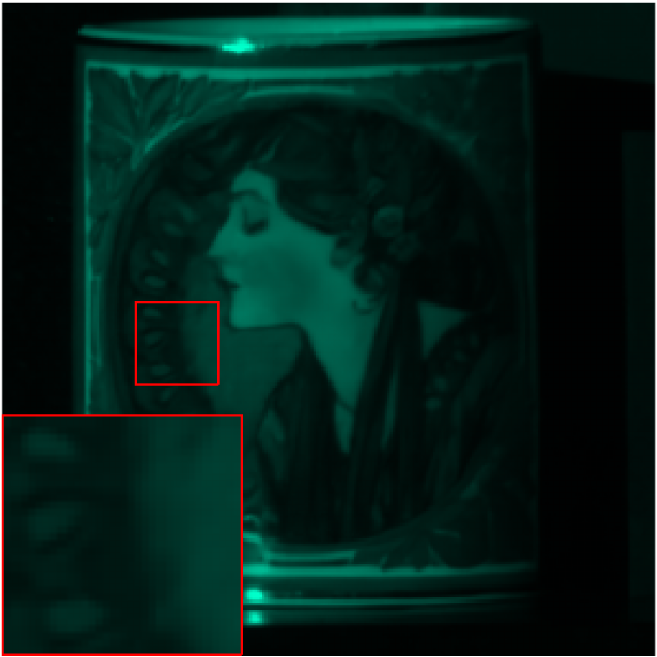}\\
            \includegraphics[width=0.1\linewidth]{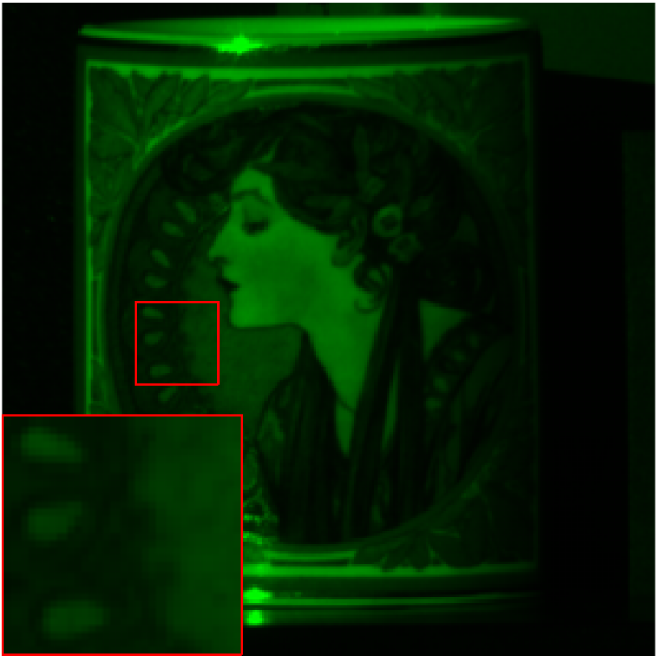}&
			\includegraphics[width=0.1\linewidth]{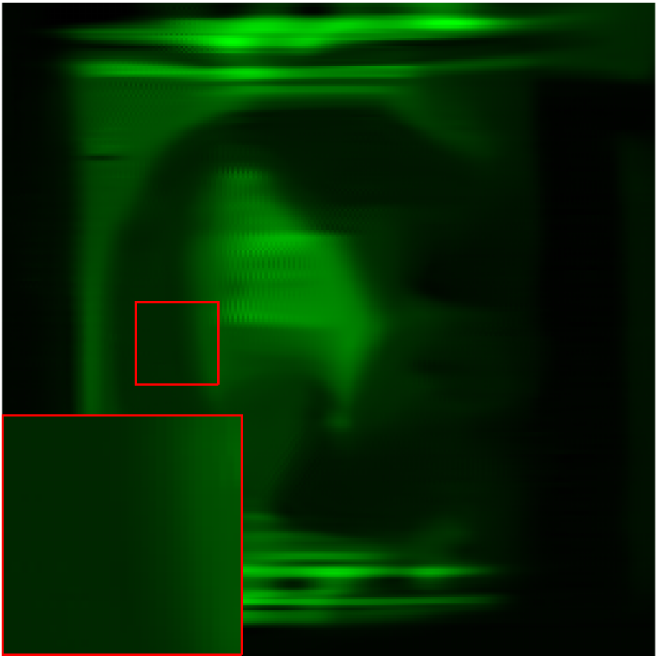}&
			\includegraphics[width=0.1\linewidth]{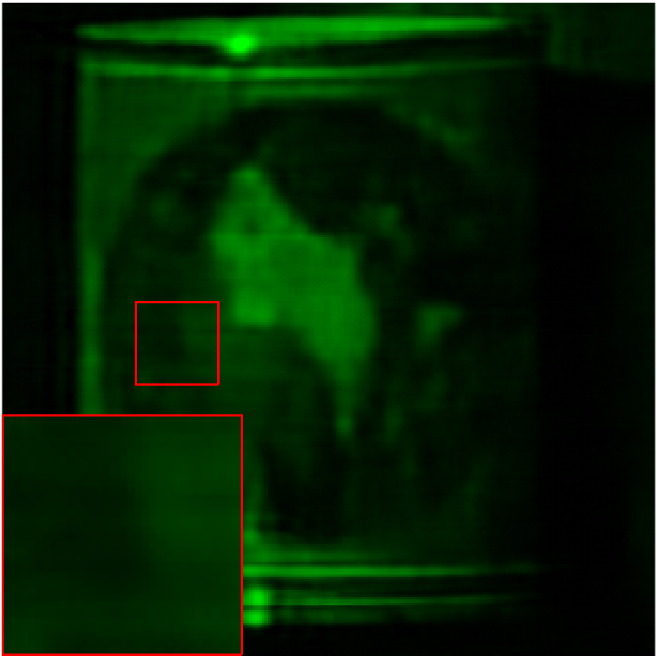}&
			\includegraphics[width=0.1\linewidth]{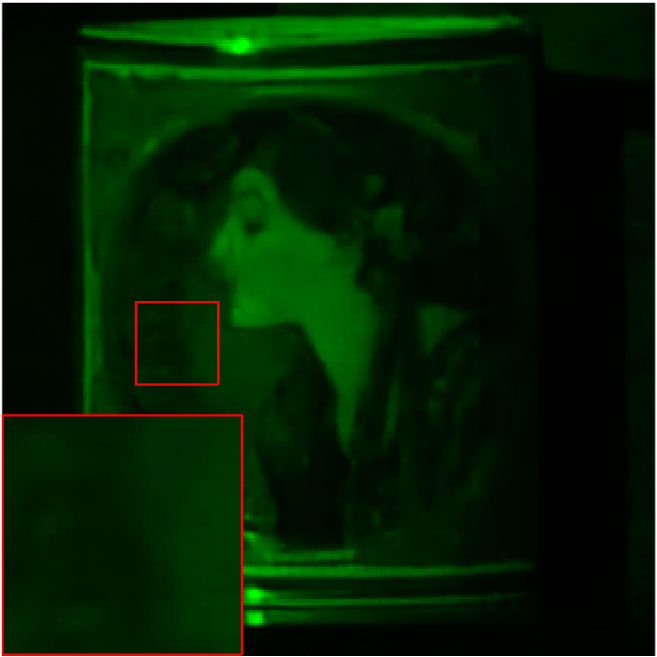}&
			\includegraphics[width=0.1\linewidth]{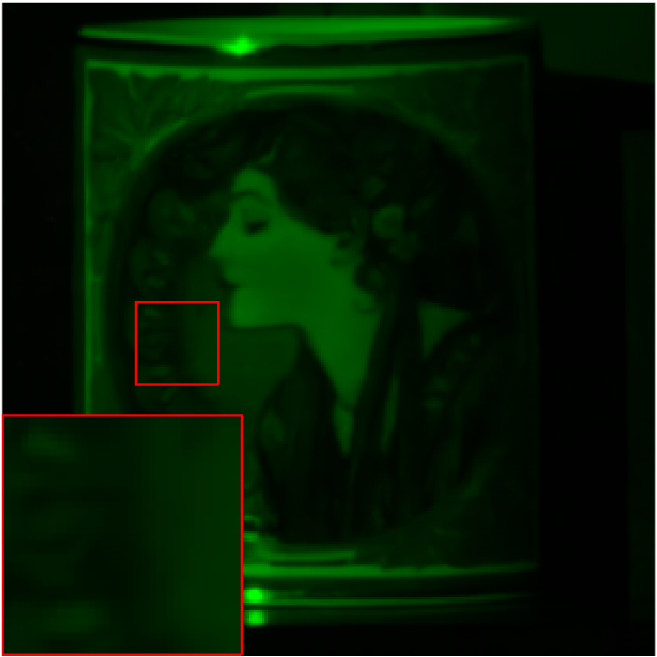}&
                \includegraphics[width=0.1\linewidth]{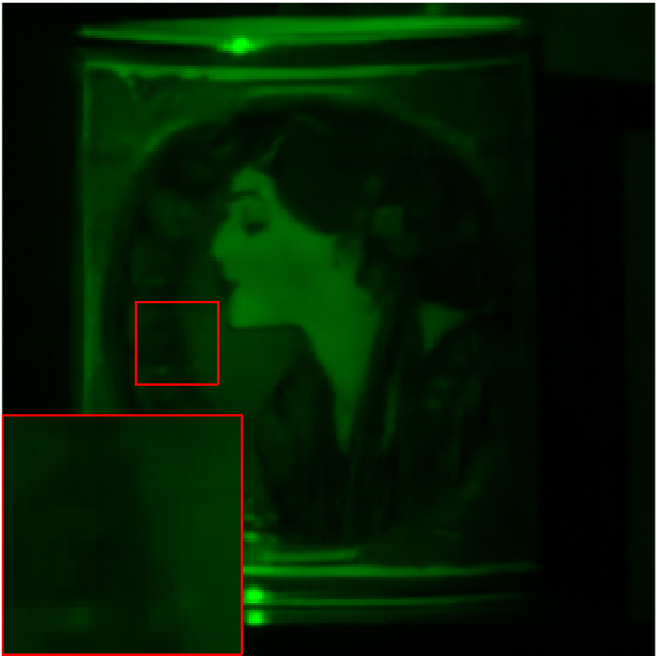}&
			\includegraphics[width=0.1\linewidth]{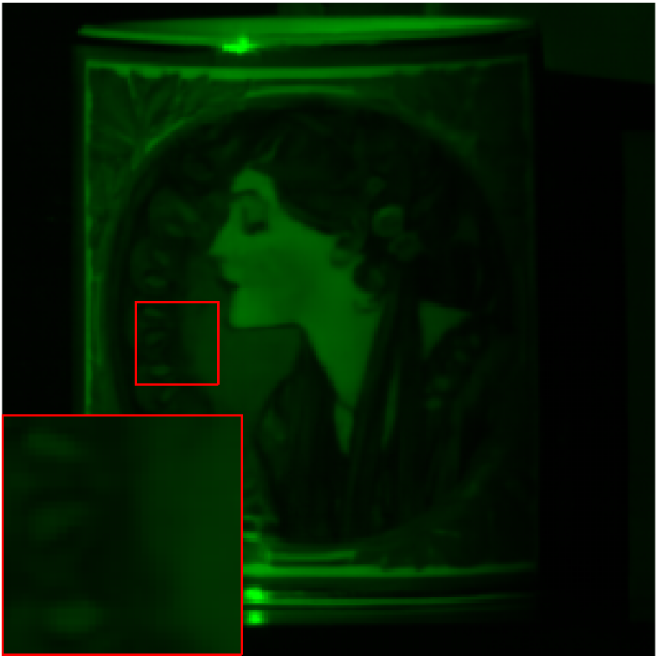}&
			\includegraphics[width=0.1\linewidth]{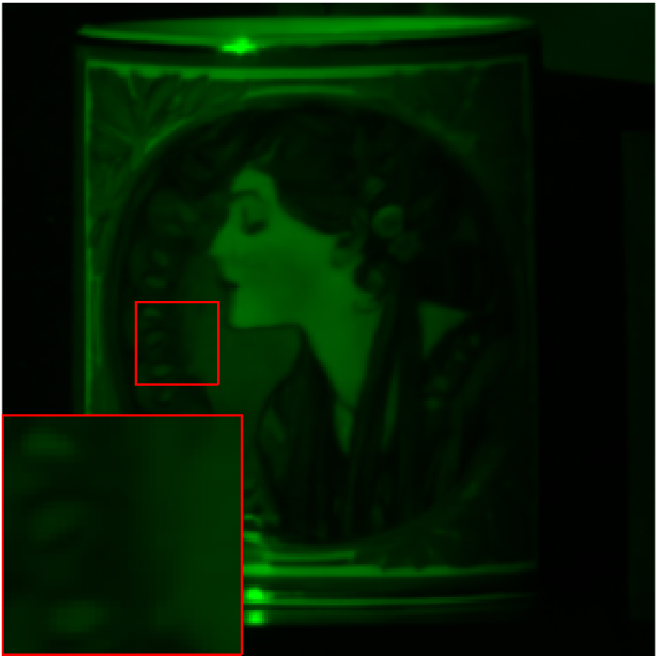}&
			\includegraphics[width=0.1\linewidth]{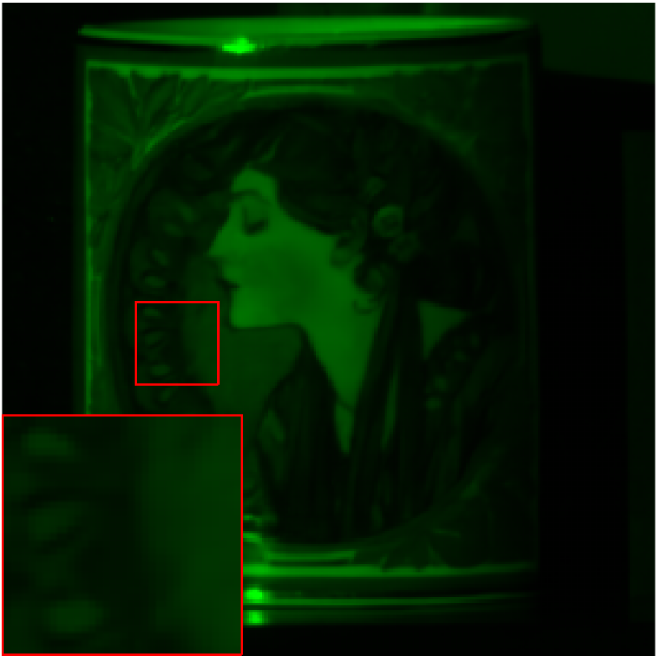}&
			\includegraphics[width=0.1\linewidth]{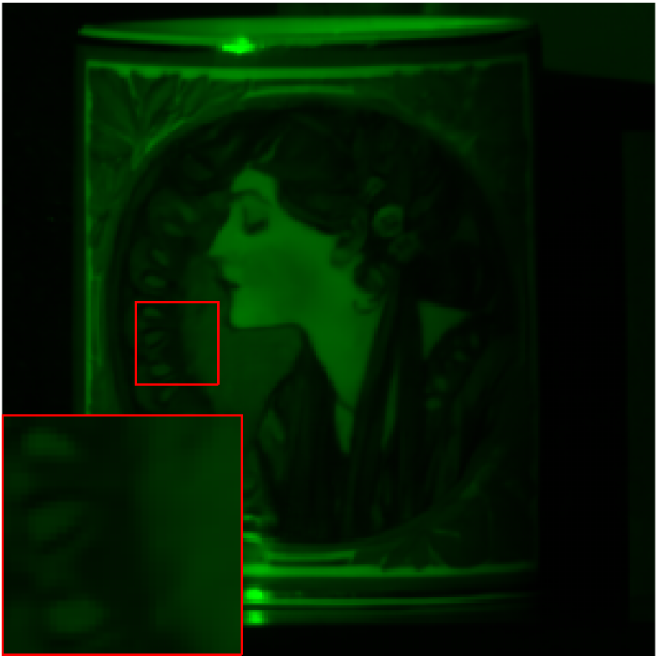}\\
            \includegraphics[width=0.1\linewidth]{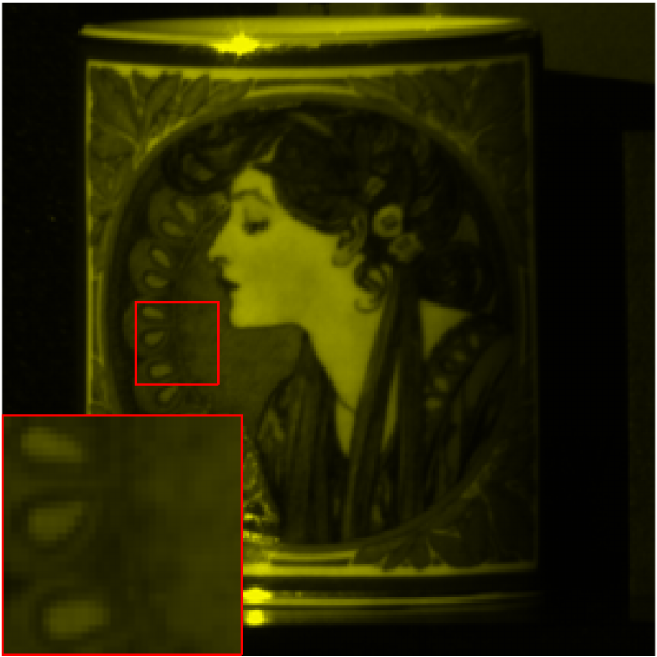}&
			\includegraphics[width=0.1\linewidth]{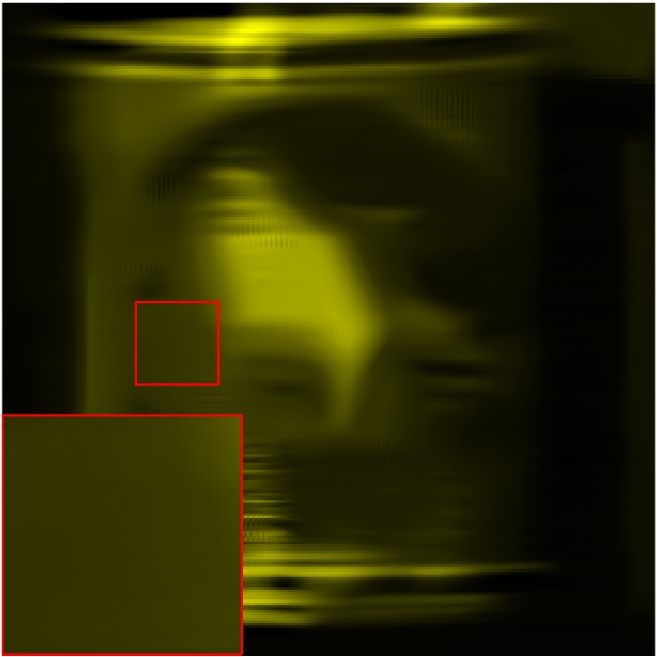}&
			\includegraphics[width=0.1\linewidth]{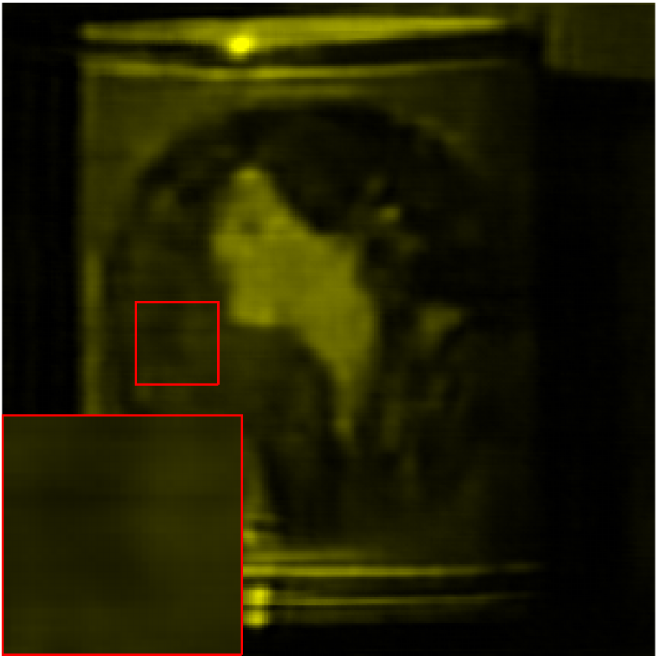}&
			\includegraphics[width=0.1\linewidth]{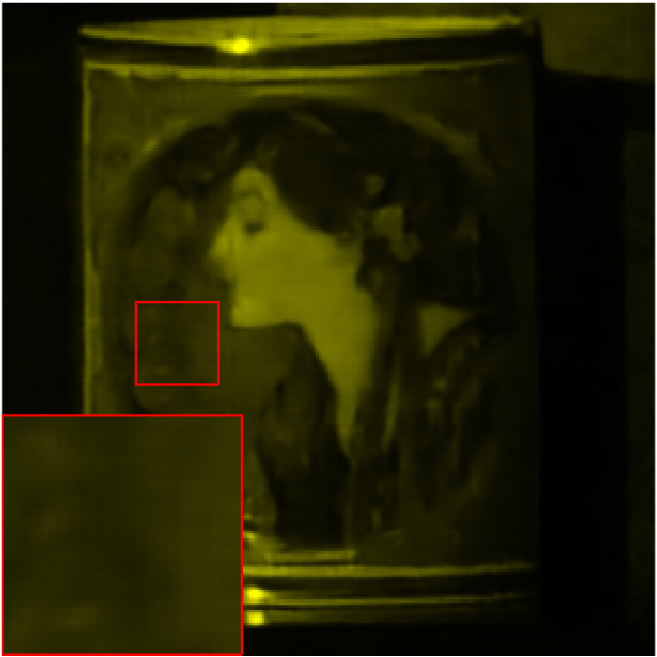}&
			\includegraphics[width=0.1\linewidth]{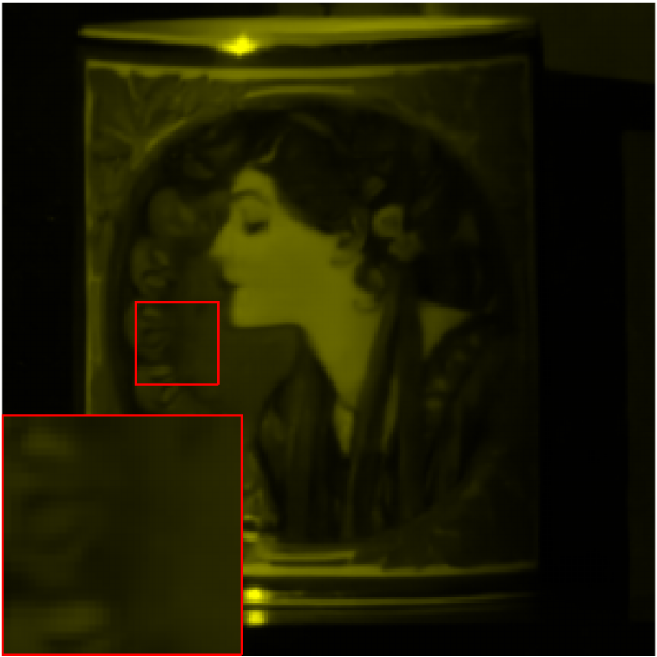}&
                \includegraphics[width=0.1\linewidth]{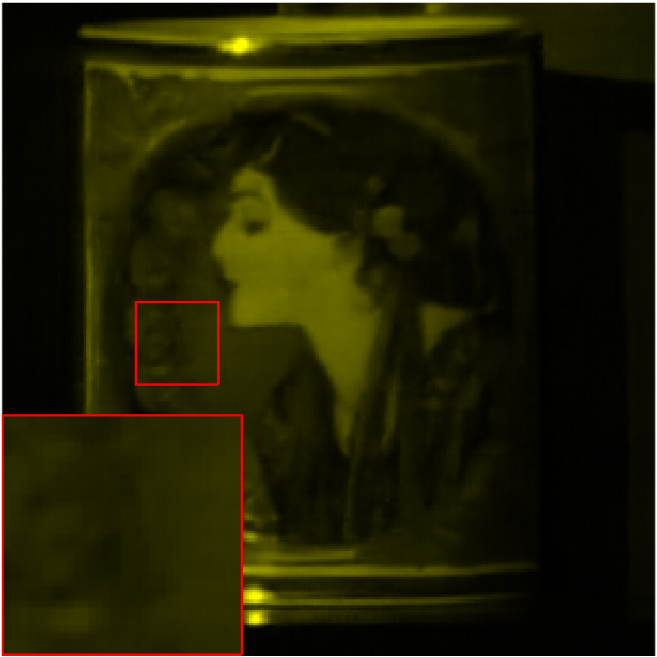}&
			\includegraphics[width=0.1\linewidth]{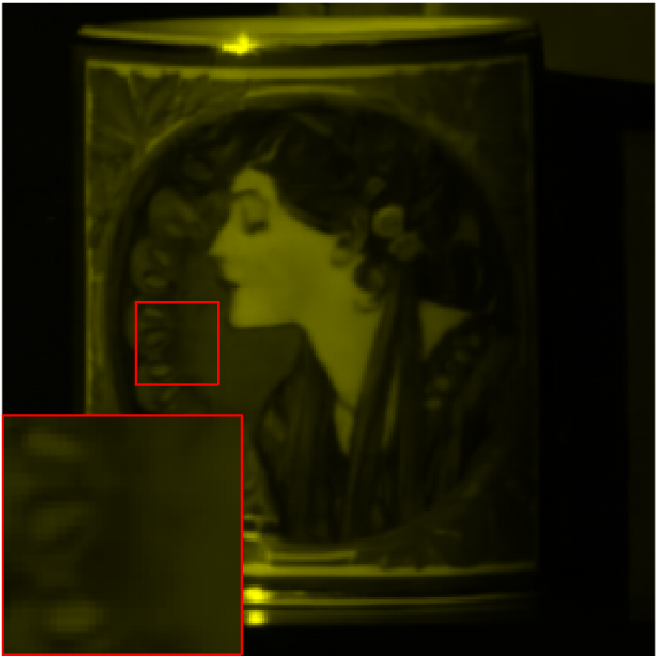}&
			\includegraphics[width=0.1\linewidth]{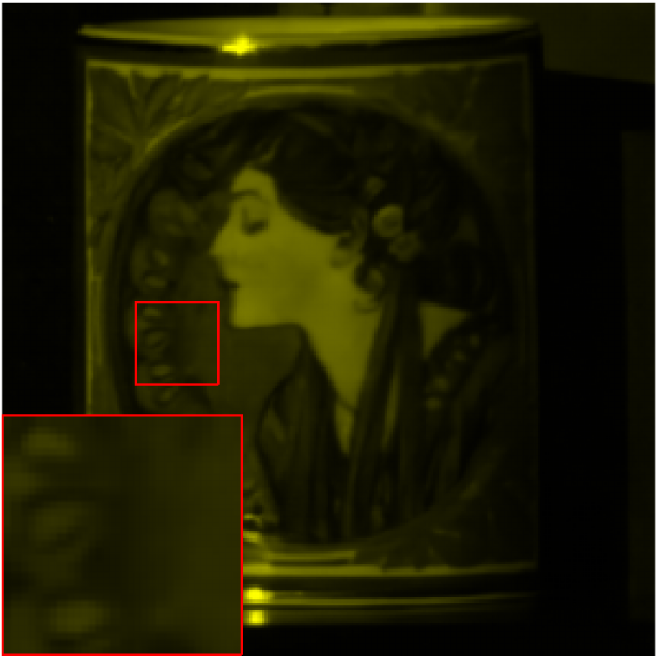}&
			\includegraphics[width=0.1\linewidth]{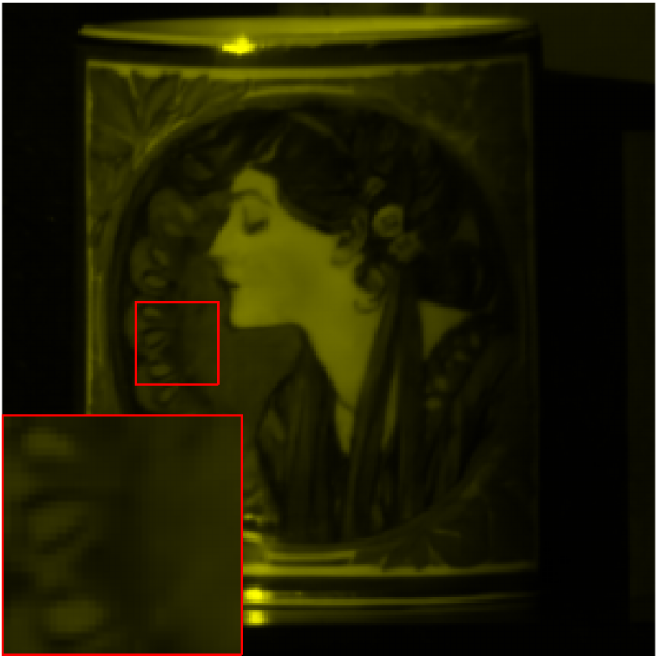}&
			\includegraphics[width=0.1\linewidth]{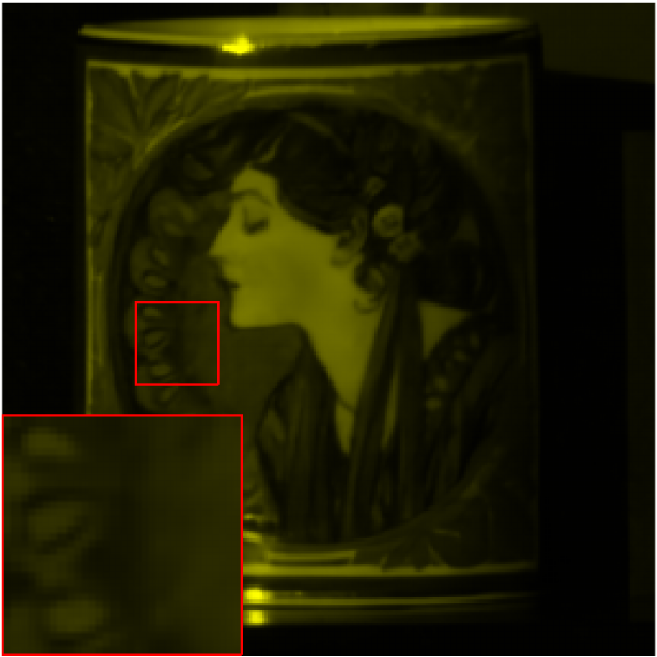}\\
            \includegraphics[width=0.1\linewidth]{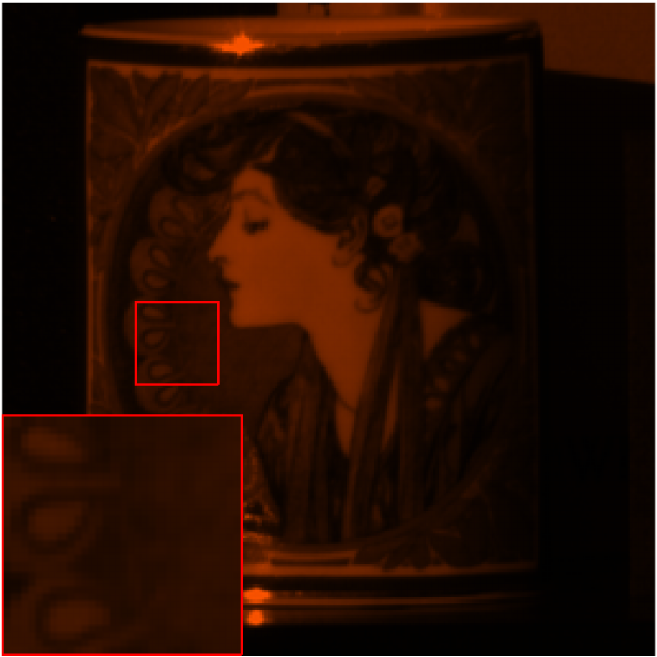}&
			\includegraphics[width=0.1\linewidth]{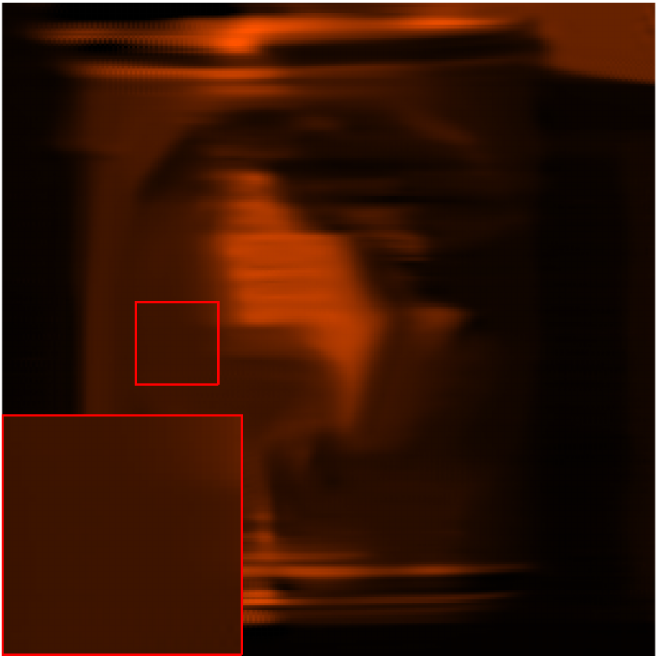}&
			\includegraphics[width=0.1\linewidth]{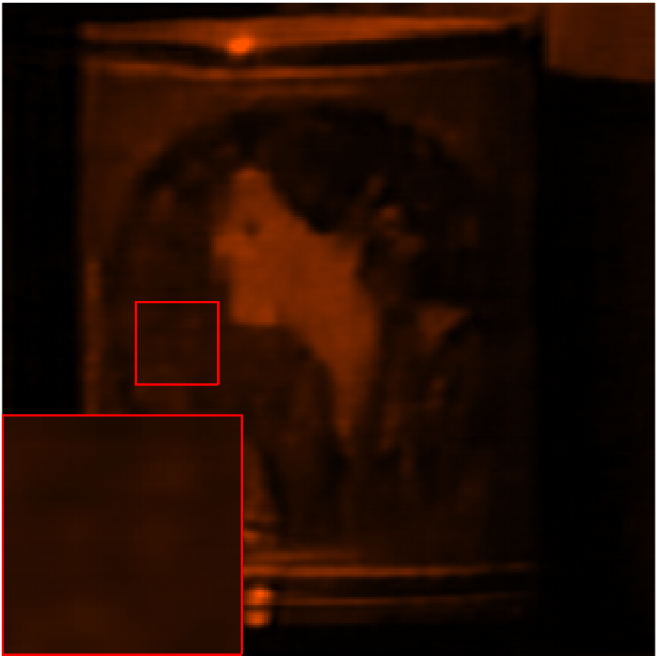}&
			\includegraphics[width=0.1\linewidth]{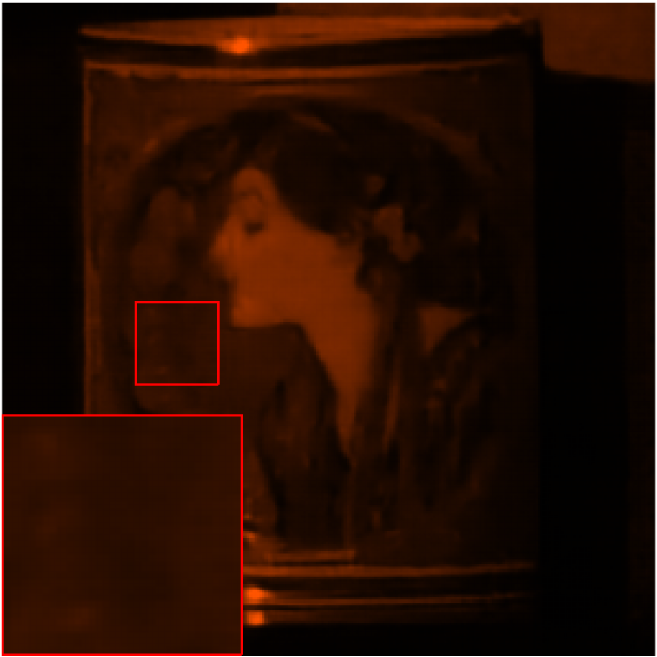}&
			\includegraphics[width=0.1\linewidth]{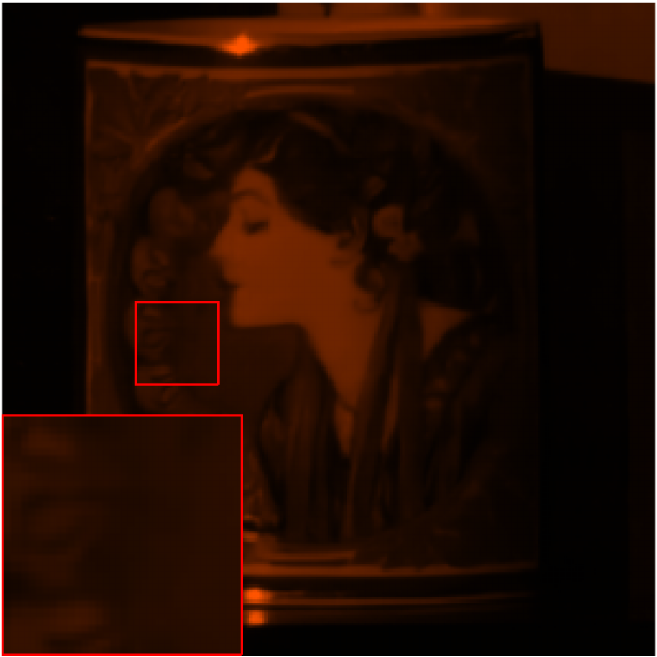}&
                \includegraphics[width=0.1\linewidth]{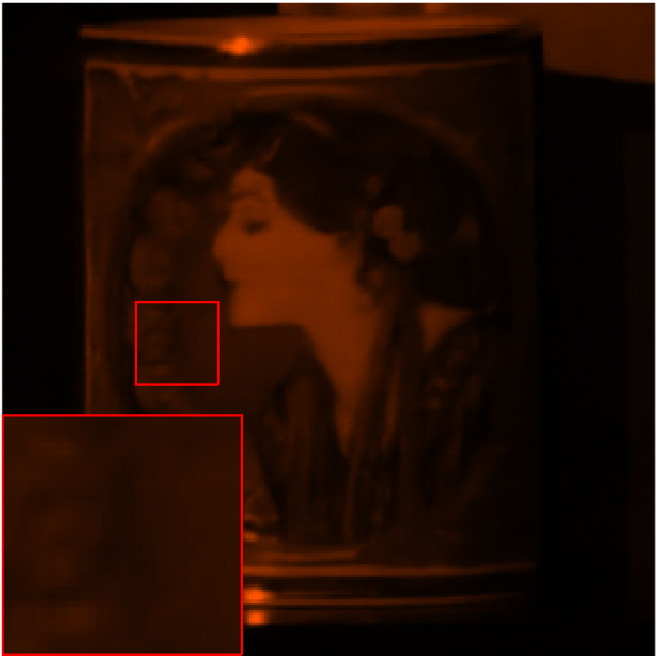}&
			\includegraphics[width=0.1\linewidth]{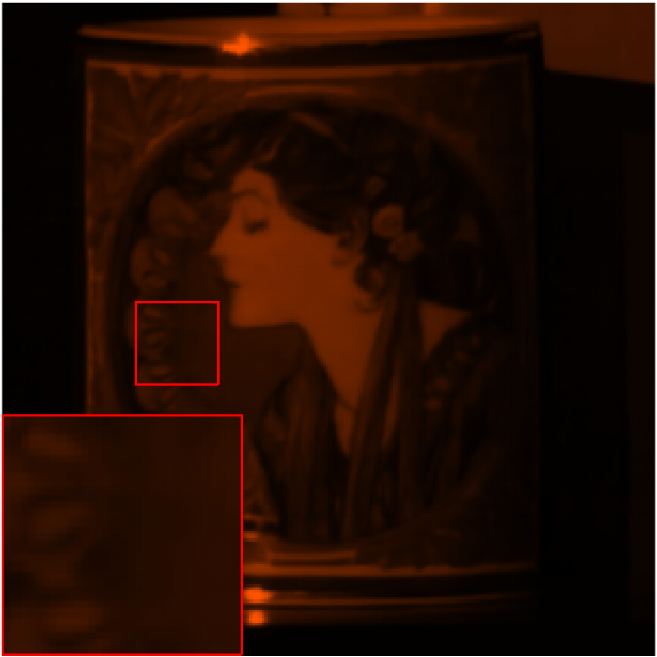}&
			\includegraphics[width=0.1\linewidth]{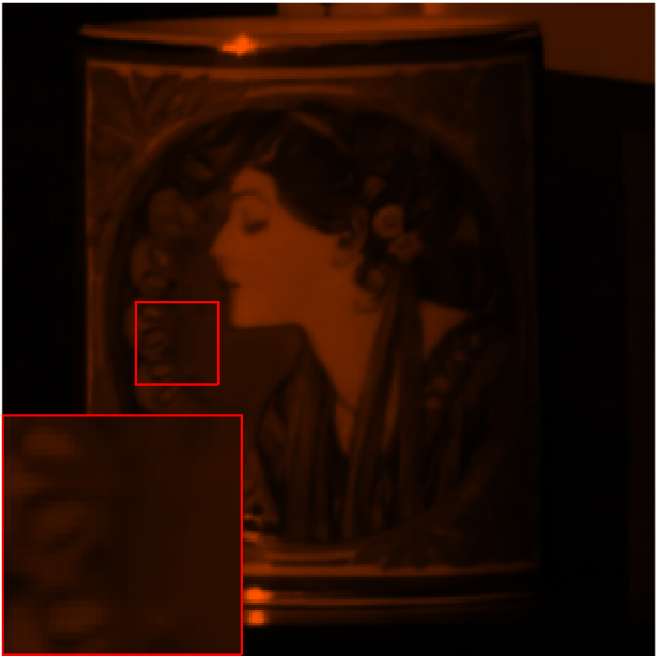}&
			\includegraphics[width=0.1\linewidth]{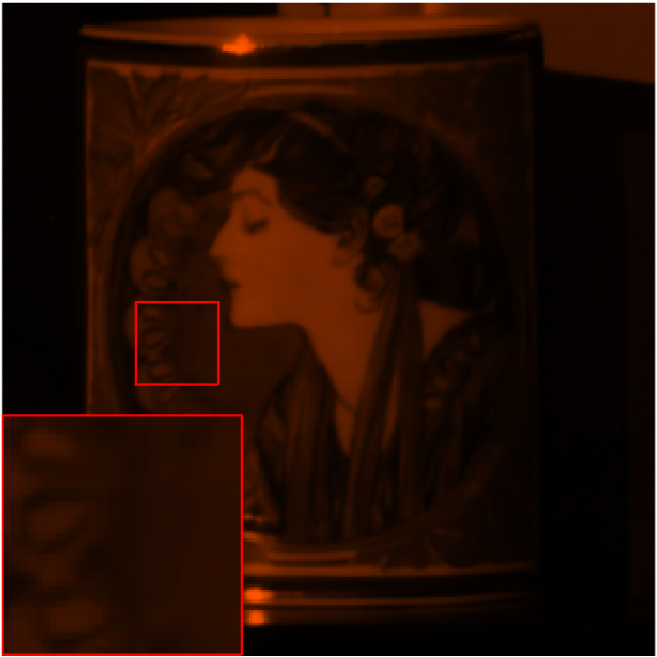}&
			\includegraphics[width=0.1\linewidth]{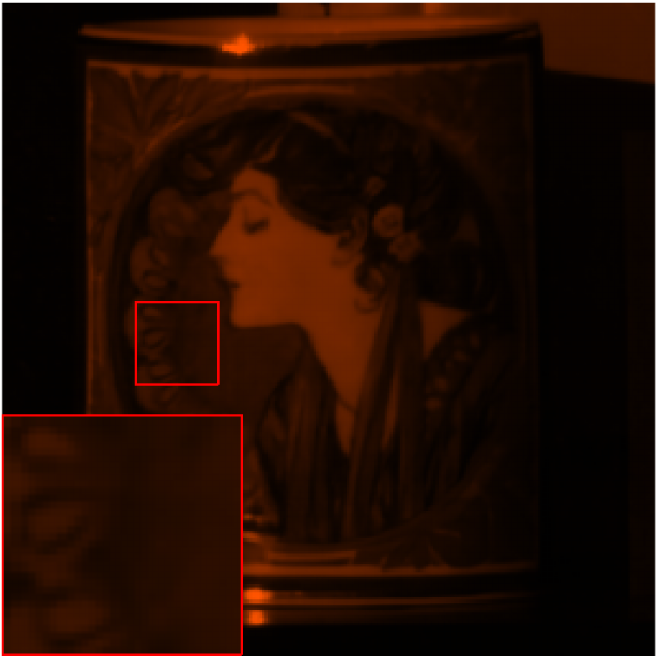}\\
            Truth&DeSCI&$\lambda$-Net&TSA-Net&BIRNAT&GAP-Net&PSRSCI&RCUMP&Block-9&LoRun-9
		\end{tabular}} 
		\caption{CASSI results of \emph{scene01} in different bands, from top to bottom $\lambda \in \{\SI{481.5}{\nano\meter}, \SI{492.5}{\nano\meter}, \SI{522.5}{\nano\meter}, \SI{567.5}{\nano\meter}, \SI{604.5}{\nano\meter}\}$.}
		\label{cassi_dbands}
        \vspace{-6mm}
	\end{center}
\end{figure*}

\begin{figure*}
	\footnotesize
	\setlength{\tabcolsep}{1pt}
 \newcommand{\tabincell}[2]{\begin{tabular}{@{}#1@{}}#2\end{tabular}}
	\begin{center}
            \scalebox{0.94}{
		\begin{tabular}{cccccccccc}
			\includegraphics[width=0.1\linewidth]{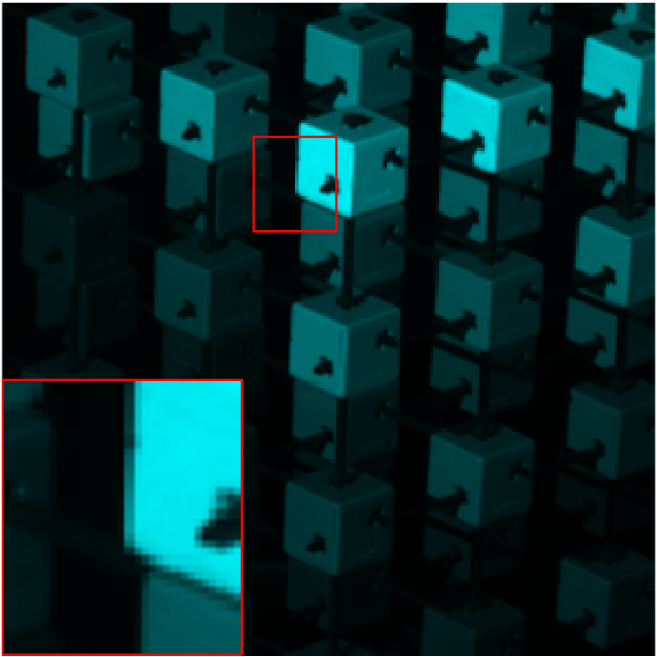}&
			\includegraphics[width=0.1\linewidth]{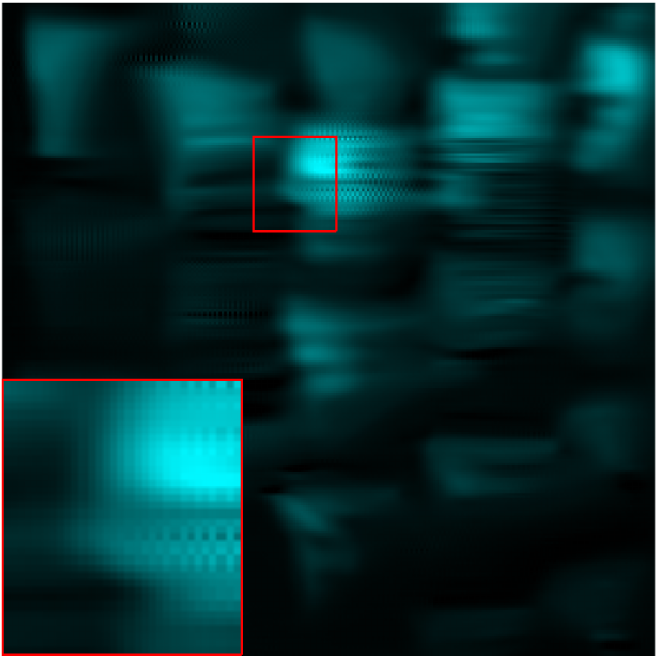}&
			\includegraphics[width=0.1\linewidth]{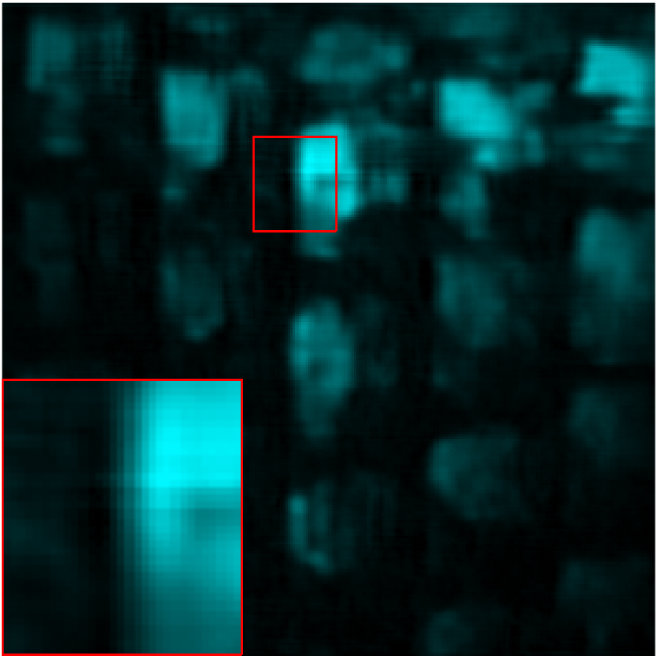}&
			\includegraphics[width=0.1\linewidth]{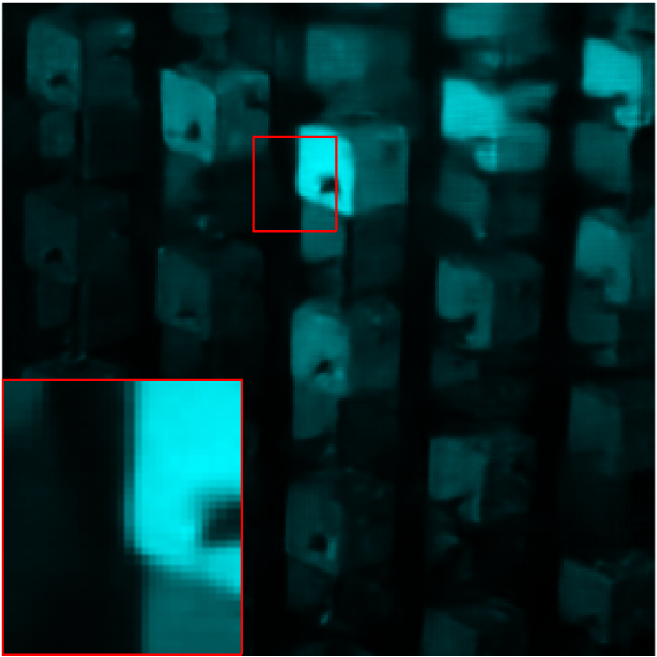}&
			\includegraphics[width=0.1\linewidth]{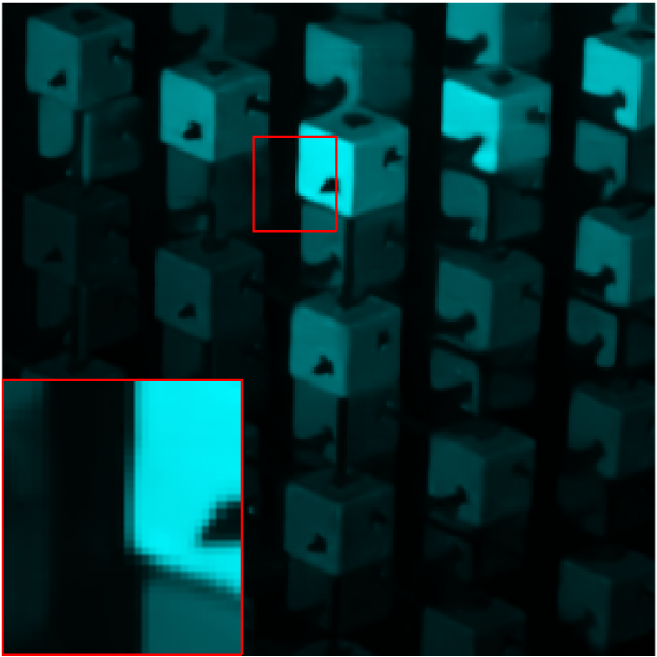}&
                \includegraphics[width=0.1\linewidth]{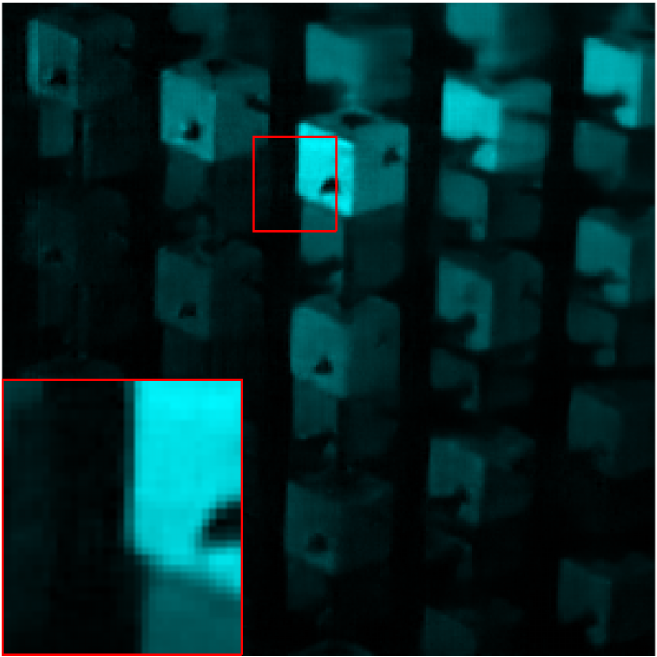}&
			\includegraphics[width=0.1\linewidth]{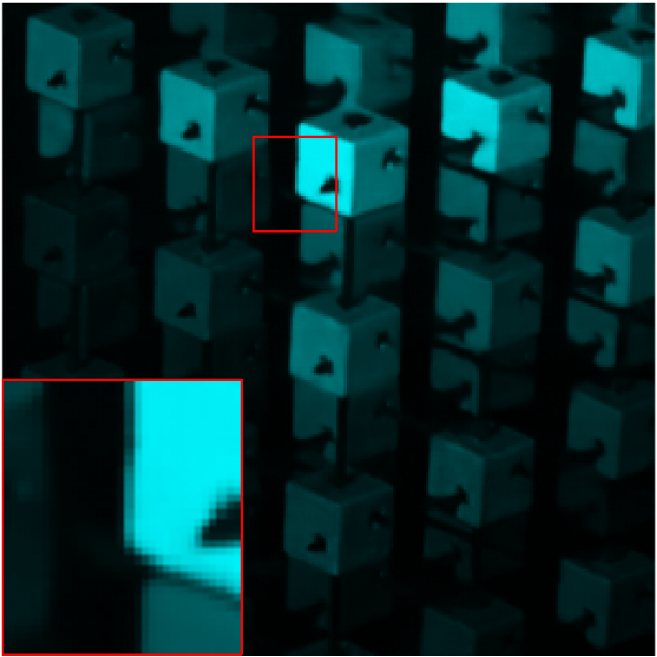}&
			\includegraphics[width=0.1\linewidth]{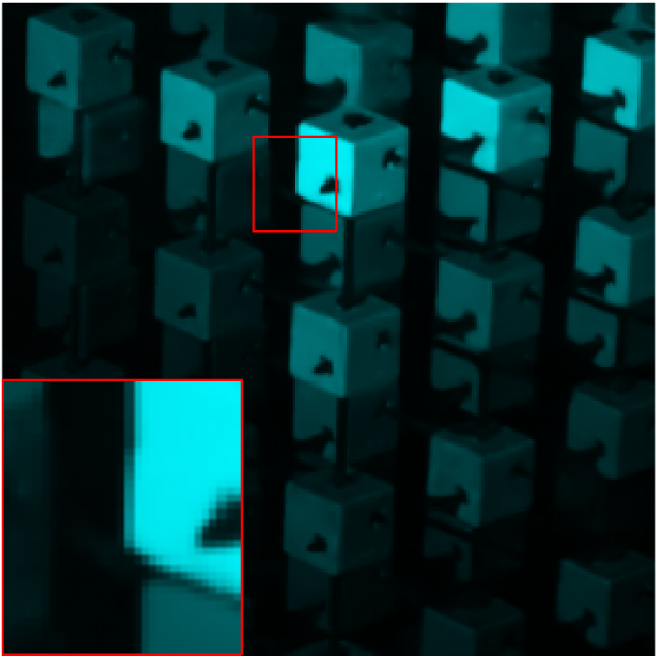}&
			\includegraphics[width=0.1\linewidth]{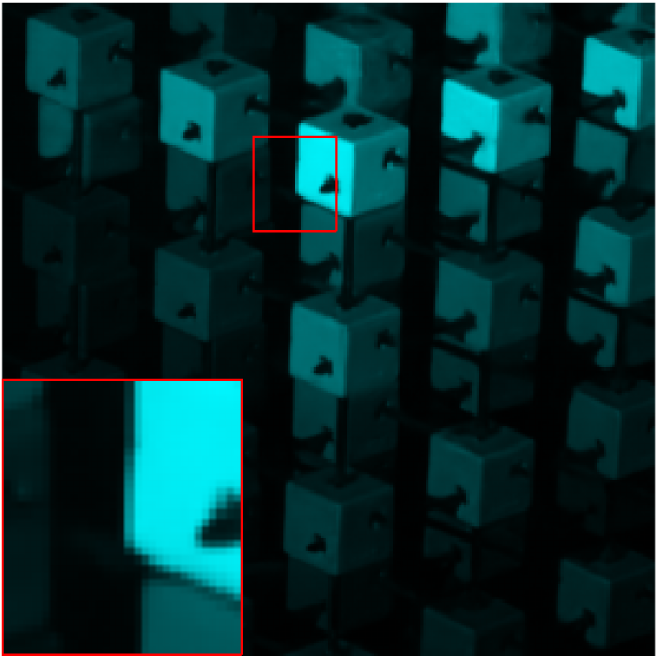}&
			\includegraphics[width=0.1\linewidth]{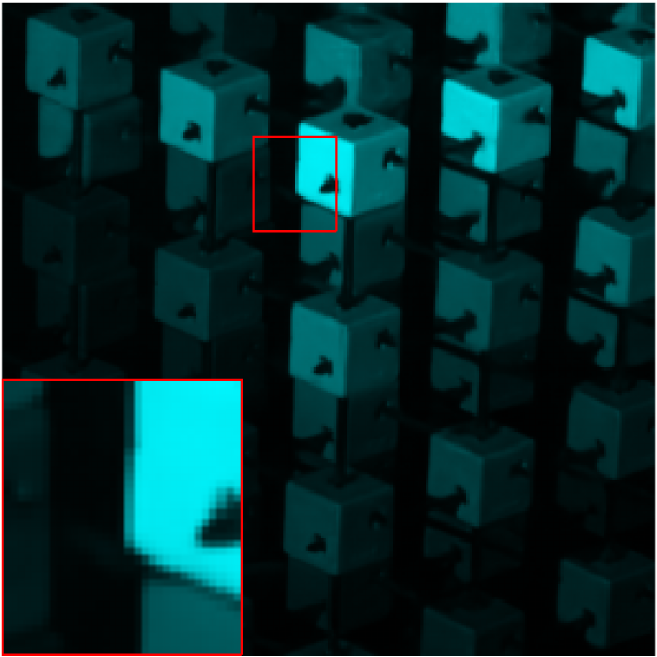}\\
            \includegraphics[width=0.1\linewidth]{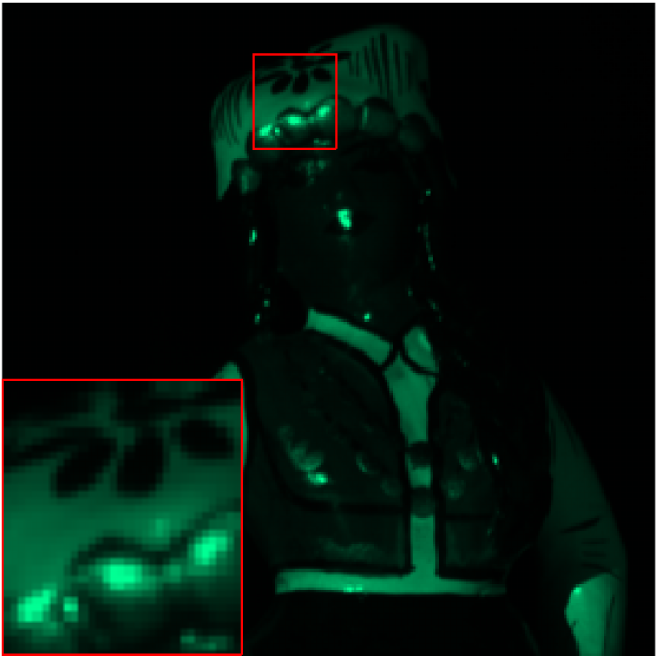}&
			\includegraphics[width=0.1\linewidth]{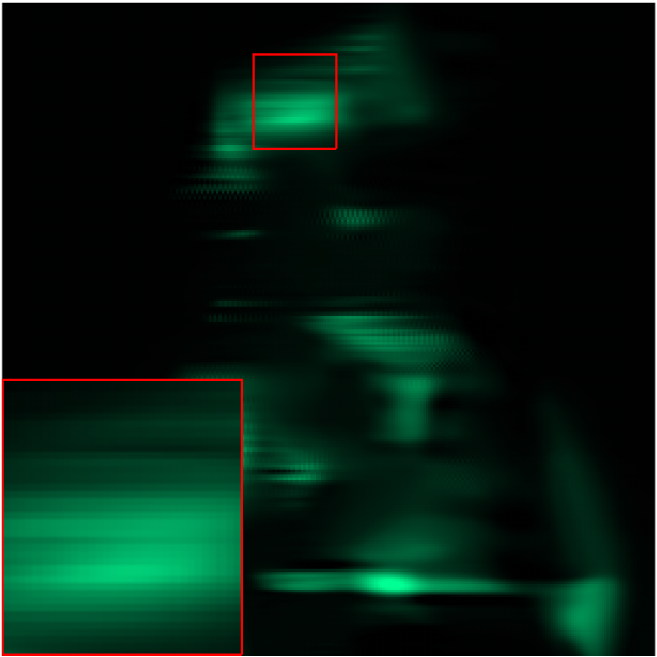}&
			\includegraphics[width=0.1\linewidth]{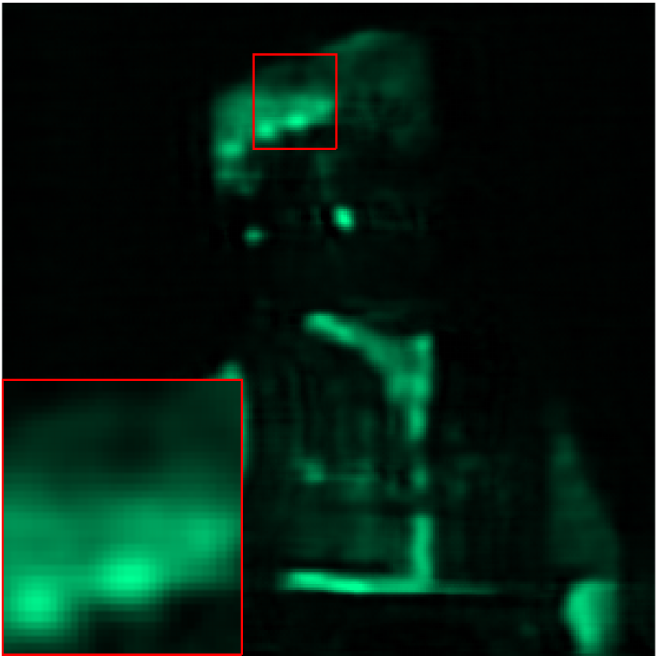}&
			\includegraphics[width=0.1\linewidth]{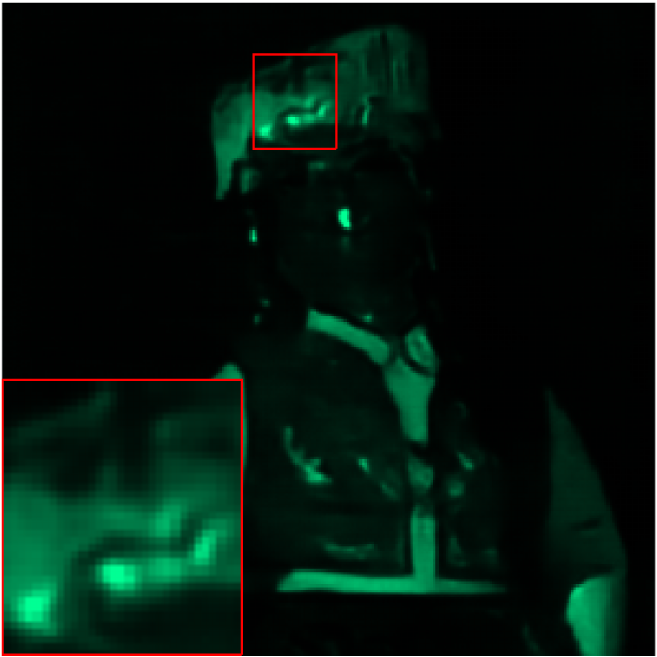}&
			\includegraphics[width=0.1\linewidth]{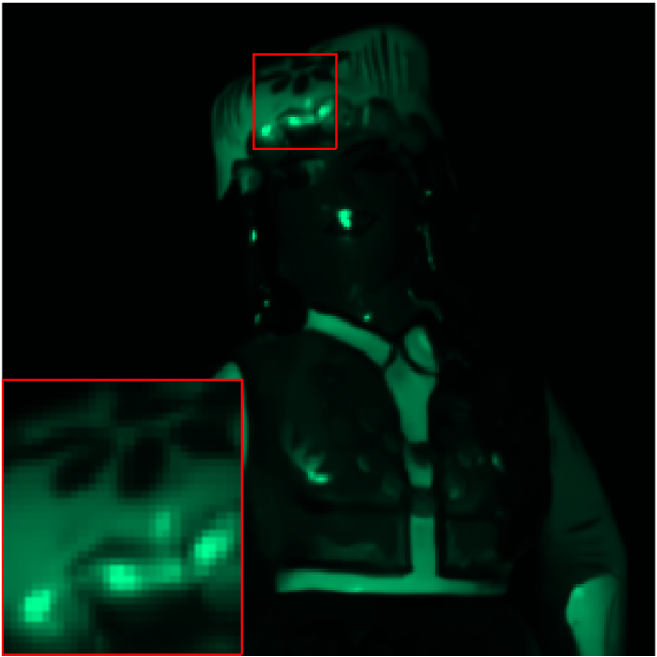}&
                \includegraphics[width=0.1\linewidth]{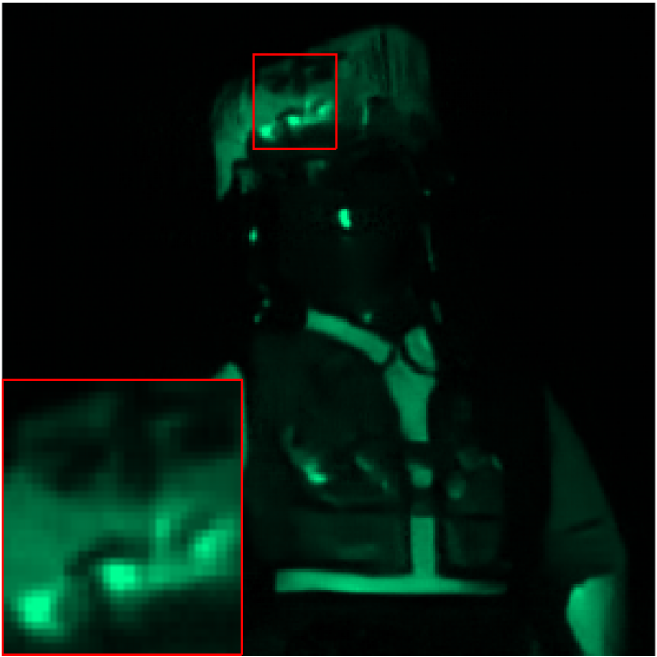}&
			\includegraphics[width=0.1\linewidth]{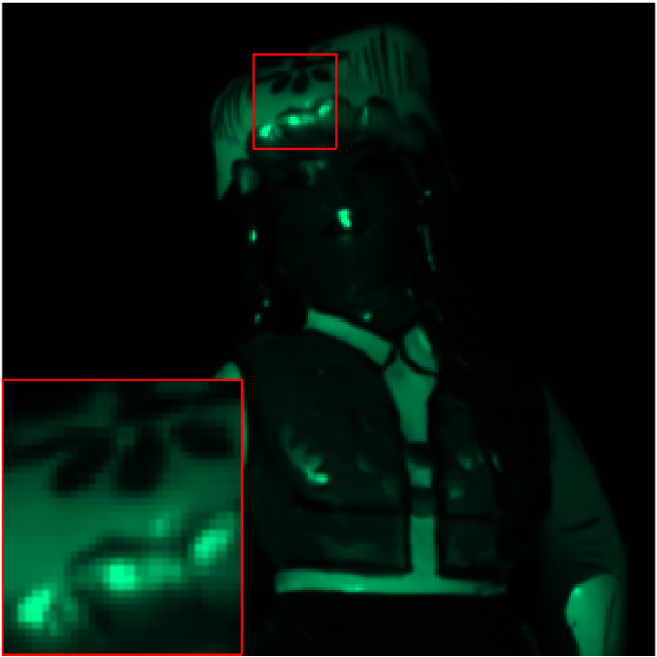}&
			\includegraphics[width=0.1\linewidth]{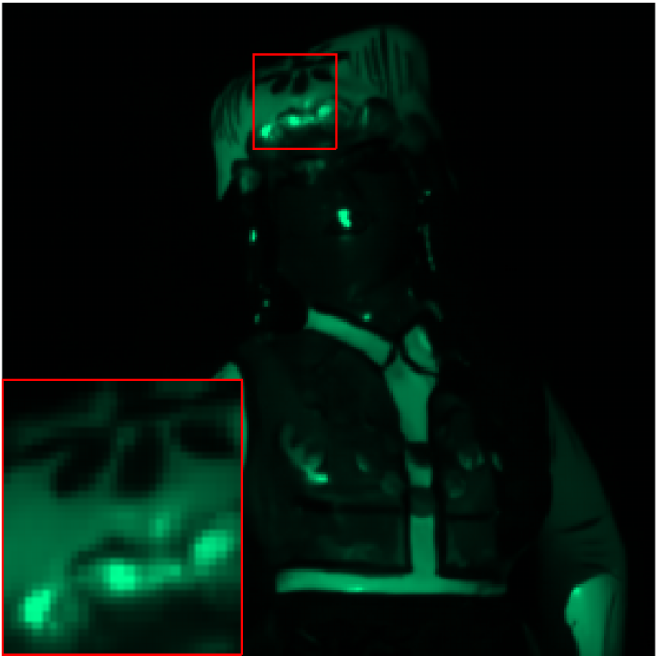}&
			\includegraphics[width=0.1\linewidth]{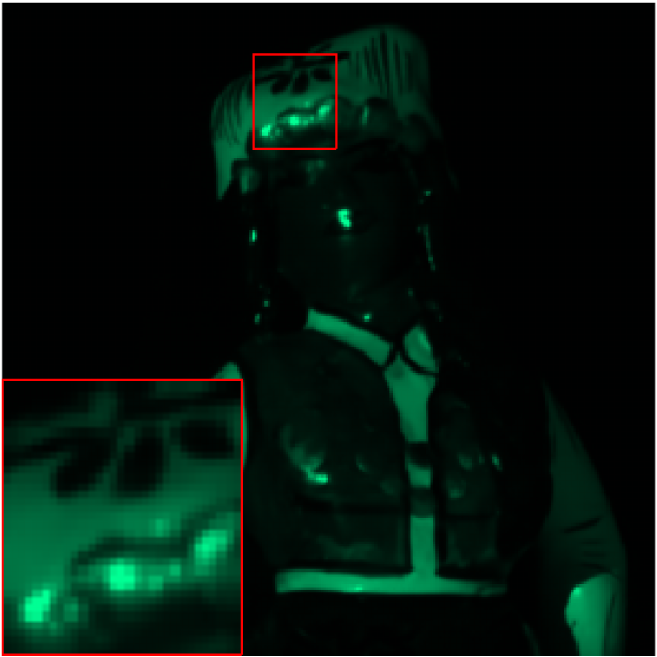}&
			\includegraphics[width=0.1\linewidth]{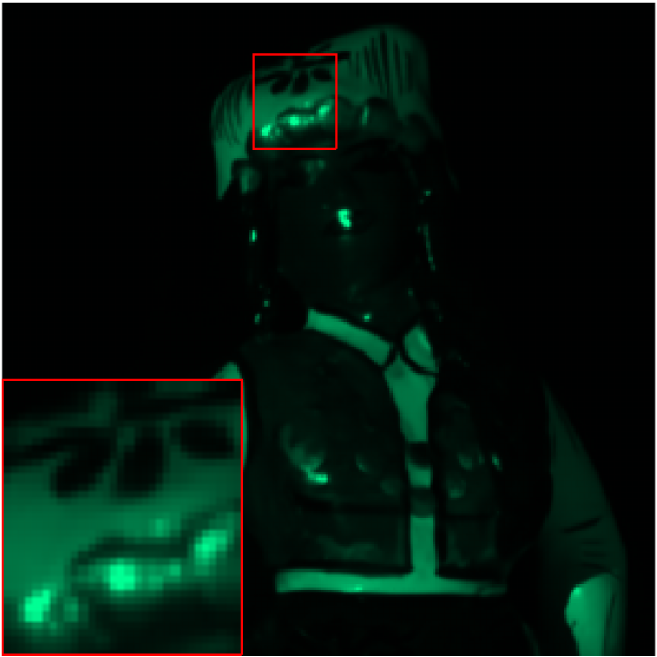}\\
            \includegraphics[width=0.1\linewidth]{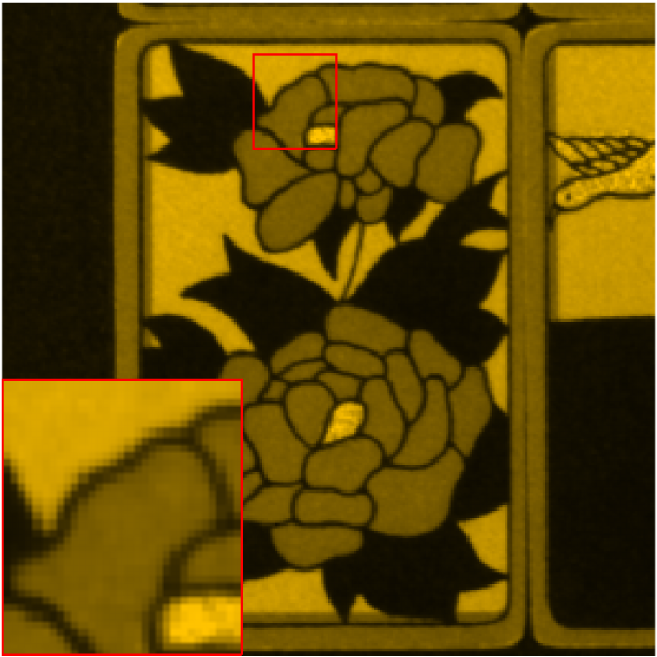}&
			\includegraphics[width=0.1\linewidth]{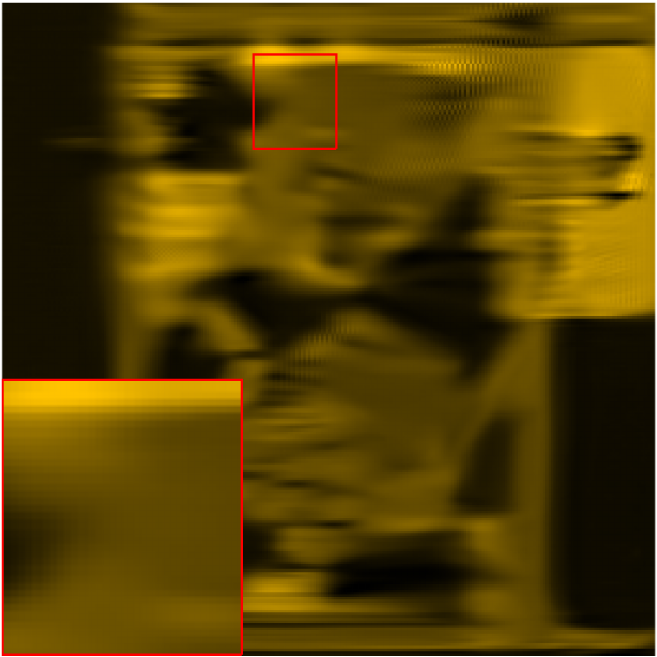}&
			\includegraphics[width=0.1\linewidth]{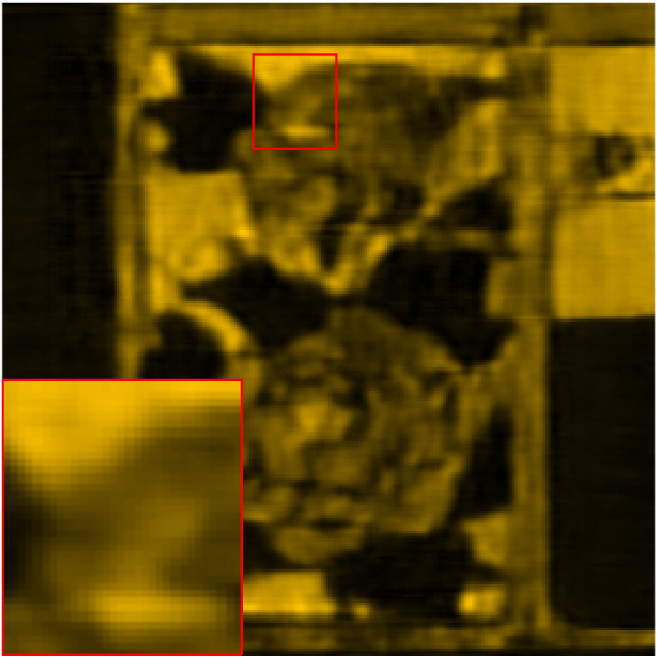}&
			\includegraphics[width=0.1\linewidth]{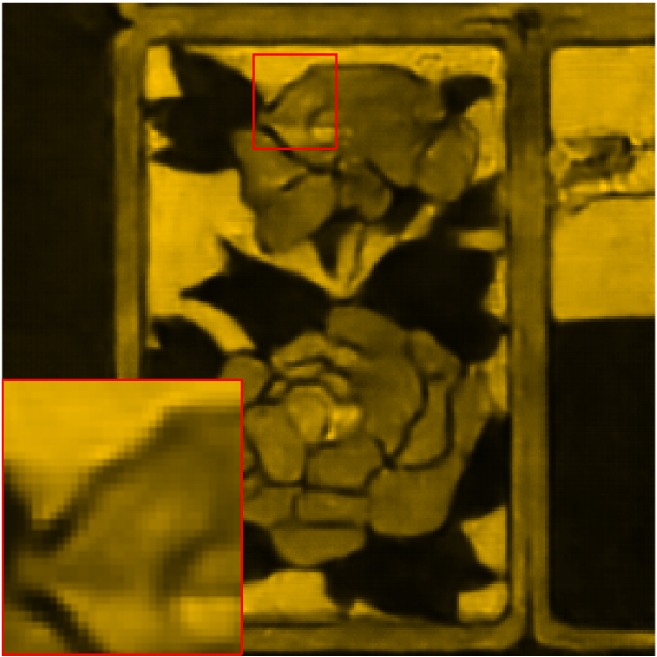}&
			\includegraphics[width=0.1\linewidth]{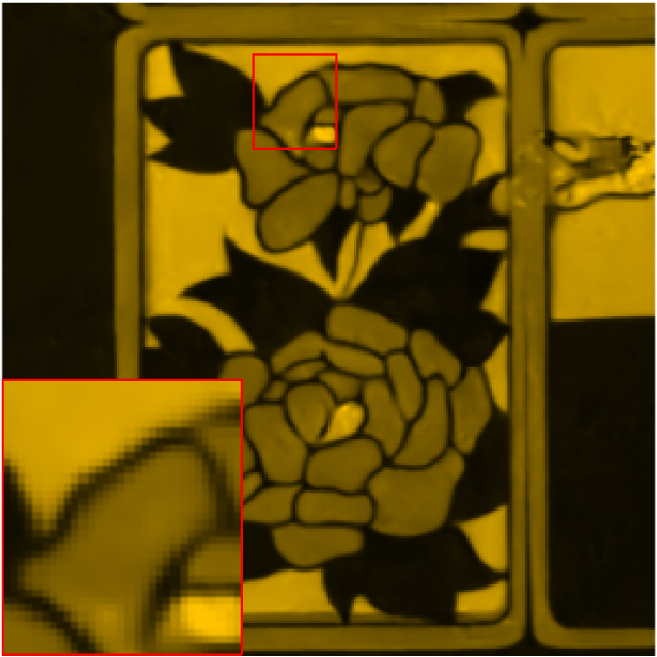}&
                \includegraphics[width=0.1\linewidth]{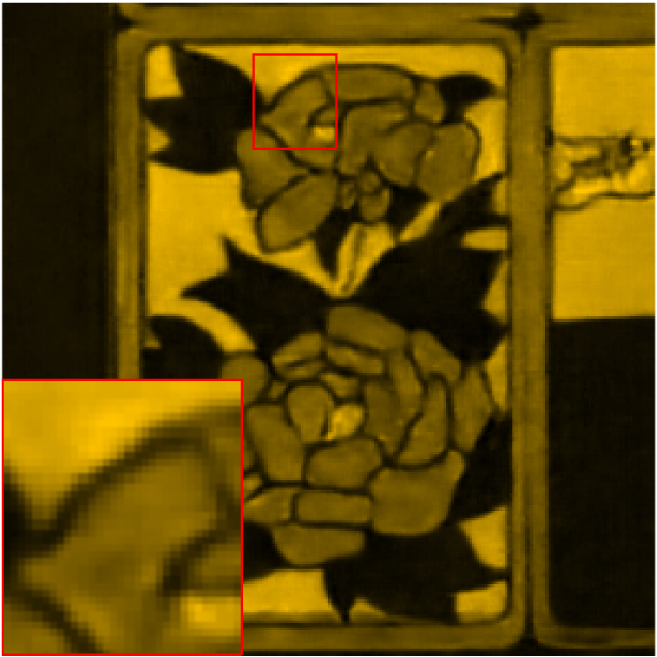}&
			\includegraphics[width=0.1\linewidth]{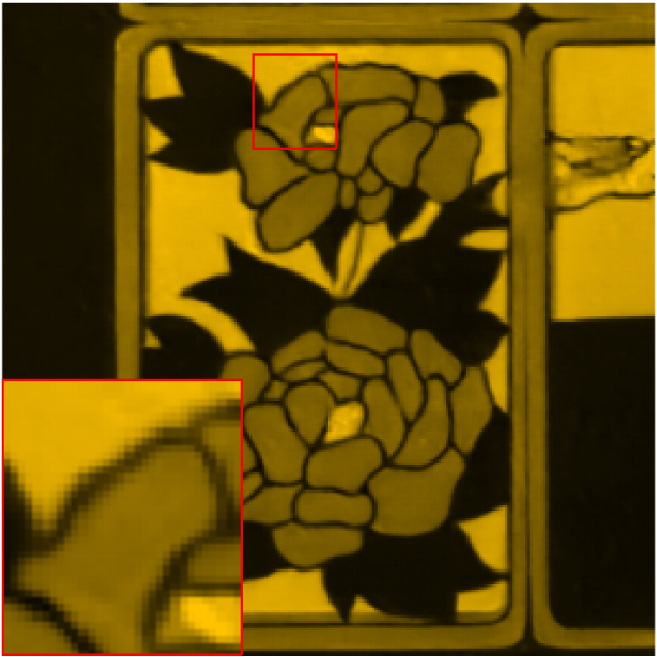}&
			\includegraphics[width=0.1\linewidth]{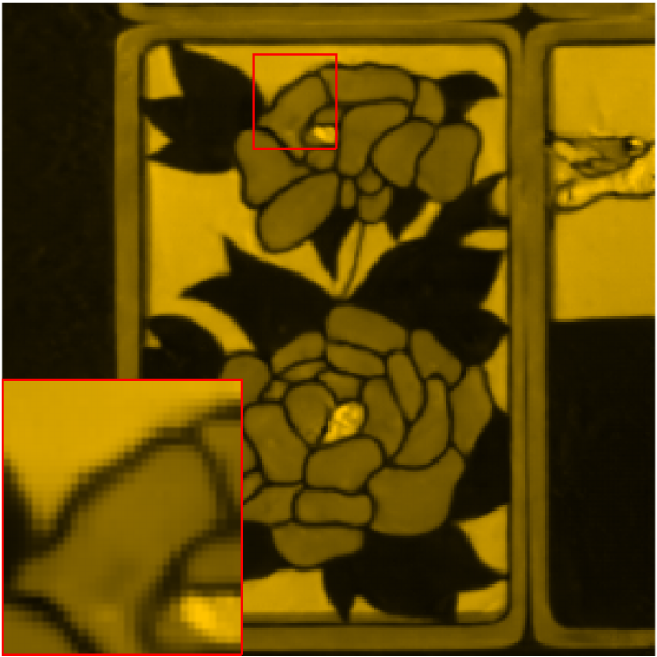}&
			\includegraphics[width=0.1\linewidth]{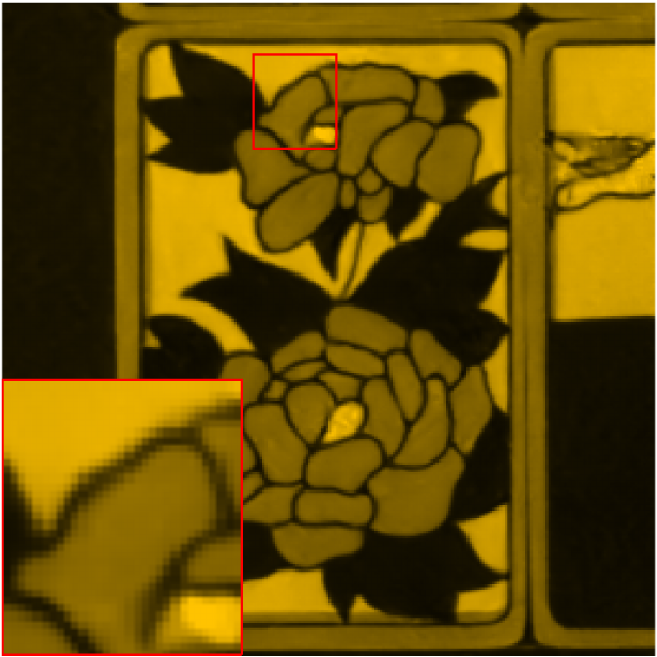}&
			\includegraphics[width=0.1\linewidth]{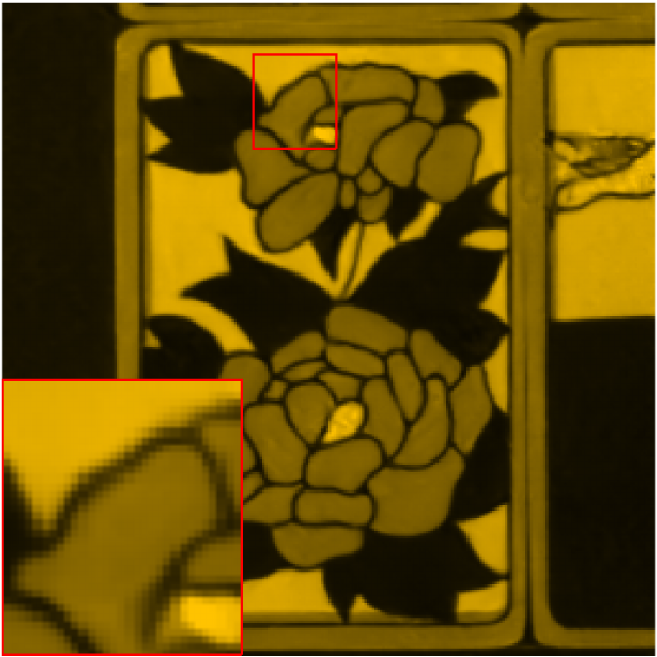}\\
            \includegraphics[width=0.1\linewidth]{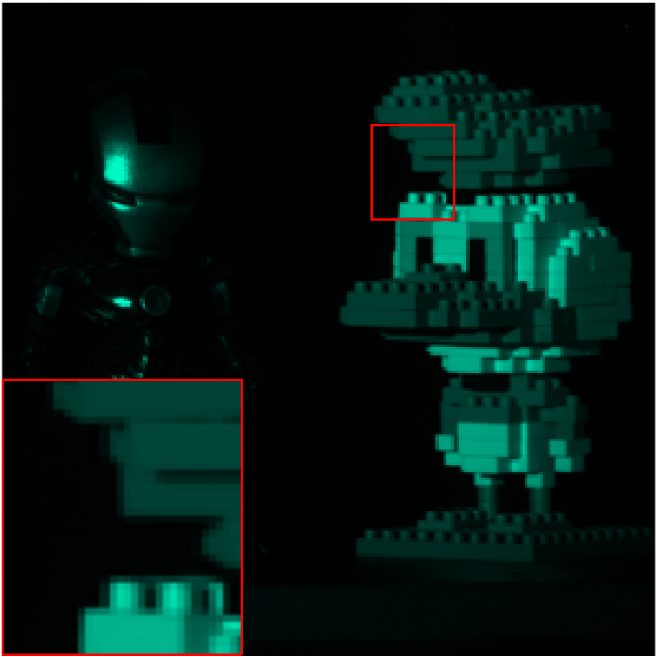}&
			\includegraphics[width=0.1\linewidth]{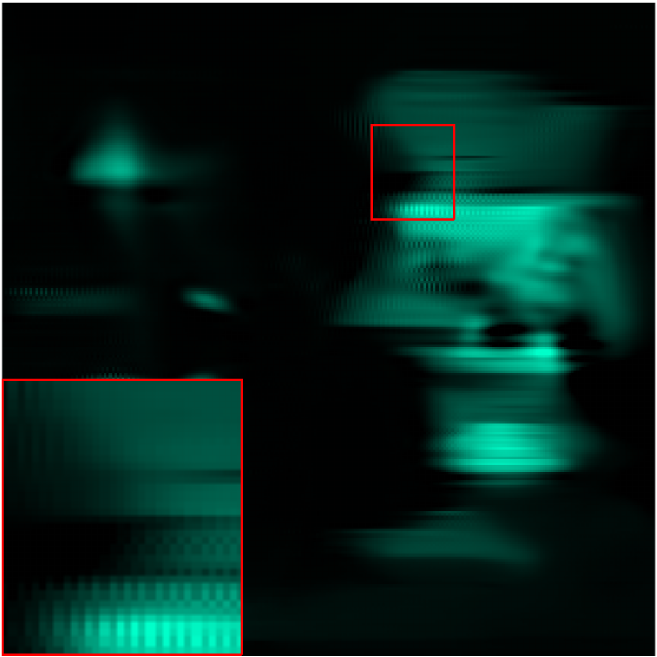}&
			\includegraphics[width=0.1\linewidth]{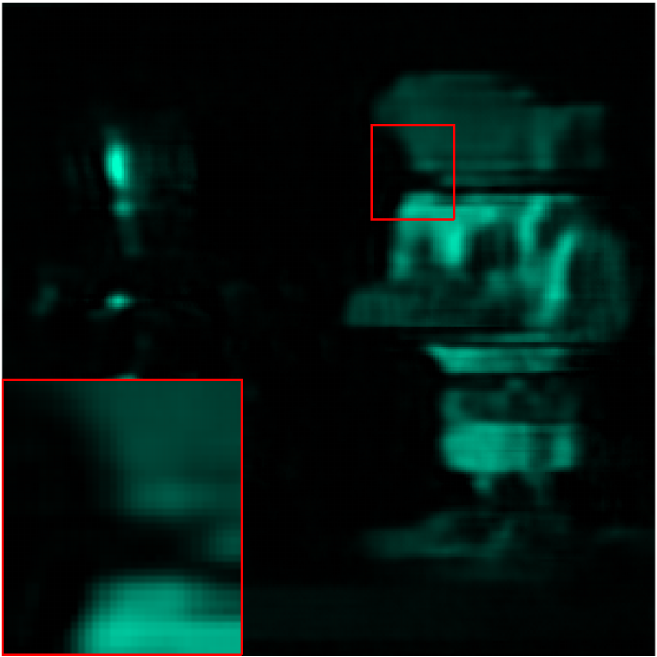}&
			\includegraphics[width=0.1\linewidth]{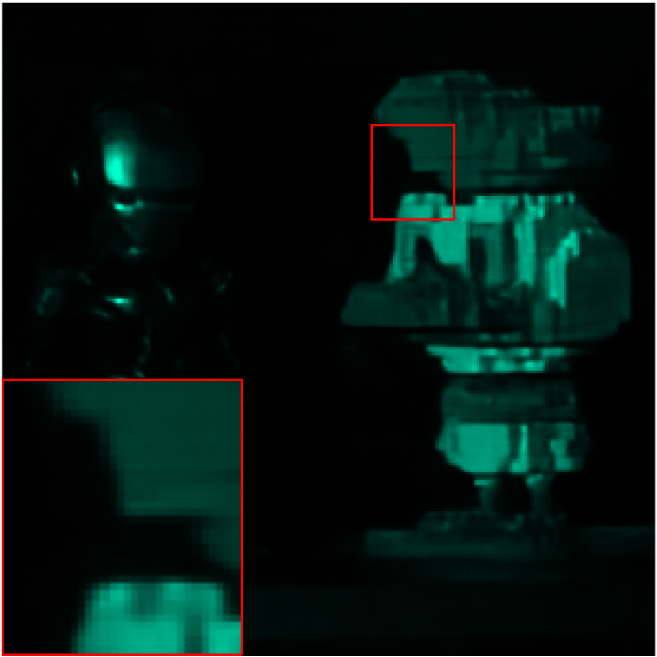}&
			\includegraphics[width=0.1\linewidth]{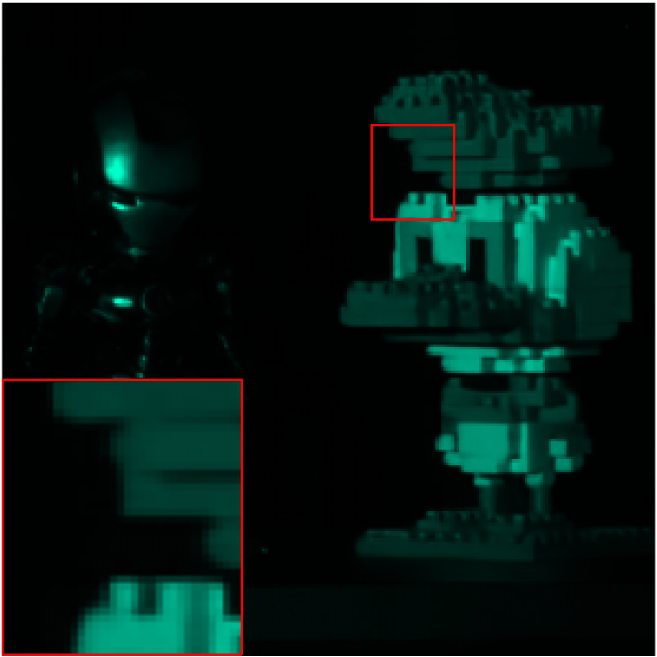}&
                \includegraphics[width=0.1\linewidth]{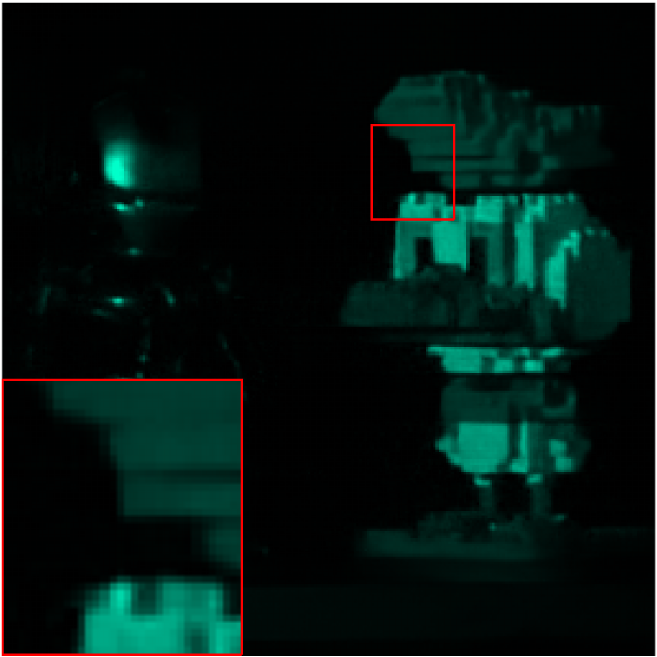}&
			\includegraphics[width=0.1\linewidth]{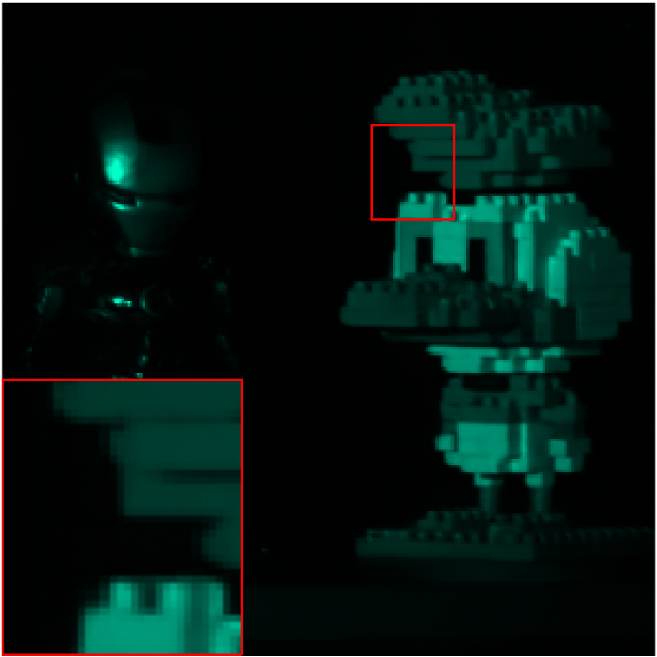}&
			\includegraphics[width=0.1\linewidth]{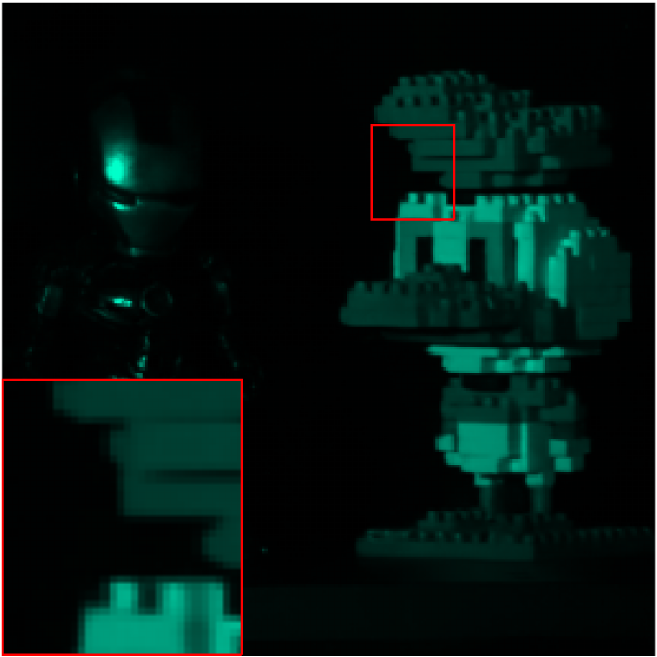}&
			\includegraphics[width=0.1\linewidth]{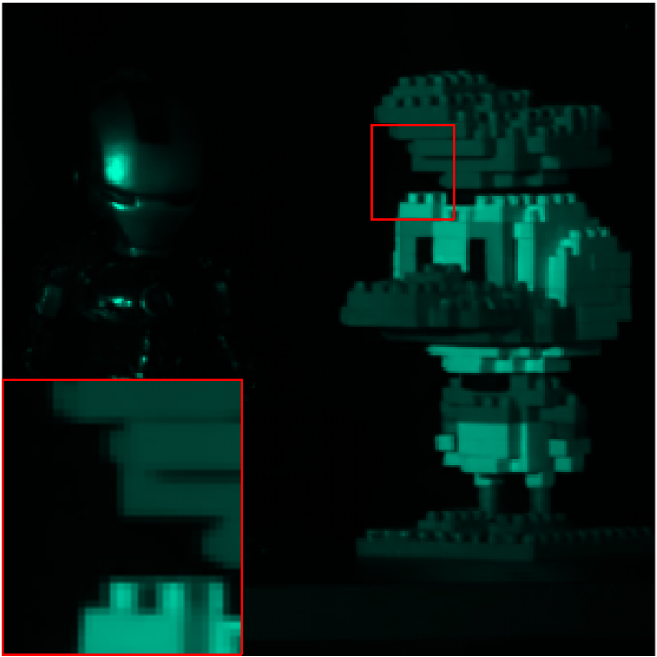}&
			\includegraphics[width=0.1\linewidth]{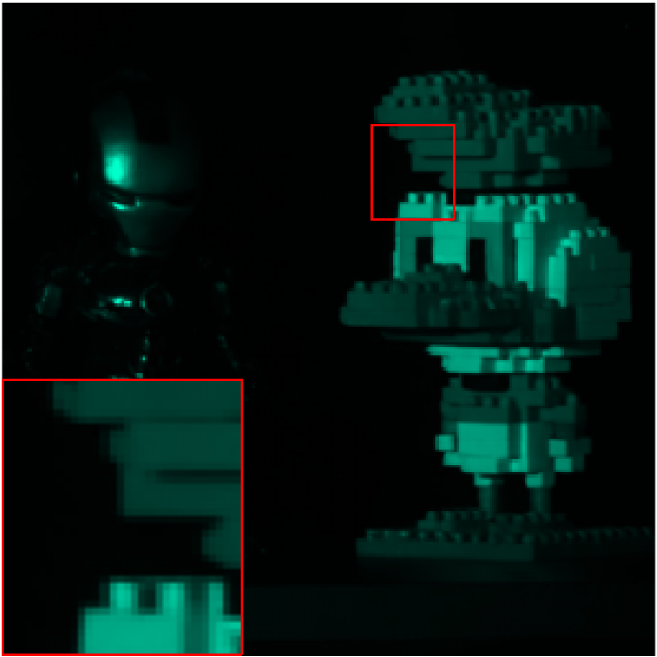}\\
            \includegraphics[width=0.1\linewidth]{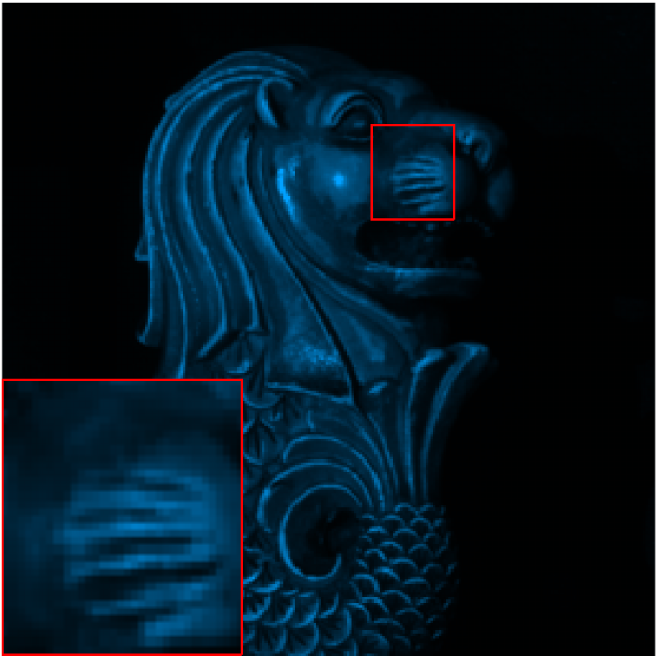}&
			\includegraphics[width=0.1\linewidth]{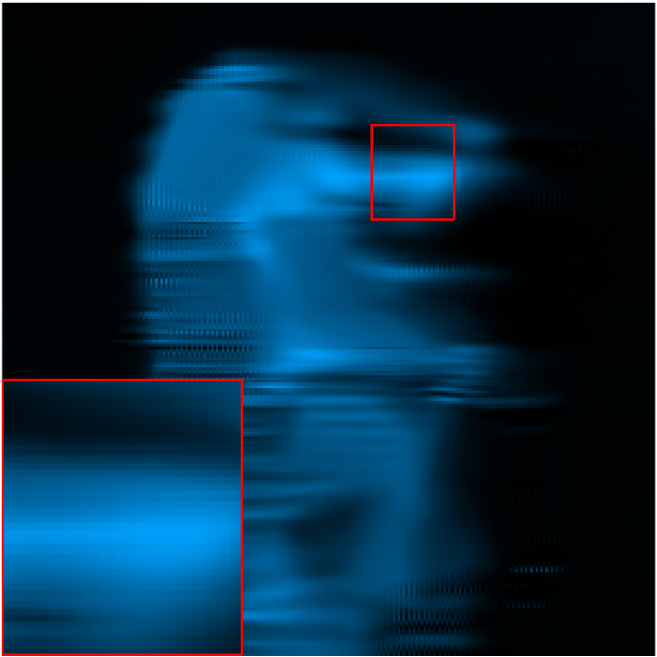}&
			\includegraphics[width=0.1\linewidth]{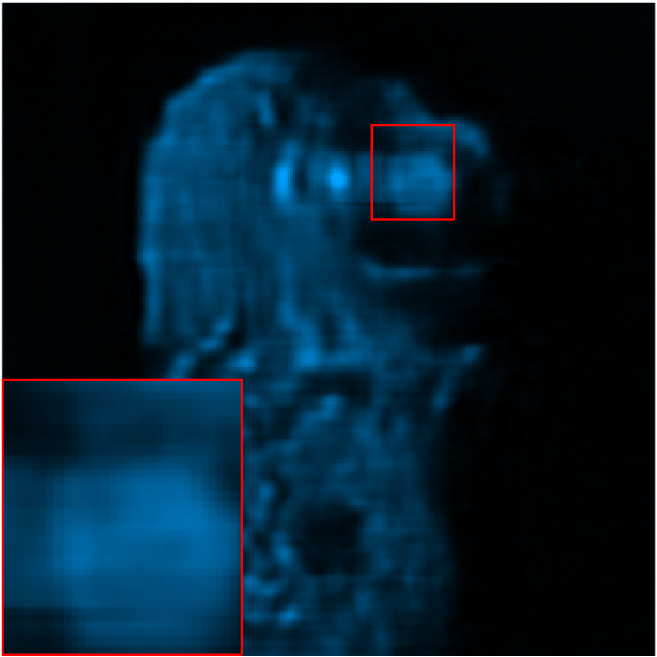}&
			\includegraphics[width=0.1\linewidth]{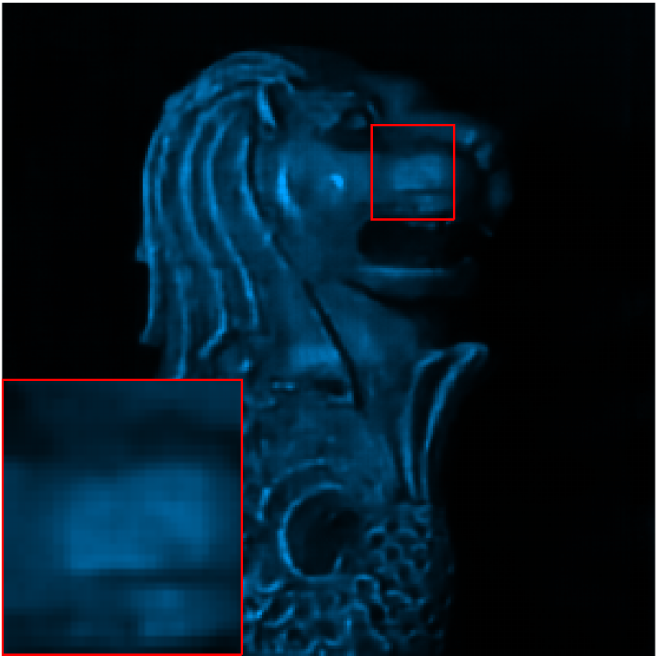}&
			\includegraphics[width=0.1\linewidth]{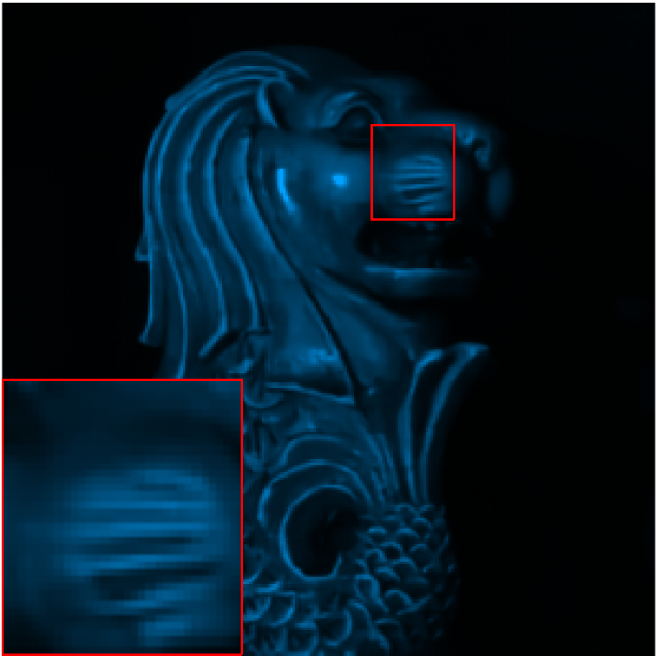}&
                \includegraphics[width=0.1\linewidth]{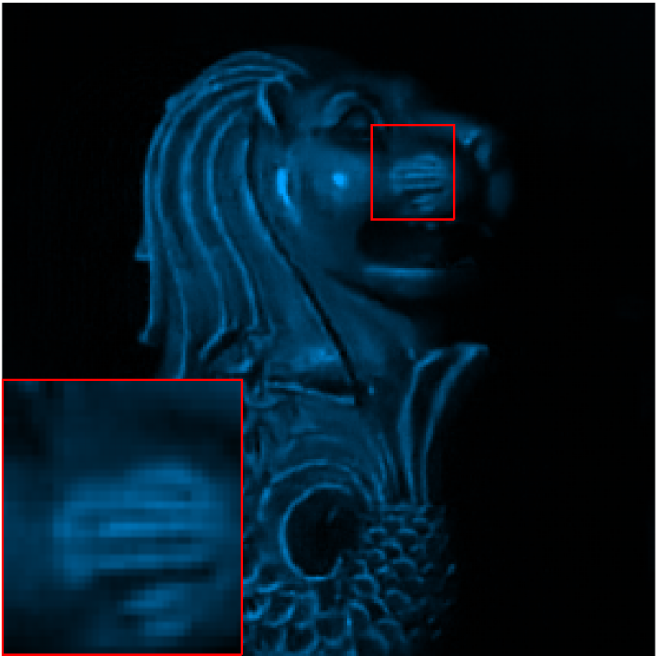}&
			\includegraphics[width=0.1\linewidth]{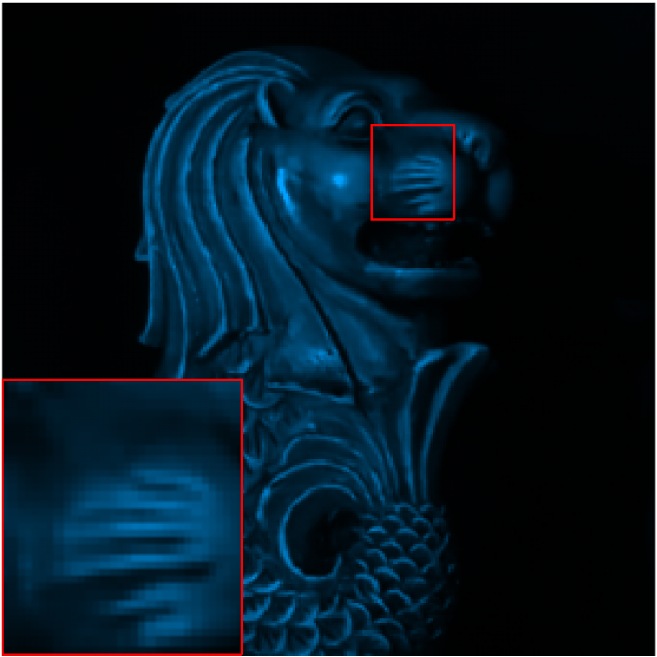}&
			\includegraphics[width=0.1\linewidth]{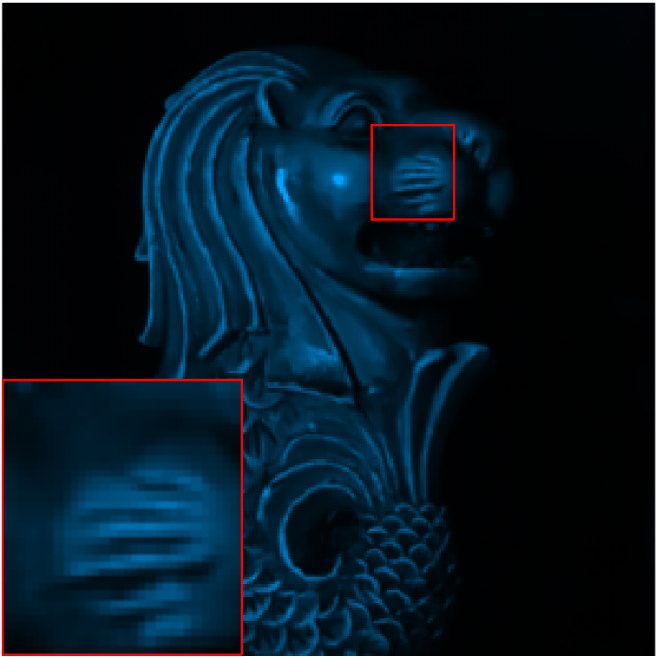}&
			\includegraphics[width=0.1\linewidth]{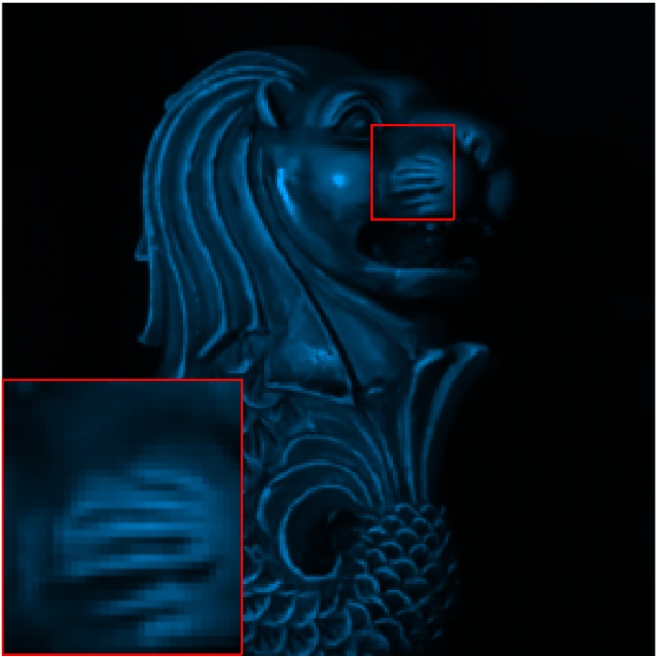}&
			\includegraphics[width=0.1\linewidth]{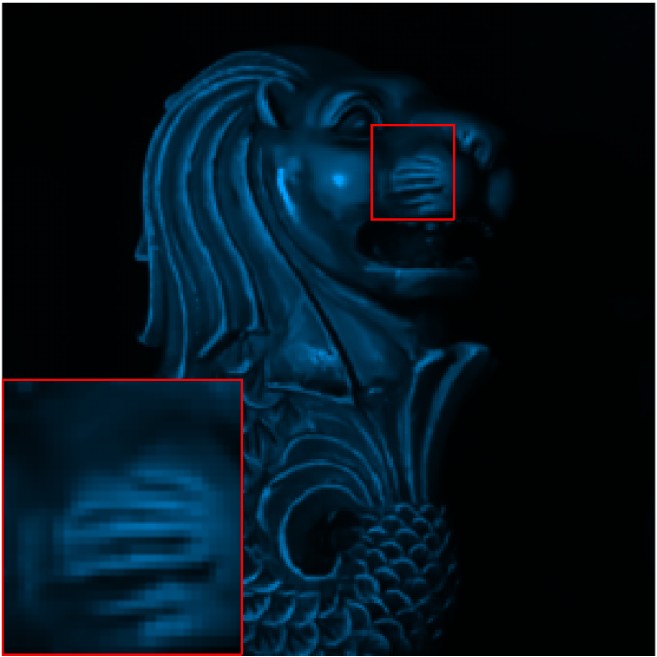}\\
            Truth&DeSCI&$\lambda$-Net&TSA-Net&BIRNAT&GAP-Net&PSRSCI&RCUMP&Block-9&LoRun-9
		\end{tabular}} 
		\caption{CASSI results of different scenes in different bands, from top to bottom $\lambda \in \{\SI{487.0}{\nano\meter}, \SI{498.0}{\nano\meter}, \SI{584.5}{\nano\meter}, \SI{492.5}{\nano\meter}, \SI{476.5}{\nano\meter}\}$.}
		\label{cassi_dimgs}
	\end{center}
    \vspace{-6mm}
\end{figure*}

\begin{figure*}
	\footnotesize
	\setlength{\tabcolsep}{1pt}
 \newcommand{\tabincell}[2]{\begin{tabular}{@{}#1@{}}#2\end{tabular}}
	\begin{center}
            \scalebox{0.94}{
		\begin{tabular}{cccccccccc}
			\includegraphics[width=0.1\linewidth]{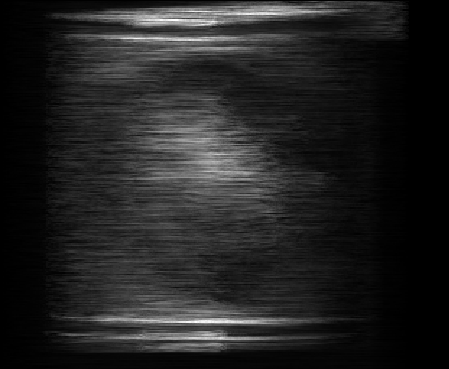}&
			\includegraphics[width=0.1\linewidth]{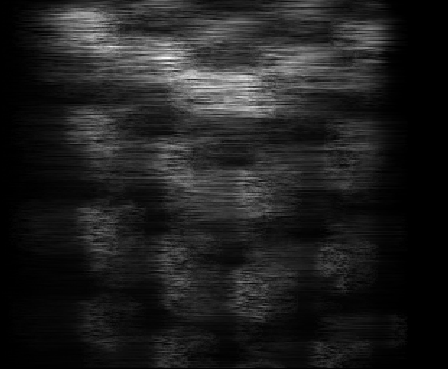}&
			\includegraphics[width=0.1\linewidth]{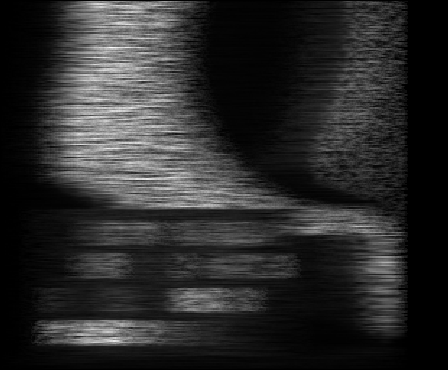}&
			\includegraphics[width=0.1\linewidth]{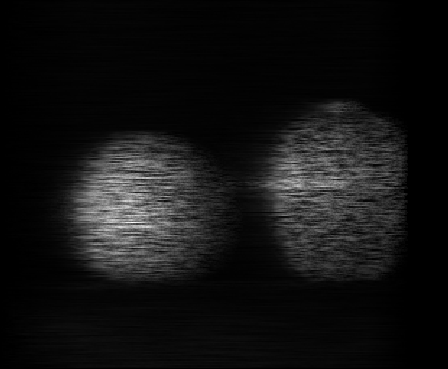}&
			\includegraphics[width=0.1\linewidth]{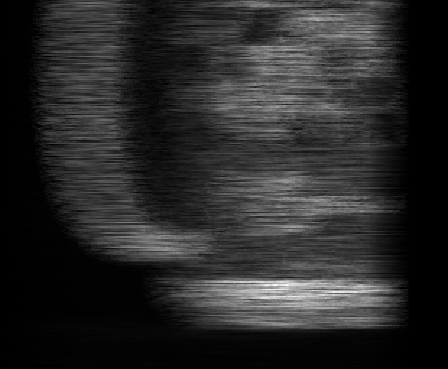}&
                \includegraphics[width=0.1\linewidth]{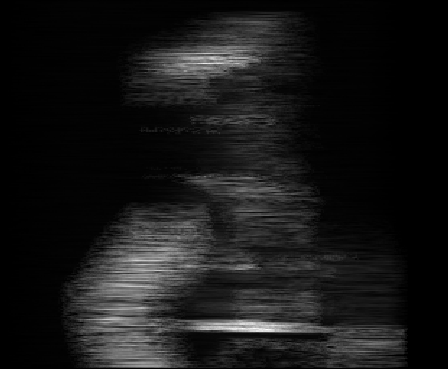}&
			\includegraphics[width=0.1\linewidth]{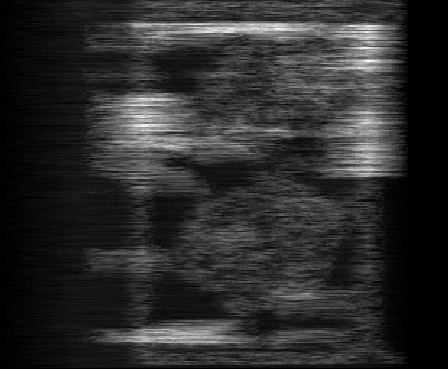}&
			\includegraphics[width=0.1\linewidth]{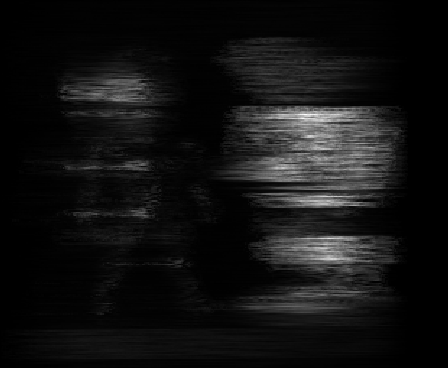}&
			\includegraphics[width=0.1\linewidth]{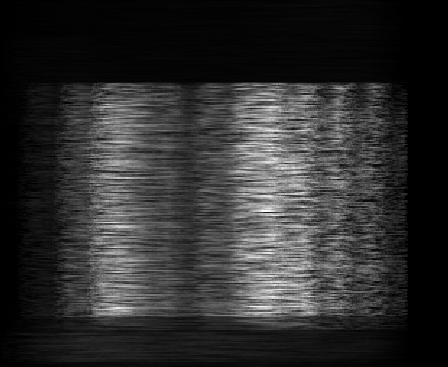}&
			\includegraphics[width=0.1\linewidth]{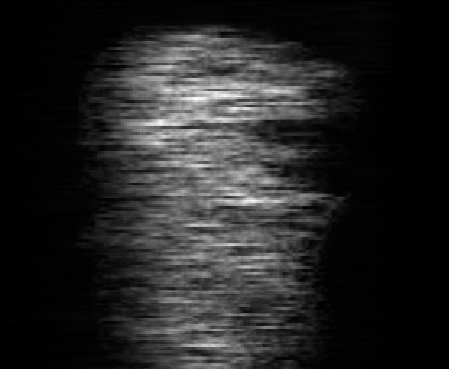}\\
            scene1&scene2&scene3&scene4&scene5&scene6&scene7&scene8&scene9&scene10
		\end{tabular}} 
		\caption{CASSI measurements of different scenes, each scene has the size of $256 \times 310$.}
		\label{cassi_meas}
	\end{center}
\end{figure*}

\begin{table}
    \centering
    \newcommand{\tabincell}[2]{\begin{tabular}{@{}#1@{}}#2\end{tabular}}
    \setlength{\tabcolsep}{3pt}
    \caption{Comparisons on CASSI datasets with U-Net denoiser. PSNR (upper entry), SSIM (lower entry) and parameters are reported.}
    \vspace{-2mm}
    \scalebox{0.9}{
    \begin{tabular}{cccccccc}
    \toprule
        Method & scene01 & scene02 & scene03 & scene04 & scene05 & Avg & Params.\\
    \midrule
        Block-9 & 
        \tabincell{c}{\textbf{34.87}\\\textbf{0.94}} &
        \tabincell{c}{\textbf{35.32}\\\textbf{0.94}}& 
        \tabincell{c}{\textbf{35.48}\\\textbf{0.95}}&
        \tabincell{c}{40.55\\0.97}&
        \tabincell{c}{\textbf{32.74}\\\textbf{0.95}}&
        \tabincell{c}{\textbf{35.79}\\\textbf{0.95}}&
        \tabincell{c}{4.80M}\\
    \midrule
        \tabincell{c}{LoRun-9\\$\gamma$=10}&
        \tabincell{c}{34.20\\0.93}&
        \tabincell{c}{34.40\\0.92}&
        \tabincell{c}{35.23\\\textbf{0.95}}&
        \tabincell{c}{\textbf{41.26}\\\textbf{0.98}}&
        \tabincell{c}{32.15\\0.94}&
        \tabincell{c}{35.45\\0.94}&
        \tabincell{c}{1.43M}\\
    \toprule
    \end{tabular}
    }
    \label{tab:snapshot_unet_s1-5}
    \vspace{-6mm}
\end{table}

\vspace{-2mm}
\subsection{Image Super Resolution}

\noindent\textbf{Background.}
SR aims to reconstruct a high-resolution image $\mathbf{x} \in \mathbb{R}^{m}$ from its low-resolution observation $\mathbf{y} \in \mathbb{R}^{n}$.
The degradation process is typically modeled as
\begin{equation}
\label{degradation}
\mathbf{y} = (\mathbf{x} \ast \mathbf{k}) \downarrow_s + \mathbf{n},
\end{equation}
where $\ast$ denotes convolution with blur kernel $\mathbf{k}$, $\downarrow_s$ represents downsampling with Scale Factor (SF) $s$, and $\mathbf{n}$ is additive noise.
Alternatively, Eq. \eqref{degradation} can be rewritten in matrix form:
\begin{equation}
\label{degradation_matrix}
\mathbf{y} = \mathbf{D} \mathbf{H} \mathbf{x} + \mathbf{n},
\end{equation}
where $\mathbf{H} \in \mathbb{R}^{m \times m}$ denotes the blur operator and $\mathbf{D} \in \mathbb{R}^{n \times m}$ denotes the downsampling matrix (thus $\Phi(\cdot)$ in Eq. \eqref{model} and Eq. \eqref{optimization} is the $\mathbf{D} \mathbf{H}$ operator).
Based on the iterative reconstruction scheme, we unfold the model following Eq. \eqref{gradient_descent} to Eq. \eqref{proximal_mapping_simple}, allowing interpretable learning of the mapping from $\mathbf{y}$ to $\mathbf{x}$.

\begin{figure*}
	\footnotesize
	\setlength{\tabcolsep}{1pt}
 \newcommand{\tabincell}[2]{\begin{tabular}{@{}#1@{}}#2\end{tabular}}
	\begin{center}
            \scalebox{0.97}{
		\begin{tabular}{cccccccccccc}
			\includegraphics[width=0.08\textwidth]{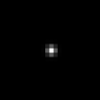}&
			\includegraphics[width=0.08\textwidth]{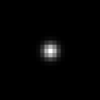}&
			\includegraphics[width=0.08\textwidth]{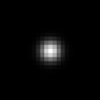}&
			\includegraphics[width=0.08\textwidth]{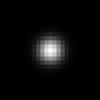}&
			\includegraphics[width=0.08\textwidth]{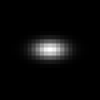}&
                \includegraphics[width=0.08\textwidth]{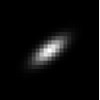}&
			\includegraphics[width=0.08\textwidth]{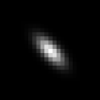}&
			\includegraphics[width=0.08\textwidth]{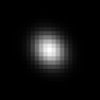}&
			\includegraphics[width=0.08\textwidth]{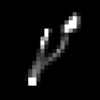}&
			\includegraphics[width=0.08\textwidth]{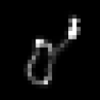}&
                \includegraphics[width=0.08\textwidth]{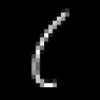}&
			\includegraphics[width=0.08\textwidth]{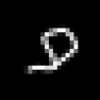}\\
            \textbf{K1}& \textbf{K2}&\textbf{K3}&\textbf{K4}&\textbf{K5}&\textbf{K6}&\textbf{K7}&\textbf{K8}&\textbf{K9}&\textbf{K10}&\textbf{K11}&\textbf{K12}\\
		\end{tabular}} 
		\caption{Presentation of 4 isotropic Gaussian kernels with different widths (0.7, 1.2, 1.6, 2.0), 4 anisotropic Gaussian kernels and 4 motion blur kernels.}
		\label{kernel}
	\end{center}
    \vspace{0mm}
\end{figure*}

\noindent\textbf{Experiment Settings and Datasets.}
In the experiment, we set the SF as $\{ \times 2, \times3, \times4 \}$ for 12 different blur kernels (4 isotropic Gaussian kernels with different widths $\{0.7, 1.2, 1.6, 2.0\}$, 4 anisotropic Gaussian kernels and 4 motion blur kernels which are presented in Fig. \ref{kernel}).
\emph{DIV2K} and \emph{Flickr2K} datasets are adopted as training samples.
And \emph{BSD68} and \emph{Set5} datasets are used to test the performance.

\noindent\textbf{Comparisons and Results.}
We compare the Block-9 strategy with our LoRun method in Table \ref{tab:sr} under different SFs and blur kernels on \emph{BSD68} datasets.
Our LoRun has reduced $63\%$ parameters of Block-9 while obtained better SR results in many kernels $\{\mathbf{K1, K2, K3, K4, K5, K6, K8}\}$, which shows that LoRun has the potential to be more generalized to adapt different degradation situations.
Fig. \ref{sr_compare} visualizes the SR reconstruction results of Block-1, Block-9 and our LoRun-9 on $\mathbf{K5, K9}$ blur kernels with different SFs, which demonstrate that LoRun framework has better visual performance than current DUN training methods.

\begin{table*}
	\caption{Comparisons between our LoRun ($\gamma = 10$) and Block-9 for \emph{BSD68} datasets \textbf{SR} in different SFs and different blur kernels.}
	\centering
  \newcommand{\tabincell}[2]{\begin{tabular}{@{}#1@{}}#2\end{tabular}}
	\resizebox{0.97\textwidth}{!}{
        \centering
        \begin{tabular}{ccccccccccccccc}
        \bottomrule
   Method & SF & \textbf{K1} &\textbf{K2}&\textbf{K3}&\textbf{K4}&\textbf{K5}&\textbf{K6}&\textbf{K7}&\textbf{K8}&\textbf{K9}&\textbf{K10}&\textbf{K11}&\textbf{K12}&Params.\\
        \midrule
	\multirow{2}[2]{*}{Block-9}&$\times 2$&\textbf{29.90}&\textbf{30.31}&\textbf{29.79}&28.70&28.41&\textbf{28.29}&\textbf{28.49}&\textbf{27.63}&\textbf{29.57}&\textbf{29.49}&\textbf{29.54}&\textbf{29.52}&\multirow{2}[2]{*}{5.3M}\\
                & $\times 3$&
26.70&27.37&27.51&27.46&27.27&27.21&\textbf{27.30}&\textbf{27.10}&\textbf{26.89}&\textbf{26.78}&\textbf{26.68}&\textbf{26.52}&\\
                & $\times 4$&
\textbf{24.75}&25.60&25.84&25.94&25.86&25.75&\textbf{25.79}&25.95&\textbf{25.18}&\textbf{24.92}&\textbf{24.97}&\textbf{24.96}&\\

    \midrule
			    \multirow{2}[2]{*}{\makecell{LoRun-9 \\ $\gamma = 10$}}
                & $\times 2$&
29.88&30.17&29.73&\textbf{28.72}&\textbf{28.45}&28.25&28.42&27.60&29.03&29.31&29.43&29.43&\multirow{2}[2]{*}{\textbf{1.96M}}\\
                & $\times 3$&
\textbf{26.73}&\textbf{27.38}&\textbf{27.52}&\textbf{27.47}&\textbf{27.29}&\textbf{27.25}&27.22&27.02&26.65&26.68&26.43&26.33&\\
                & $\times 4$&
24.74&\textbf{25.65}&\textbf{25.89}&\textbf{25.99}&\textbf{25.90}&\textbf{25.80}&25.75&\textbf{25.99}&25.05&24.82&24.72&24.83&\\
			\toprule
		\end{tabular}
	}
	\label{tab:sr}
    \vspace{-2mm}
\end{table*}

\begin{figure*}
	\footnotesize
	\setlength{\tabcolsep}{1pt}
 \newcommand{\tabincell}[2]{\begin{tabular}{@{}#1@{}}#2\end{tabular}}
	\begin{center}
            \scalebox{1}{
		\begin{tabular}{cccccc}
			\includegraphics[width=0.19\linewidth]{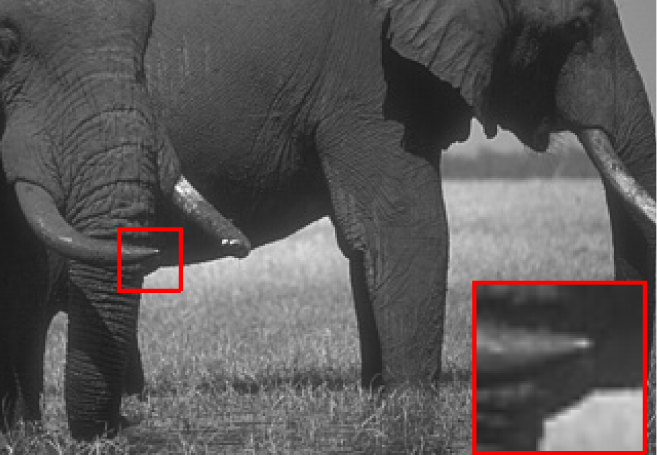}&
			\includegraphics[width=0.19\linewidth]{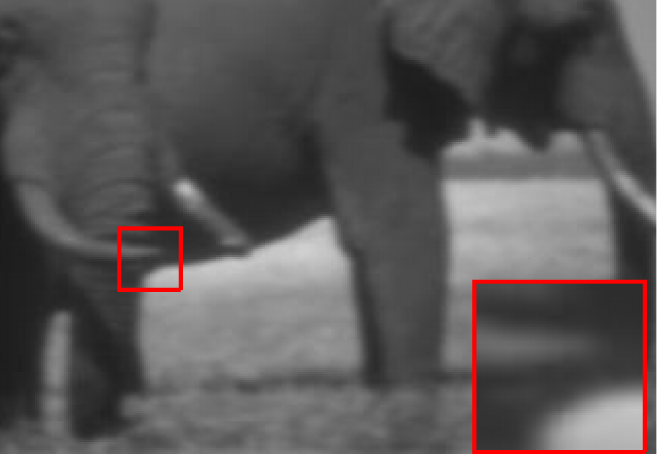}&
			\includegraphics[width=0.19\linewidth]{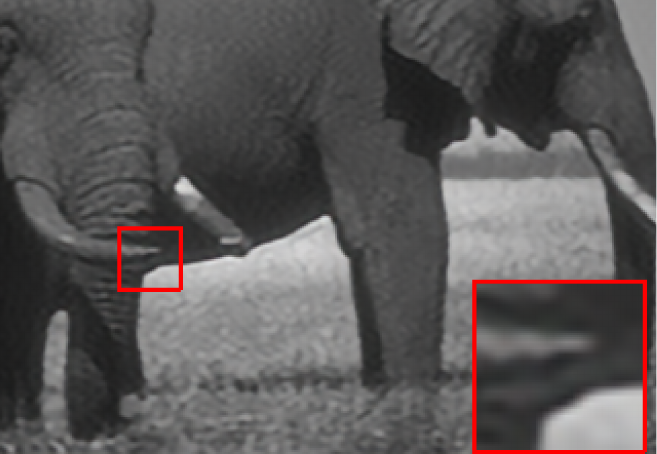}&
			\includegraphics[width=0.19\linewidth]{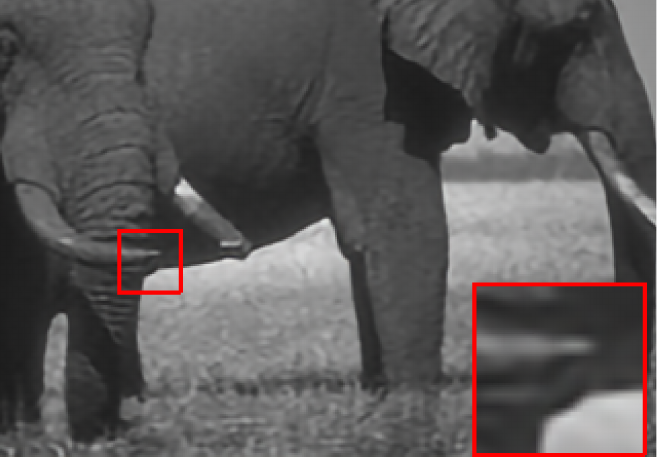}&
                \includegraphics[width=0.19\linewidth]{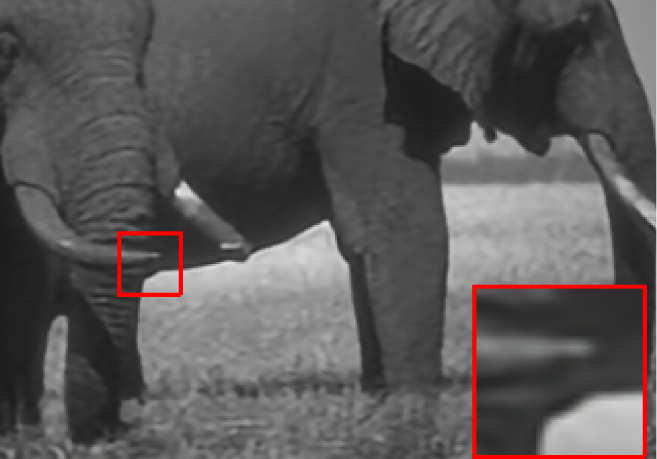}\\
			\includegraphics[width=0.19\linewidth]{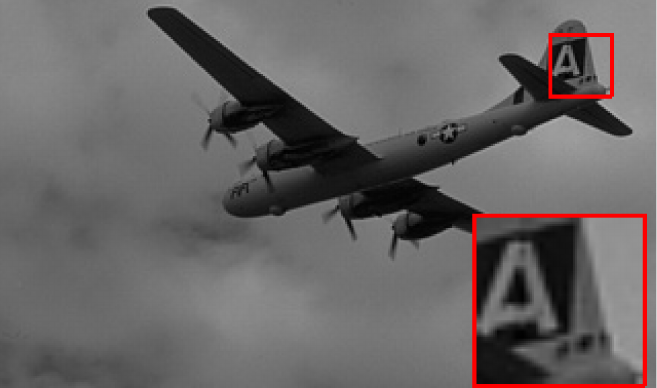}&
			\includegraphics[width=0.19\linewidth]{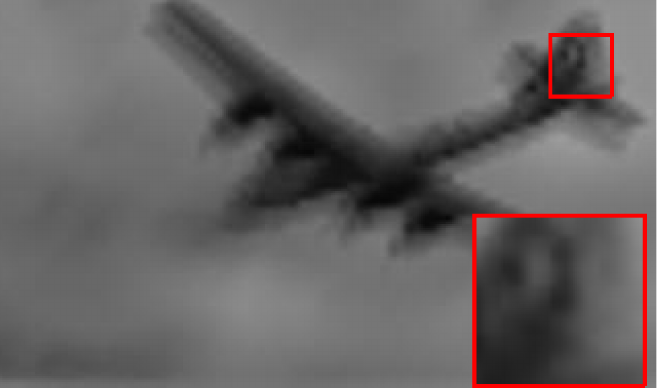}&
			\includegraphics[width=0.19\linewidth]{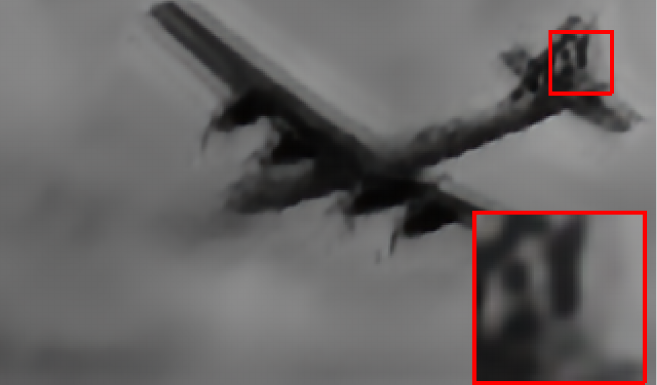}&
                \includegraphics[width=0.19\linewidth]{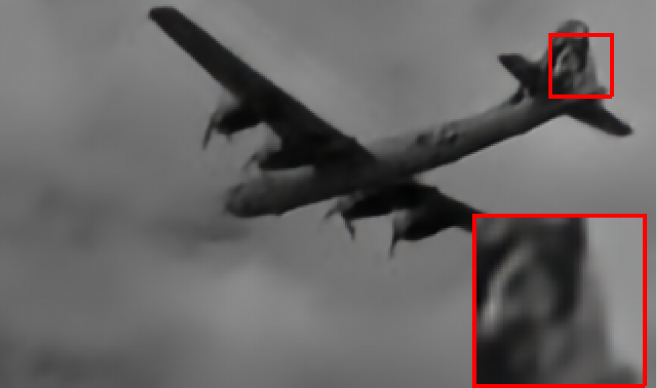}&
			\includegraphics[width=0.19\linewidth]{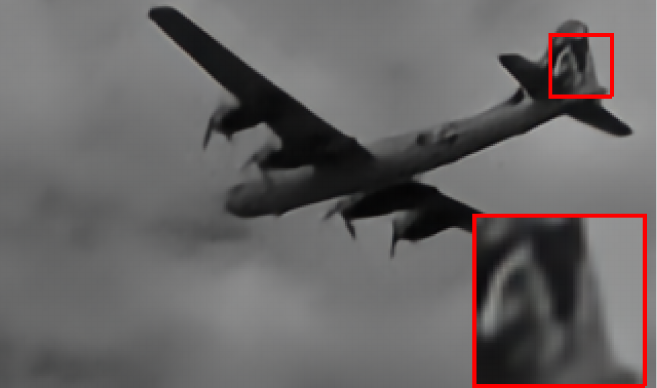}\\

            Truth&Zoomed-LR&Block-1&Block-9&LoRun-9\\
		\end{tabular}}
		\caption{SR results of different blur kernels and SFs. (Up: \textbf{K5}, SF=$\times 2$. Down: \textbf{K9}, SF=$\times 4$.)}
        \vspace{-4mm}
		\label{sr_compare}
	\end{center}
\end{figure*}

\subsection{Ablation Analysis}
In this section, we provide the ablation analysis of our LoRun framework on below four aspects: 
(1) the effect of the frozen denoiser (2) the effect of the rank factor $\gamma$ (3) the effect of the GDM and (4) the effect of the pre-training procedure.
Relevant analyses and explanations are presented, and their efficiency is experimentally demonstrated.

\noindent\textbf{Effect of the Frozen Denoiser.}
The frozen denoiser is the backbone of our LoRun.
And the information quality and denoising capability of the base module are important for the update of the LoRA module.
We test the effect of the basic frozen denoiser in CS application, and U-Net serves as the denoiser.
Generally, in CS a DUN model should be trained for each CS ratio individually.
As the CS ratio increases, the denoiser in the model can obtain more information and achieve better performance.
In this ablation experiment, we train our LoRun framework by incorporating the frozen backbone, which is a U-Net denoiser pretrained under different CS ratios (denoted as $\beta$-pretrained).
Then we product the LoRA fine-tuning procedure.
The numerical results (PSNR and SSIM) with different $beta$-pretrained denoisers are demonstrated in Table.\ref{tab:cs_unet_pretrainedRatio_ablation}.
And for each CS ratio, our LoRun models induced with pretrained denoiser under larger CS ratio exhibits better performance.
Thus a more suitable pretrained benchmark denoiser can be able to better guide the optimization direction of the LoRA parameter updates, then perform better multi-level denoising abilities.
Moreover, this experiments show that even we adopt the denoiser which was trained under CS ratio $\beta=1\%$, our LoRun can also obtain comparable results.
Therefore, we can decouple the backbone from LoRA, enabling efficient and flexible switching between different task settings by simply replacing the LoRA modules.

\begin{table}
    \centering
     \caption{Ablation experiments on $\beta$-pretrained based on the \textbf{U-Net} denoiser in \textbf{CS}.}
     \setlength{\tabcolsep}{3pt}
    \newcommand{\tabincell}[2]{\begin{tabular}{@{}#1@{}}#2\end{tabular}}
    \scalebox{0.87}{
    \begin{tabular}{cccccccc}
    \toprule
        Method & $\beta$=1\% & $\beta$=4\% & $\beta$=10\% & $\beta$=25\% & $\beta$=30\% & $\beta$=40\% & $\beta$=50\%\\
    \midrule
        \tabincell{c}{LoRun-9\\$\beta$-pretrained=1\%} & 
        \tabincell{c}{21.07\\\textbf{0.55}}& 
        \tabincell{c}{25.41\\\textbf{0.79}}&
        \tabincell{c}{29.15\\\textbf{0.89}}&
        \tabincell{c}{34.47\\0.95}&
        \tabincell{c}{35.72\\\textbf{0.96}}&
        \tabincell{c}{38.09\\0.97}&
        \tabincell{c}{40.26\\\textbf{0.98}}\\
    \midrule
        \tabincell{c}{LoRun-9\\$\beta$-pretrained=50\%}&
        \tabincell{c}{\textbf{21.08}\\\textbf{0.55}}&
        \tabincell{c}{\textbf{25.66}\\\textbf{0.79}}&
        \tabincell{c}{\textbf{29.57}\\\textbf{0.89}}&
        \tabincell{c}{\textbf{35.07}\\\textbf{0.96}}&
        \tabincell{c}{\textbf{36.31}\\\textbf{0.96}}&
        \tabincell{c}{\textbf{38.65}\\\textbf{0.98}}&
        \tabincell{c}{\textbf{40.58}\\\textbf{0.98}}\\
    \toprule
    \end{tabular}}
    \vspace{-3mm}
\label{tab:cs_unet_pretrainedRatio_ablation}
    
\end{table}

\noindent\textbf{Effect of the Rank Factor $\gamma$.}
Parameter $\gamma$ controls the rank of the weights in LoRA modules as shown in Eq. \eqref{r_gamma}.
Table \ref{tab:cs_transformer_r_ablation} shows the compressive sensing results of the proposed LoRun model with Transformer denoiser.
As the values of $\gamma$ rises, the rank of the weights also increases with more parameters.
When $\gamma$ is low, few parameters may not be able to tune the basis denoiser to a good level and when $\gamma$ is too high, the weights are not well guaranteed to be low-rank, which leads to redundant parameters and sub-optimal performance.

\begin{table}[!t]
    \centering
    \caption{Ablation on $\gamma$ based on the \textbf{Transformer} denoiser in \textbf{CS}.}
    \setlength{\tabcolsep}{3pt}
    \newcommand{\tabincell}[2]{\begin{tabular}{@{}#1@{}}#2\end{tabular}}
    \scalebox{1}{
    \begin{tabular}{cccccccc}
    \toprule
        Method & $\beta$=1\% & $\beta$=4\% & $\beta$=10\% & $\beta$=25\% & $\beta$=30\% & $\beta$=40\% & $\beta$=50\%\\
    \midrule
        \tabincell{c}{LoRun-9\\$\gamma$=10}&
        \tabincell{c}{21.02\\\textbf{0.55}}&
        \tabincell{c}{\textbf{25.17}\\\textbf{0.75}}&
        \tabincell{c}{\textbf{29.14}\\\textbf{0.89}}&
        \tabincell{c}{34.16\\\textbf{0.95}}&
        \tabincell{c}{35.54\\\textbf{0.95}}&
        \tabincell{c}{37.88\\\textbf{0.97}}&
        \tabincell{c}{40.10\\\textbf{0.98}}\\
    \midrule
        \tabincell{c}{LoRun-9\\$\gamma$=20}&
        \tabincell{c}{\textbf{21.08}\\\textbf{0.55}}&
        \tabincell{c}{\textbf{25.17}\\\textbf{0.75}}&
        \tabincell{c}{29.13\\\textbf{0.89}}&
        \tabincell{c}{\textbf{34.24}\\\textbf{0.95}}&
        \tabincell{c}{\textbf{35.56}\\\textbf{0.95}}&
        \tabincell{c}{\textbf{37.95}\\\textbf{0.97}}&
        \tabincell{c}{\textbf{40.14}\\\textbf{0.98}}\\
    \midrule
        \tabincell{c}{LoRun-9\\$\gamma$=80}&
        \tabincell{c}{20.97\\\textbf{0.55}}&
        \tabincell{c}{25.01\\0.74}&
        \tabincell{c}{28.83\\0.88}&
        \tabincell{c}{33.93\\\textbf{0.95}}&
        \tabincell{c}{35.30\\\textbf{0.95}}&
        \tabincell{c}{37.74\\\textbf{0.97}}&
        \tabincell{c}{39.98\\\textbf{0.98}}\\
    \toprule
    \end{tabular}
    }
    \label{tab:cs_transformer_r_ablation}
    \vspace{-2mm}
\end{table}

\noindent\textbf{Effect of the GDM.}
In addition to the interpretability and robustness in the derivation of optimization theory, the GDM also matters in the degradation-specific learning tasks, which encodes the degradation matrix information into next PMM, introducing a guided denoising process.
We conduct experiments between Block-1 model and the version of Block-1 w/o GDM.
Table \ref{tab:sr_ablation} demonstrates the importance of GDM for SR with SF = $\times 3$ on \emph{Set5} and GDM brings 1.16 dB better performance on average in Block-1 setting, which demonstrates the degradation-specific advantages of GDM.
\begin{table}
    \centering
    \caption{Ablation analysis for GDM in SR with SF = $\times 3$ on \emph{Set5}.}
     \newcommand{\tabincell}[2]{\begin{tabular}{@{}#1@{}}#2\end{tabular}}
     \scalebox{1}{
    \begin{tabular}{cccccc}
\toprule
\textbf{Method} & \textbf{baby} &\textbf{bird}&\textbf{head}&\textbf{Average}&\textbf{Params.} \\
    \midrule
Block-1&\textbf{29.62}&\textbf{27.55}&\textbf{28.55}&\textbf{28.57}&0.59M\\
    \midrule
        \tabincell{c}{Block-1\\w/o GDM} &29.14&27.28&25.80&27.41&0.59M\\
\toprule
    \end{tabular}}
    \label{tab:sr_ablation}
    \vspace{-2mm}
\end{table}

\begin{figure}
    \centering
    \includegraphics[width=0.98\linewidth]{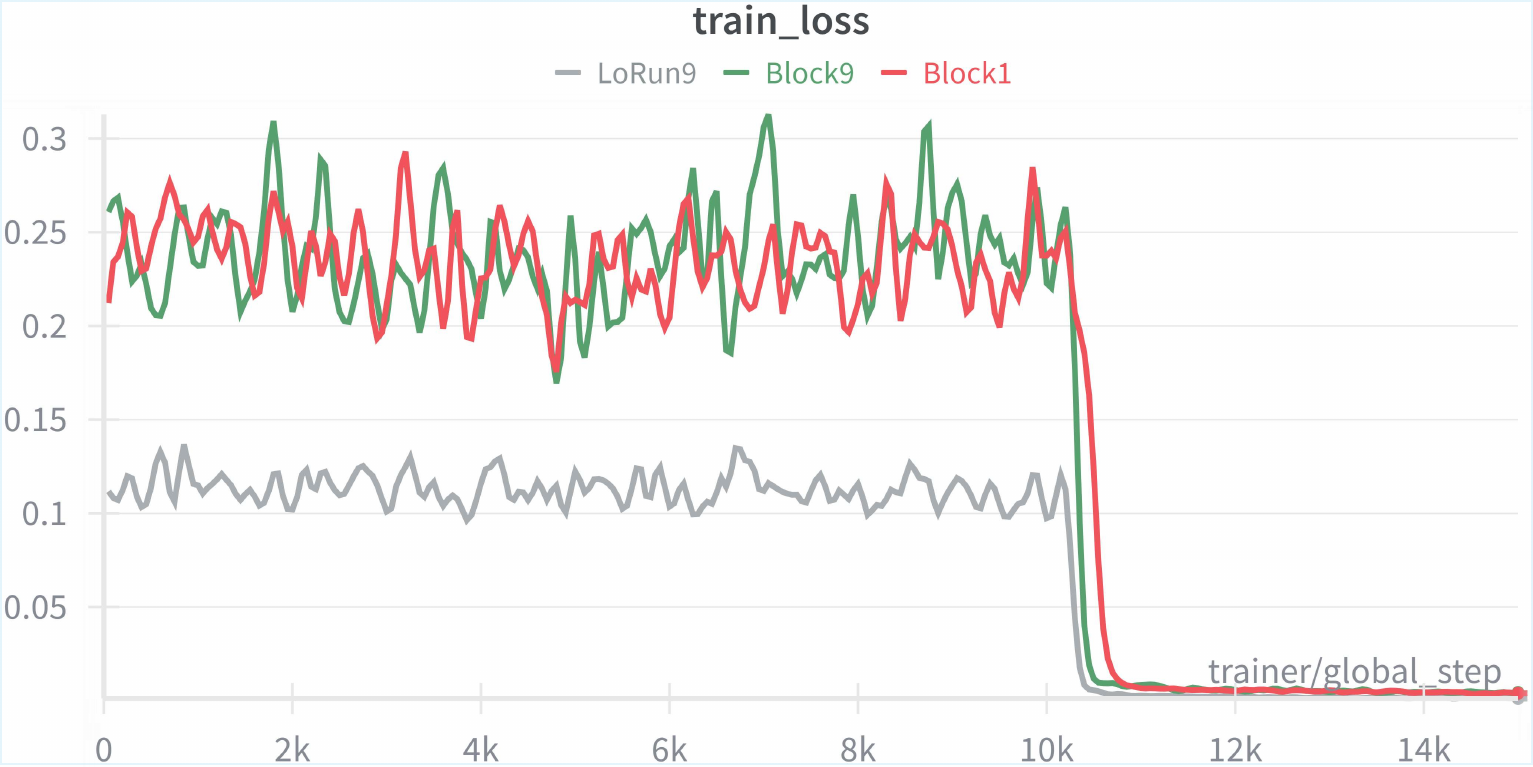}
    \vspace{-1mm}
    \caption{ Loss curves with training iterations among different training strategies.}
    \label{fig:loss_curves}
    \vspace{-2mm}
\end{figure}

\noindent\textbf{Effect of the Pre-training Procedure.}
Pre-training constructs the basic denoising abilities for the backbone denoiser, which is leveraged across stages.
This strategy not only reduces the redundant parameters for different blocks, but also obtains more stable and fast loss convergence as evidenced in Fig. \ref{fig:loss_curves}, which shows the relation between loss and training steps on CS task under $25\%$ ratio.
Noted that Block1 denotes the DUN framework with only one block, Block9 has 9 independent blocks, Block9-share represents one block within 9 stages and LoRun9-25 means our LoRun method with 9 stages under $25\%$ CS ratio.
In this figure, all the current full training strategies have more violent fluctuations and higher losses compared with our LoRun.
Moreover based on the inflection point of the loss in the figure, LoRun converges faster than all other strategies.

{
\noindent\textbf{Effect of the number $K$ of the Blocks.}
Table \ref{tab:blocks_ablation} shows the effect of the number $K$ of the DUN blocks.
Noted that when $K=1$, our LoRun just add the LoRA modules into each weight of the block.
The results demonstrate the performance trend of our LoRun with varying block numbers: larger blocks often present better results, while when the number of blocks is large enough the effect rises slowly.
Our LoRun can obtains comparable results against the regular Block-K models across the varying $K$ blocks.}

\begin{table}[ht]
\centering
\caption{ Ablation analysis for the effect of the number $K$ of blocks.}
\vspace{-1mm}
\scalebox{1}{
\begin{tabular}{lcccc}
\toprule
 & $K=1$ & $K=4$ & $K=9$ & $K=12$ \\
\hline
Original & 
\makecell{31.08 \\ 0.9032} & 
\makecell{32.28 \\ 0.9085} & 
\makecell{32.84 \\ 0.9128} & 
\makecell{32.91 \\ 0.9133} \\
\hline
LoRun (Ours) & 
\makecell{31.28 \\ 0.9062} & 
\makecell{32.24 \\ 0.9090} & 
\makecell{32.78 \\ 0.9125} & 
\makecell{32.81 \\ 0.9130} \\
\toprule
\end{tabular}}
\label{tab:blocks_ablation}
\vspace{-2mm}
\end{table}

\section{Conclusion}
\label{conclusion}
In this paper, we proposed a generalized deep low-rank adaptation unfolding network for IR, in which only one denoiser can rule all, compressing memory usage with no additional inference latency.
By injecting a pretrained module into each block of the DUN model, the backbone of the model is constructed with most frozen parameters.
Next, we integrate LoRA into the DUN model and enable dynamic adaptation for each denoiser with few independent parameters.
Finally under the guidance of the fundamental denoiser, we train the whole model end-to-end, which learns specific information for each block with low-rank property.
Comprehensive experiments in the three IR tasks verify the effectiveness of our method, which reduces redundant parameters and achieves promising results.

\bibliographystyle{IEEEtran}
\bibliography{main}

\clearpage
\appendix

In the supplemental materials, we provide the derivation of the PGD and HQS algorithms, LoRA visualization and the details of the denoisers adopted in three tasks, such as the U-Net, Transformer and the other complex structures which achieve SOTA performance on the target application.
We first provide the classical U-Net and Transformer frameworks, then we present the architectures according to the application.
{

\noindent\textbf{Derivation of the PGD and HQS Algorithms}
\textbf{PGD: } IR problems formulate the vectorized degraded measurement $\mathbf{y} \in \mathbb{R}^n$ and the vectorized original image $\mathbf{x} \in \mathbb{R}^m$ as  
\begin{equation}
    \mathbf{y} = \Phi \mathbf{x} + \mathbf{n},
\end{equation}
where $\mathbf{n} \in \mathbb{R}^n$ denotes the noise or errors in the imaging system, and $\Phi \in \mathbb{R}^{n \times m}$ represents the degradation matrix.  
The above equation can be reformulated as an optimization model:  
\begin{equation}
    \hat{\mathbf{x}} = \arg\min_{\mathbf{x}} \frac{1}{2} \left\| \mathbf{y} - \Phi \mathbf{x} \right\|_2^2 + \lambda \Psi(\mathbf{x}),
\end{equation}
where $\Psi(\cdot)$ denotes the regularization term which depends on the image prior assumed in the model.  
Based on the representation and the optimization model, PGD uses the following update steps to iteratively solve the IR problem:  
\begin{equation}
    \mathbf{z}^{(k)} = \mathbf{x}^{(k-1)} - \rho \Phi^{\top} \left( \Phi \mathbf{x}^{(k-1)} - \mathbf{y} \right),
\end{equation}
\begin{equation}
    \mathbf{x}^{(k)} = \operatorname{prox}(\mathbf{z}^{(k)}) = \arg\min_{\mathbf{x}} \left( \lambda \Psi(\mathbf{x}) + \frac{1}{2\rho} \left\| \mathbf{x} - \mathbf{z}^{(k)} \right\|_2^2 \right),
\end{equation}
\begin{equation}
    \mathbf{x}^{(k)} = \arg\min_{\mathbf{x}} \left( \frac{1}{2} \left\| \mathbf{x} - \mathbf{z}^{(k)} \right\|_2^2 + \rho \lambda \Psi(\mathbf{x}) \right),
\end{equation}
where $k$ denotes the iteration step index, $\rho$ is the step size, and $\Psi(\cdot)$ is a handcrafted or deep prior.  
Moreover, the above formulation can be regarded as a denoising process with Gaussian noise at level $\sqrt{\rho \lambda}$.  
This can be simplified as:  
\begin{equation}
    \mathbf{x}^{(k)} = \mathrm{denoiser}_{\Psi}\left( \mathbf{z}^{(k)}, \rho, \lambda \right),
\end{equation}
which denotes the denoiser targeted to $\Psi$ with inputs $\mathbf{z}^{(k)}$, $\rho$, and $\lambda$.

\textbf{HQS:} HQS aims to decouple the fidelity term and the regularization term by introducing an auxiliary variable for optimization problems.  
According to the IR problem and model shown below:  
\begin{equation}
    \mathbf{y} = \Phi \mathbf{x} + \mathbf{n},
\end{equation}
\begin{equation}
    \hat{\mathbf{x}} = \arg\min_{\mathbf{x}} \frac{1}{2} \left\| \mathbf{y} - \Phi \mathbf{x} \right\|_2^2 + \lambda \Psi(\mathbf{x}),
\end{equation}
HQS first introduces an auxiliary variable $\mathbf{w}$ into the optimization model:  
\begin{equation}
    \hat{\mathbf{x}} = \arg\min_{\mathbf{x}} \left\{ \frac{1}{2} \left\| \mathbf{y} - \Phi \mathbf{x} \right\|_2^2 + \lambda \Psi(\mathbf{w}) \right\}, \quad \text{s.t.} \quad \mathbf{w} = \mathbf{x}.
\end{equation}
This constrained model can be transformed into the following augmented Lagrangian (or penalty) objective:  
\begin{equation}
    \mathcal{L}(\mu, \mathbf{x}, \mathbf{w}) = \frac{1}{2} \left\| \mathbf{y} - \Phi \mathbf{x} \right\|_2^2 + \lambda \Psi(\mathbf{w}) + \frac{\mu}{2} \left\| \mathbf{w} - \mathbf{x} \right\|_2^2,
\end{equation}
where $\mu$ is the penalty parameter.  
As a result, the objective function can be naturally divided into two subproblems, which are solved for $\mathbf{x}$ and $\mathbf{w}$ respectively. The iterative updates are as follows:  
\begin{equation}
    \mathbf{x}^{(k)} = \arg\min_{\mathbf{x}} \left\{ \left\| \mathbf{y} - \Phi \mathbf{x} \right\|_2^2 + \mu \left\| \mathbf{w}^{(k-1)} - \mathbf{x} \right\|_2^2 \right\},
\end{equation}
\begin{equation}
    \mathbf{x}^{(k)} = \left( \Phi^{\top} \Phi + \mu I \right)^{-1} \left( \Phi^{\top} \mathbf{y} + \mu \mathbf{w}^{(k-1)} \right),
\end{equation}
and  
\begin{equation}
    \mathbf{w}^{(k)} = \arg\min_{\mathbf{w}} \left\{ \frac{1}{2} \left\| \mathbf{w} - \mathbf{x}^{(k)} \right\|_2^2 + \frac{\lambda}{\mu} \Psi(\mathbf{w}) \right\}.
\end{equation}
Thus, different optimization algorithms lead to different effective regularization parameters. This update can be simplified as:  
\begin{equation}
    \mathbf{x}^{(k)} = \mathrm{denoiser}_{\Psi}\left( \mathbf{z}^{(k)}, \mu, \lambda \right),
\end{equation}
which denotes the denoiser targeted to $\Psi$ with inputs $\mathbf{z}^{(k)}$, $\mu$, and $\lambda$.

\noindent\textbf{Visualization of the LoRA modules}
The visualization of the LoRA modules are shown in Fig. \ref{fig:lora_visual} which clearly illustrates the distinct, structured patterns in the weight matrices for different stages, verifying that LoRun is dynamic and stage-specific.
\begin{figure}
    \centering
    \includegraphics[width=1\linewidth]{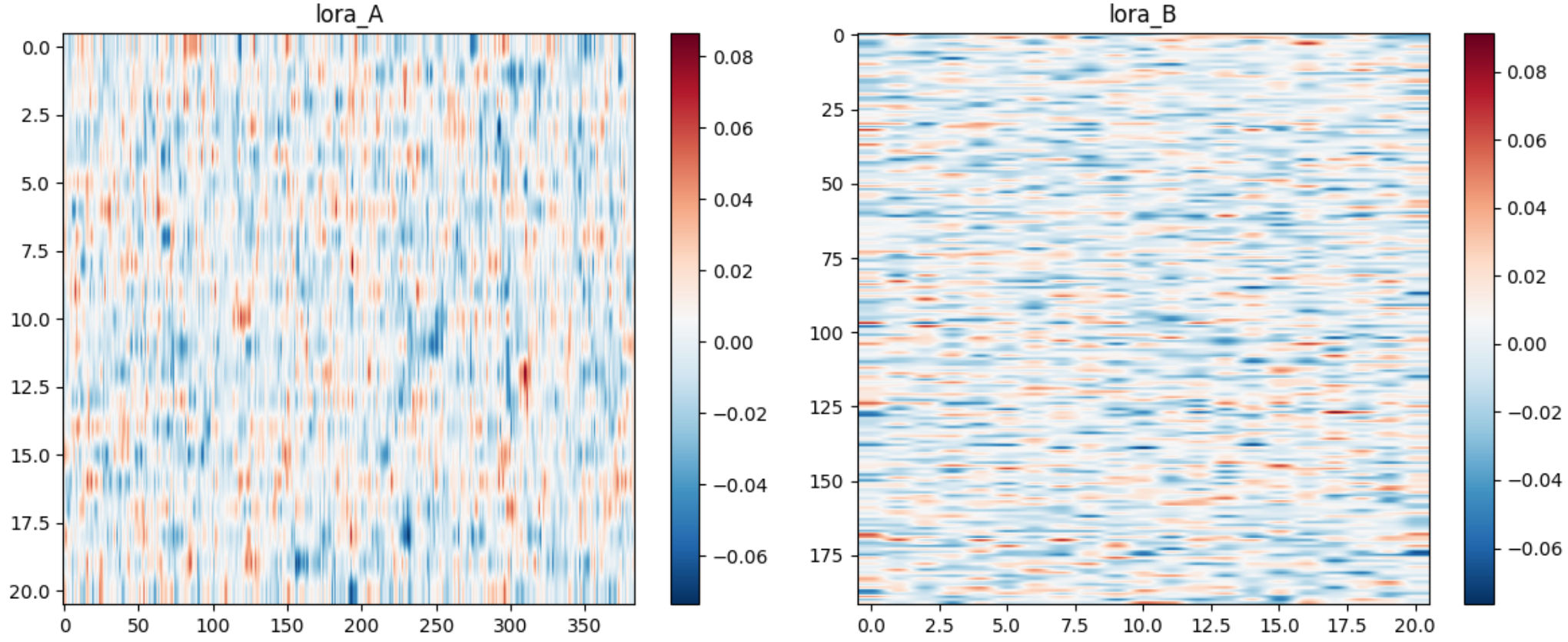}\\
    \includegraphics[width=1\linewidth]{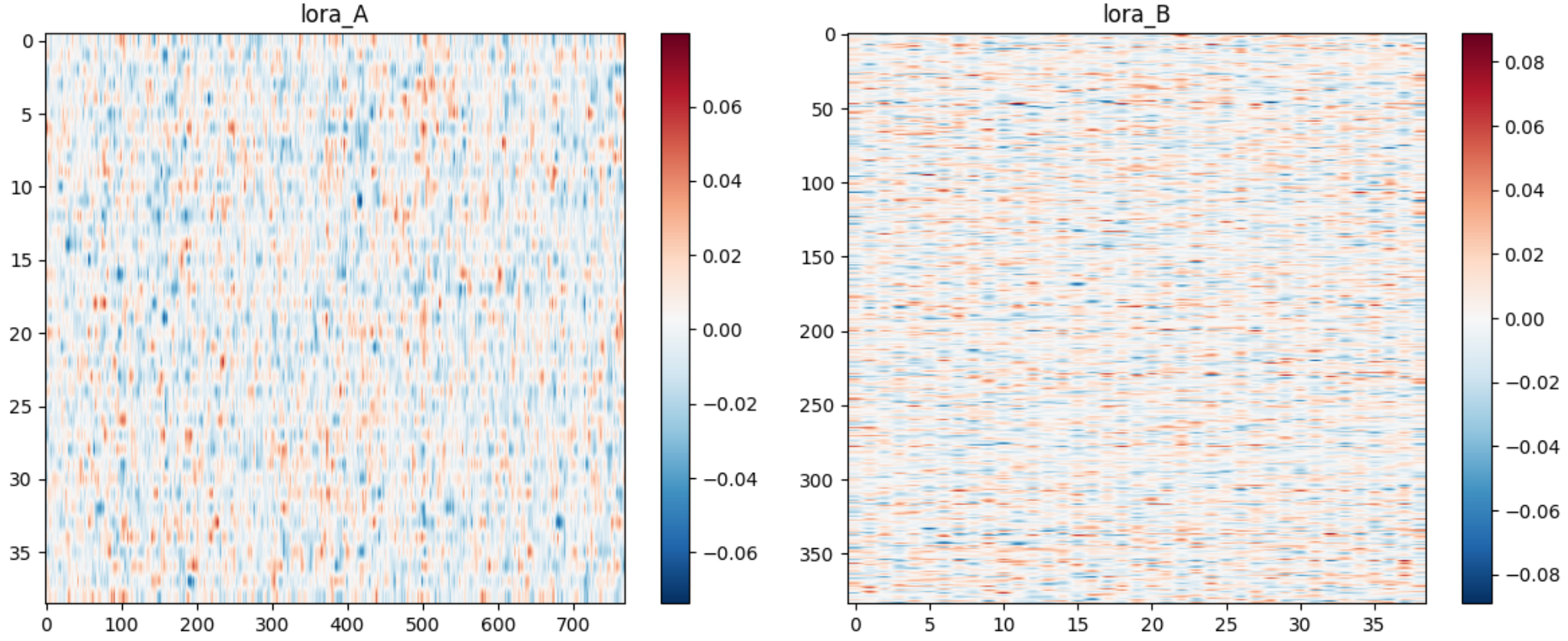}
    \caption{Visualization of the selected LoRA modules.}
    \label{fig:lora_visual}
\end{figure}
}

\begin{figure}
	\footnotesize
	\setlength{\tabcolsep}{1pt}
 \newcommand{\tabincell}[2]{\begin{tabular}{@{}#1@{}}#2\end{tabular}}
	\begin{center}
            \scalebox{1}{
		\begin{tabular}{cc}
			\includegraphics[width=0.15\textwidth]{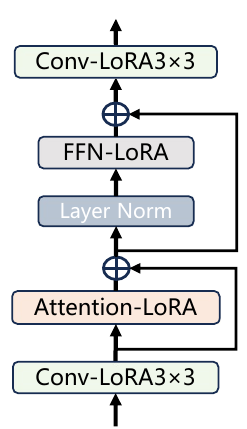}&
			\includegraphics[width=0.15\textwidth]{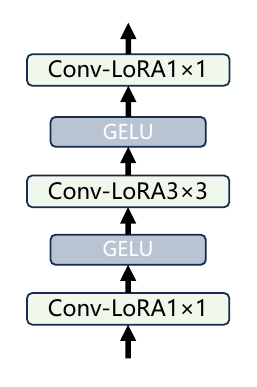}\\
          (a) Transformer Denoiser &(b) Feed Forward Network (FFN)\\
			\multicolumn{2}{c}{\includegraphics[width=0.43\textwidth]{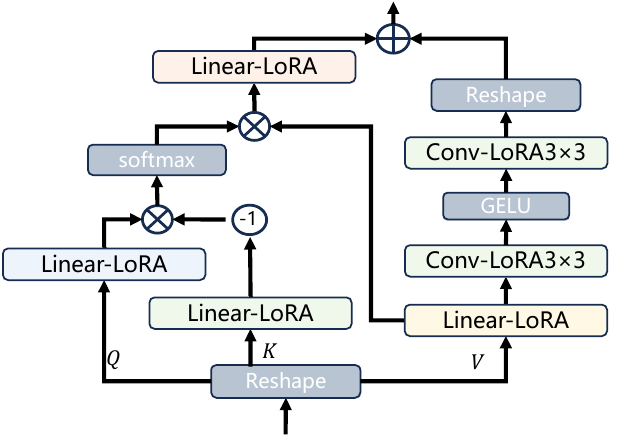}}\\
            \multicolumn{2}{c}{(c) Attention-LoRA Network}
		\end{tabular}}
\caption{Structures of the Transformer denoiser and its components. (a) The Transformer denoiser which aims to demonstrate the performance of our LoRun in simple but classical modules. (b) The feed forward network of the Transformer. (c) The attention network induced with LoRA of the Transformer.}
		\label{fig:transformer_denoiser_structure}
	\end{center}
\end{figure}

\begin{figure}
	\footnotesize
	\setlength{\tabcolsep}{1pt}
 \newcommand{\tabincell}[2]{\begin{tabular}{@{}#1@{}}#2\end{tabular}}
	\begin{center}
            \scalebox{1}{
		\begin{tabular}{cc}
			\includegraphics[width=0.33\textwidth]{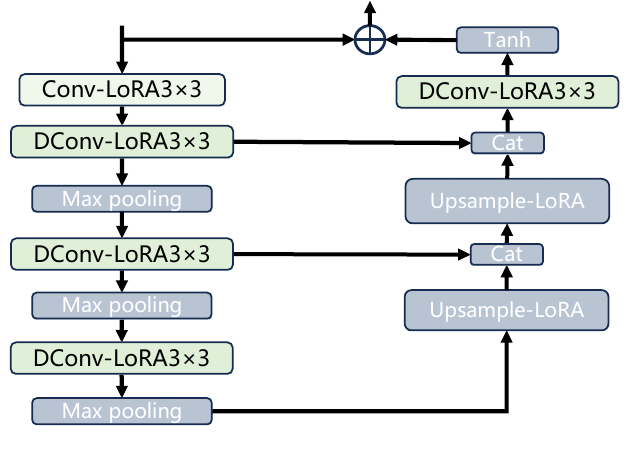}&
			\includegraphics[width=0.15\textwidth]{fig/ffn_lora.pdf}\\
          (a) U-Net Denoiser &(b) Dconv-LoRA\\
		\end{tabular}}
\caption{Structures of the U-Net denoiser and its components. (a) The U-Net denoiser which aims to demonstrate the performance of our LoRun in simple but classical modules. (b) The double-convolution module induced with LoRA.}
		\label{fig:unet_denoiser_structure}
	\end{center}
\end{figure}

\noindent\textbf{U-Net and Transformer}
Fig. \ref{fig:transformer_denoiser_structure} and Fig. \ref{fig:unet_denoiser_structure} present the detailed frameworks about the classical Transformer and U-Net denoisers.
Fig. \ref{fig:transformer_denoiser_structure} (a) is the main module, (b) is the feed forward network (FFN) of the Transforme and (c) is the attention network induced with LoRA.
Fig. \ref{fig:unet_denoiser_structure} (a) is the main U-Net framework, which contains the (b) double convolution (DConv) module and the (c) upsample network for better resolution alignment.

\begin{figure}
	\footnotesize
	\setlength{\tabcolsep}{1pt}
 \newcommand{\tabincell}[2]{\begin{tabular}{@{}#1@{}}#2\end{tabular}}
	\begin{center}
            \scalebox{1}{
		\begin{tabular}{cc}
			\includegraphics[width=0.24\textwidth]{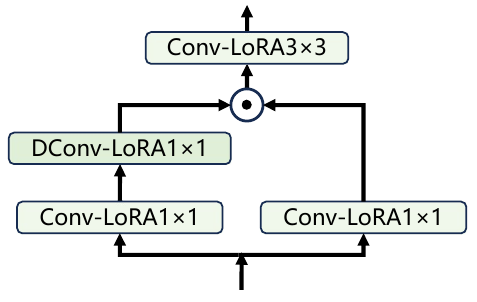}&
			\includegraphics[width=0.24\textwidth]{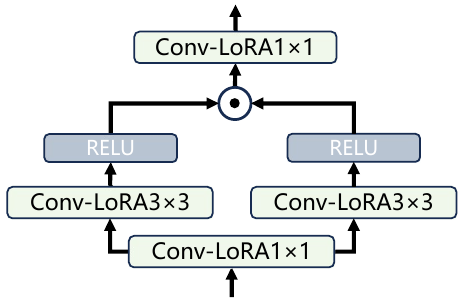}\\
          (a) Convolution Modulated Block (CMB)&(b) Spatial Alignment Block (SAB)\\
			\multicolumn{2}{c}{\includegraphics[width=0.5\textwidth]{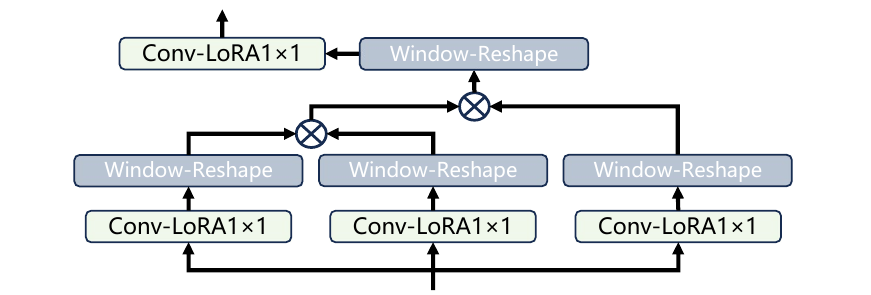}}\\
            \multicolumn{2}{c}{(c) Window-based Spectra wise Self-Attention (WSSA)}
		\end{tabular}}
\caption{Structures of the components of the denoiser for CASSI task. (a) The convolution modulated block of the denoiser. (b) The spatial alignment block. (c) The window-based spectra wise self-attention with LoRA.}
		\label{fig:CASSI_denoiser_structure}
	\end{center}
\end{figure}

\noindent\textbf{Coded Aperture Snapshot Spectral Imaging}
Fig. \ref{fig:CASSI_denoiser_structure} shows the structure of the components of the denoiser which we adopt for CASSI task.
And we follow the block parameter settings as SSR \cite{zhang2024improving} method.
In this figure, (a) denotes the convolution modulated block (CMB), (b) is the spatial alignment block (SAB) and (c) represents the window-based spectral wise self-attention (WSSA) network, which better generates the spectral priors of the multi-spectral images within multiple windows.

\noindent\textbf{Compressive Sensing}
Fig. \ref{fig:CASSI_denoiser_structure} shows the structure of the denoiser which we adopt for CS task.
We follow the block parameter settings as CPPNet \cite{guo2024cpp} method.
In this figure, (a) denotes the depth-wise channel attention block (DCAB) , (b) is the multi-scale transformer block (MSTB) and (c) represents the denoiser, which is the dual path fusion block (DPFB) network, promoting the multi-level information flow for CS.

\begin{figure*}
	\footnotesize
	\setlength{\tabcolsep}{1pt}
 \newcommand{\tabincell}[2]{\begin{tabular}{@{}#1@{}}#2\end{tabular}}
	\begin{center}
            \scalebox{0.9}{
		\begin{tabular}{cc}
			\includegraphics[width=0.4\textwidth]{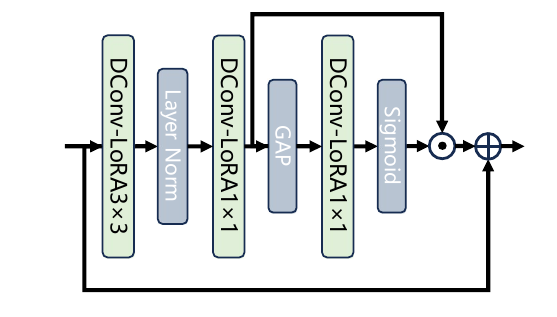}&
			\includegraphics[width=0.6\textwidth]{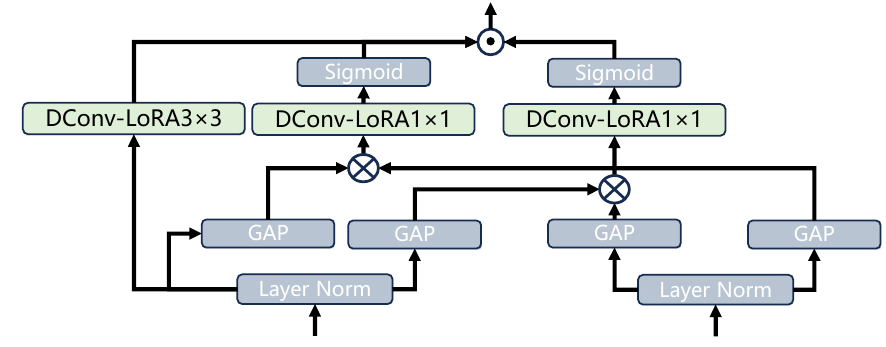}\\
          (a) Depth-wise Channel Attention &(b) Multi-Scale Transformer\\
           Block (DCAB) & Block (MSTB)\\
			\multicolumn{2}{c}{\includegraphics[width=1\textwidth]{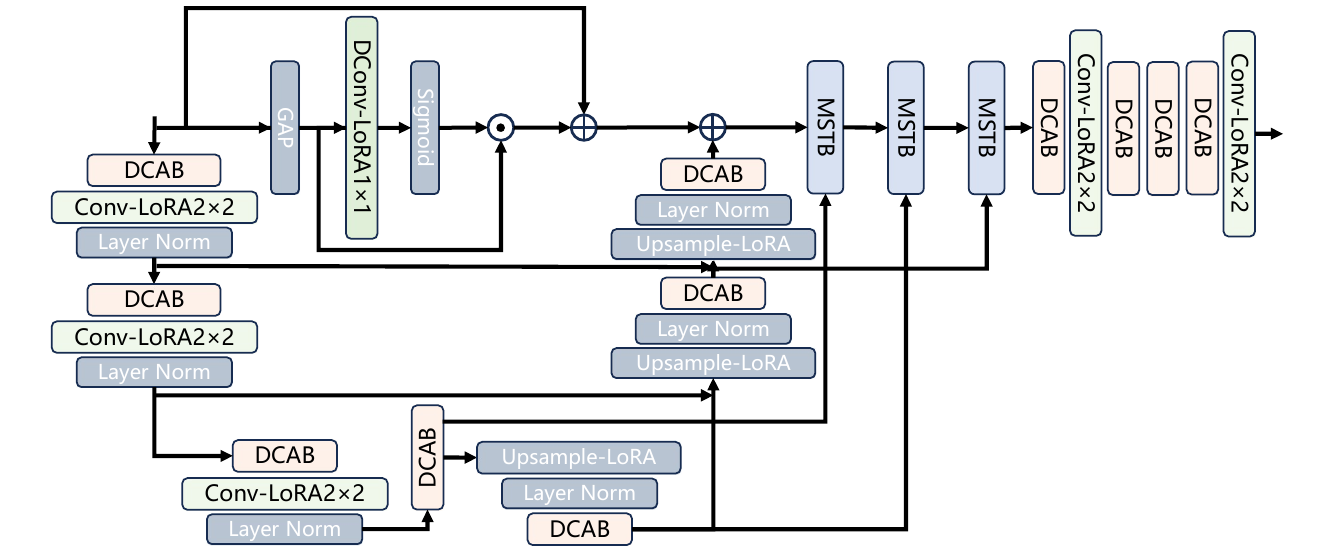}}\\
            \multicolumn{2}{c}{(c) Denoiser: Dual Path Fusion Block (DPFB)}
		\end{tabular}}
\caption{Structures of the denoiser and its components for CS task. (a) The depth-wise channel attention block. (b) The multi-scale transformer module with LoRA. (c) The denoiser which is the dual path fusion block for CS task.}
		\label{fig:CS_denoiser_structure}
	\end{center}
\end{figure*}

\begin{figure}
    \centering
    \includegraphics[width=1\linewidth]{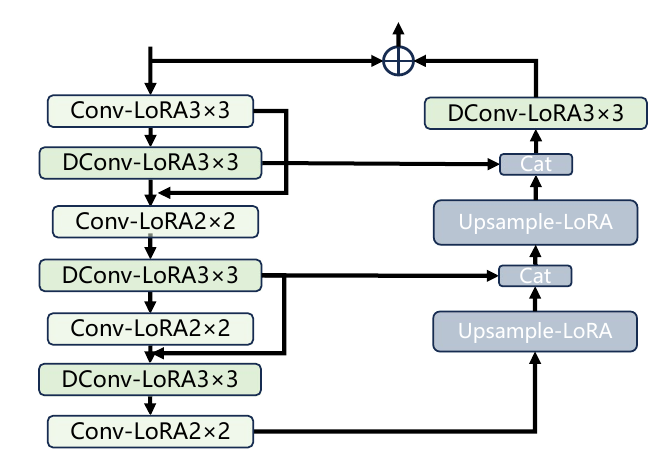}
    \caption{Structures of the denoiser and its components for SR task. This architecture is based on the convolution-LoRA module and upsample block.}
    \label{fig:SR_denoiser_structure}
\end{figure}

\noindent\textbf{Image Super Resolution}
Fig. \ref{fig:SR_denoiser_structure} shows the structure of the denoiser which we adopt for SR task.
We follow the block parameter settings as USRNet \cite{zhang2020deep} method.
The basic block is the classical residual convolution module.
\end{document}